\documentclass[sigconf,screen]{acmart}

\copyrightyear{2022} 
\acmYear{2022} 
\setcopyright{rightsretained}

\acmConference[MM '22]{Proceedings of the 30th ACM International Conference on Multimedia}{October 10--14, 2022}{Lisboa, Portugal}
\acmBooktitle{Proceedings of the 30th ACM International Conference on Multimedia (MM '22), Oct. 10--14, 2022, Lisboa, Portugal}
\acmISBN{978-1-4503-9203-7/22/10}
\acmDOI{10.1145/3503161.3548167}

\usepackage{etoolbox}
\makeatletter
\patchcmd{\maketitle}{\@copyrightpermission}{
   \begin{minipage}{0.3\columnwidth}
     \href{https://creativecommons.org/licenses/by/4.0/}{\includegraphics[width=0.90\textwidth]{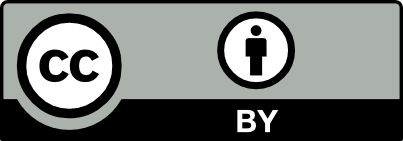}}
   \end{minipage}\hfill
   \begin{minipage}{0.7\columnwidth}
     \href{https://creativecommons.org/licenses/by/4.0/}{This work is licensed under a Creative Commons Attribution International 4.0 License.}
   \end{minipage}
  
   \vspace{5pt}
}{}{}

\makeatother

\usepackage{soul}
\usepackage[utf8]{inputenc}
\usepackage[list=true]{subcaption} %
\usepackage{graphicx}
\usepackage{amsmath}
\usepackage[titletoc]{appendix}
\usepackage{multirow}
\usepackage{algorithm}
\usepackage{algpseudocode} %
\newtheorem{theorem}{Theorem}

\begin{document}

\title{Domain Adaptation for Time-Series Classification to Mitigate Covariate Shift}

\author{Felix Ott}
\orcid{0000-0002-4392-0830}
\affiliation{%
  \institution{Fraunhofer IIS, Fraunhofer Institute for Integrated Circuits}
  \city{Nürnberg}
  \postcode{90411}
  \country{Germany}
}
\affiliation{%
  \institution{LMU Munich}
  \city{}
  \postcode{}
  \country{}
}
\email{felix.ott@iis.fraunhofer.de}

\author{David Rügamer}
\orcid{0000-0002-8772-9202}
\affiliation{%
  \institution{LMU Munich}
  \city{Munich}
  \postcode{80539}
  \country{Germany}
}
\affiliation{%
  \institution{RWTH Aachen}
  \city{}
  \postcode{}
  \country{}
}
\email{david.ruegamer@stat.uni-muenchen.de}

\author{Lucas Heublein}
\orcid{0000-0001-6670-3698}
\affiliation{%
  \institution{Fraunhofer IIS, Fraunhofer Institute for Integrated Circuits}
  \city{Nürnberg}
  \postcode{90411}
  \country{Germany}
}
\email{heublels@iis.fraunhofer.de}

\author{Bernd Bischl}
\orcid{0000-0001-6002-6980}
\affiliation{%
  \institution{LMU Munich}
  \city{Munich}
  \postcode{80539}
  \country{Germany}
}
\email{bernd.bischl@stat.uni-muenchen.de}

\author{Christopher Mutschler}
\orcid{0000-0001-8108-0230}
\affiliation{%
  \institution{Fraunhofer IIS, Fraunhofer Institute for Integrated Circuits}
  \city{Nürnberg}
  \postcode{90411}
  \country{Germany}
}
\email{christopher.mutschler@iis.fraunhofer.de}

\renewcommand{\shortauthors}{Felix Ott et al.}
\begin{abstract}
The performance of a machine learning model degrades when it is applied to data from a similar but different domain than the data it has initially been trained on. To mitigate this domain shift problem, domain adaptation (DA) techniques search for an optimal transformation that converts the (current) input data from a source domain to a target domain to learn a domain-invariant representation that reduces domain discrepancy. This paper proposes a novel supervised DA based on two steps. First, we search for an optimal class-dependent transformation from the source to the target domain from a few samples. We consider optimal transport methods such as the earth mover's distance, Sinkhorn transport and correlation alignment. Second, we use embedding similarity techniques to select the corresponding transformation at inference. We use correlation metrics and higher-order moment matching techniques. We conduct an extensive evaluation on time-series datasets with domain shift including simulated and various online handwriting datasets to demonstrate the performance.
\end{abstract}



\maketitle

\section{Introduction}
\label{chap_introduction}

\begin{figure}[t!]
	\centering
    \includegraphics[trim=10 10 10 10, clip, width=1.0\linewidth]{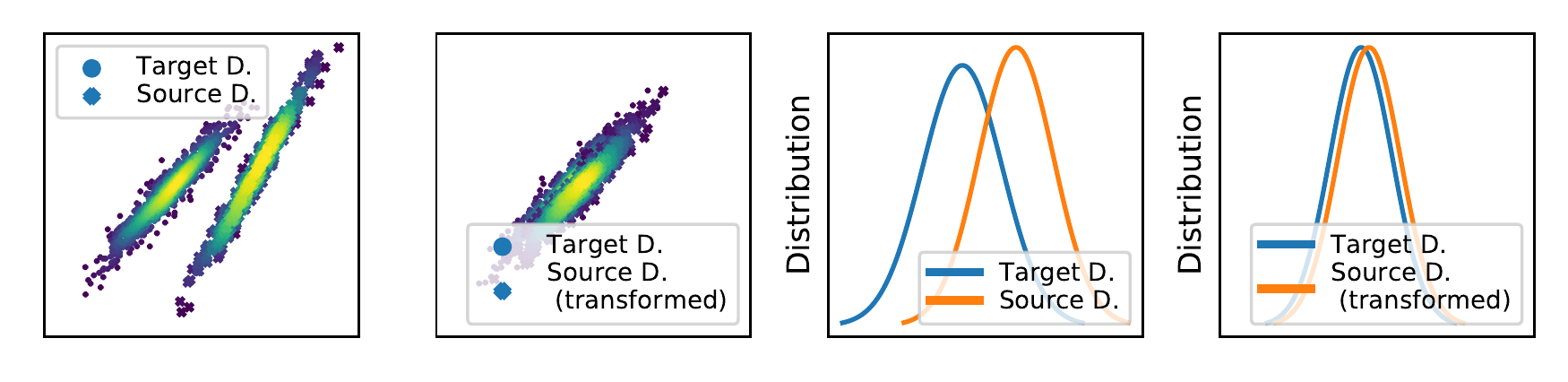}
    \vspace{-0.3cm}
    \caption{Domain adaptation. To compensate the domain shift the source data is transformed into the target data (1\textsuperscript{st}: 2D source and target domain features, before transformation, 2\textsuperscript{nd}: after transformation). 3\textsuperscript{rd} and 4\textsuperscript{th}: their distributions.}
    \label{image_domain_adaptation}
\end{figure}

Traditional machine learning (ML) algorithms assume training and test datasets to be \textit{independent and identically distributed} (i.i.d.). Hence, supervised ML only works well when the test data comes from the same distribution as the training data. As real-world data often changes over time and space, this assumption rarely holds in practice \cite{sun_feng}. Domain adaptation (DA) \cite{daume, saenko, fernando, gong, long_wang, sun_saenko, tzeng} as a special instance of transfer learning (TL) \cite{pan_yang, shao_zhu, day_khoshgoftaar, patel_gopalan} tries to compensate for this \textit{domain shift} by transferring knowledge from a \textit{source} to a \textit{target} domain, see Figure~\ref{image_domain_adaptation}. There are three types of DA: supervised, semi-supervised, and unsupervised DA. The decision which approach to use mainly depends on the number of labels available in the target domain \cite{zhang}. Most techniques are unsupervised and transform the source data by minimizing the distance to the target data. Typically, the model is (re-)trained on the transformed source domain \cite{sun_feng}.

Domain shifts appear in many applications such as classification \cite{karim_majumdar, zou_wang, zerveas_jayaraman, wang_yan_oates, he_zhang, rahimian_zabihi, fawaz_lucas}, handwriting recognition \cite{klass_lorenz}, segmentation and regression for multimedia data \cite{chen_zhang, ghifary}, for example if the background, shape deformation, or quality are different across domains. DA aims to mitigate this and has successfully been applied for object recognition \cite{gong_grauman, luo_liu, yan_wu, yang_jing}, AI planning \cite{zhuo_yang}, reinforcement learning \cite{nguyen_silander} and natural language processing \cite{zhao_hoi} (e.g., the adaptation from English to Spanish documents \cite{liu_zhang}).

\begin{figure}[t!]
	\centering
    \includegraphics[trim=9 30 11 10, clip, width=1.0\linewidth]{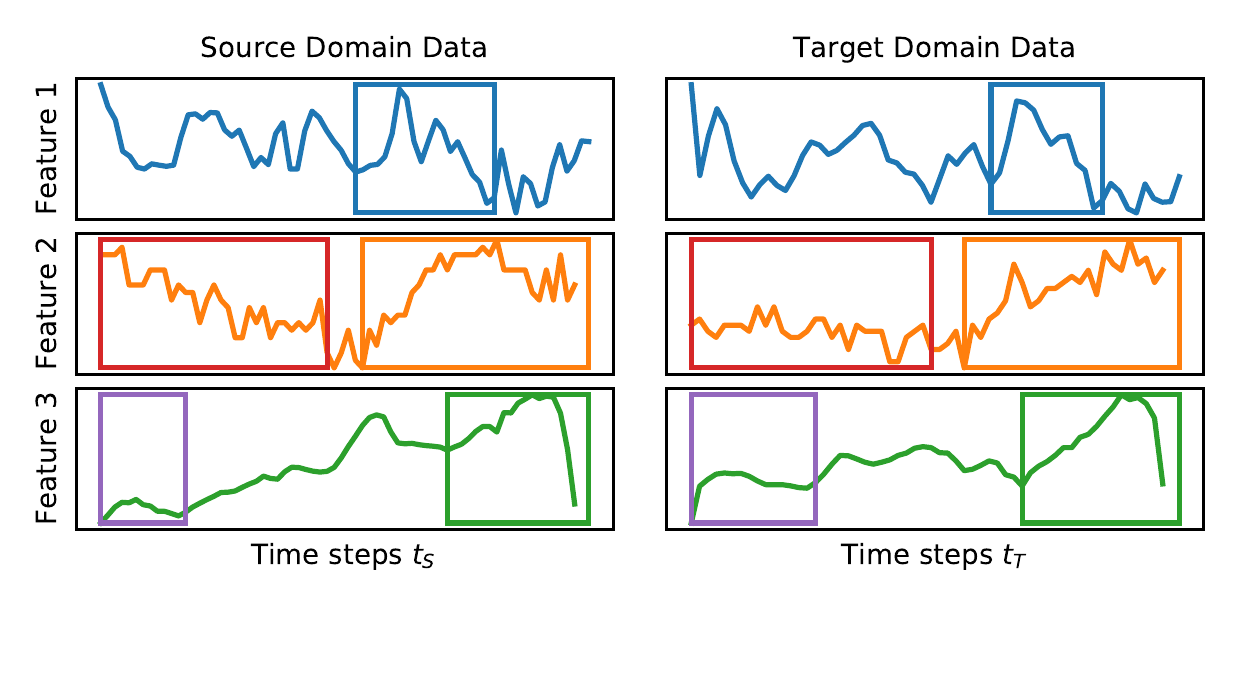}
    \caption{Domain-variant representation of MTS data of a source and target sample from a sensor pen representing the label '\texttt{5}' in an online handwriting task. Note that both samples have different number of time steps $t_S$ (left) and $t_T$ (right). Top row: data from an accelerometer; middle row: gyroscope measures; bottom row: data from a force sensor.}
    \label{image_da_time_series}
    \vspace{-0.1cm}
\end{figure}

DA is also used for multivariate time-series (MTS) classification and forecasting \cite{cai_chen, jin_park, chung_kastner, hu_tang}, which is a challenging task as the extraction of domain-invariant representations is non-trivial. Consider the different time-series of online handwriting from a sensor-enhanced pen in Figure~\ref{image_da_time_series} (right: right-handed writers, target domain; left: left-handed writers, source domain) \cite{ott, ott_tist, klass_lorenz}. The shaded areas show the discrepancy between both writers, i.e., different time step lengths and accelerations. In this case, the complex dependency of time steps in the MTS makes it challenging to extract invariant features \cite{cai_chen}. Many existing methods \cite{costa_akcay, purushotham} employ recurrent neural networks (RNNs) and assume that the conditional distributions of the source and target domain are equal, i.e., $P_S\big(y|\phi(x_1, \ldots, x_{t_S})\big) = P_T\big(y|\phi(x_1, \ldots, x_{t_T})\big)$, with the feature transformation mapping $\phi(\cdot)$ and $\mathbf{X} = \{x_1, x_2, \ldots, x_t\} \in \mathcal{X}$ being a set of training samples with $t \in \{ t_S, t_T\}$ time steps \cite{pan}. However, this assumption does typically not hold in practice as methods do not generalize across domains without additional efforts.

Classical DA methods range from feature selection \cite{bay_tuytelaars, rashedi_nez} (both domains share similarities in the features), distribution adaptation (distributions of both domains are different but share similarities), and subspace learning (a lower-dimensional shared representation). Distribution adaptation methods can be classified into three categories: (1) \textit{Marginal distribution adaptation} methods assume that the marginal distribution between the domains are different and focus on overall shape alignment. While the most established method is maximum mean discrepancy (MMD) \cite{long_cao}, e.g., used in \cite{ghifary, long_wang, long_zhu, tzeng, huang_peng, long_zhu_mmd, chen_fu}, many further techniques exist \cite{pan, dorri, duan, long_wang, baktashmotlagh, jiang}. (2) \textit{Conditional distribution adaptation} \cite{satpal_sarawagi, wang_chen_hu} assumes that the conditional distribution is varied between the domains $\big(P(\mathcal{Y}_S | \mathcal{X}_S) \neq P(\mathcal{Y}_T |\mathcal{X}_T)\big)$. (3) \textit{Joint distribution adaptation} methods \cite{long_wang2, long_wang3} minimize the joint distribution distance between the source and target domain. Subspace alignment methods (i.e., SA~\cite{fernando}, CORAL~\cite{sun_saenko}, GFK~\cite{gong}) align the source and target domains via principal component analysis with a lower dimensional space determined by the Bregman divergence. Recently, DL methods have become the predominant approach in DA. Existing techniques are, e.g., based on MMD~\cite{tzeng, long_cao}, align the second-order statistics (covariances) \cite{sun_saenko2}, use the Kullback-Leibler \cite{zhuang_cheng} or Jensen-Shannon divergence \cite{jiang_wang}, or are based on the Wasserstein distance \cite{courty_flamary, flamary_courty, pot_flamary_courty}. These methods have been broadly applied to visual object recognition and text categorization \cite{baktashmotlagh}, but rarely to time-series \cite{cai_chen}.

We propose a DA method that adapts embeddings from a small source domain (with domain shift) to embeddings from a large target domain (main training dataset on which the model has initially been trained). First, we pre-train our model on a large target domain dataset. Next, we train an optimal transformation $\mathbf{T}$ from the source to the target domain for each class with joint distribution adaptation methods, i.e., optimal transport and CORAL, from an adaptation set with few samples. At inference, we extract features with the source domain model, transform features into the target domain for each class, compute the similarity to the target domain to select the best transformation, and classify the transformed embedding with the target domain model. This allows a faster adaptation to new data without the necessity of post-training \cite{houlsby}. We apply this technique to MTS datasets and evaluate the performance of methods for the challenging task of embedding similarity comparison from time-series data \cite{cai_chen}. We show performance improvements for synthetically generated univariate sinusoidal data and on multivariate online handwriting (OnHW) datasets from pens with integrated sensors. For the challenging OnHW recognition task of new out-of-distribution writers, we propose a method that can adapt to each writer and outperforms transfer learning approaches.\footnote{Datasets and source code available at: \href{https://www.iis.fraunhofer.de/de/ff/lv/dataanalytics/anwproj/schreibtrainer/onhw-dataset.html}{www.iis.fraunhofer.de/de/ff/lv/dataanalytics/ anwproj/schreibtrainer/onhw-dataset.html}}

The remainder of this paper is organized as follows. Section~\ref{chap_related_work} discusses related work followed by our proposed methodology in Section~\ref{chap_method}. The experimental setup is described in Section~\ref{chap_experiments} and the results are discussed in Section~\ref{chap_results}. Section~\ref{chap_summary} concludes.

\section{Related Work}
\label{chap_related_work}

Research for (multivariate) time-series classification is very advanced and ranges from classical convolutional neural networks (CNNs) such as FCN~\cite{wang_yan_oates} to advanced CNNs~\cite{tan_dempster, tang_long} such as ResCNN \cite{zou_wang}, ResNet~\cite{wang_yan_oates}, XResNet~\cite{he_zhang}, XceptionTime~\cite{rahimian_zabihi} and InceptionTime~\cite{fawaz_lucas}. Spatio-temporal methods \cite{ott_tist, chung_gulcehre, karim_majumdar} became popular with the development of RNNs, i.e., long short-term memorys (LSTMs) and multi-dimensional LSTMs. Research for Transformers (e.g., TST~\cite{zerveas_jayaraman}) for time-series classification is less advanced. As their goal is to classify a (multivariate) time-series without domain shift (e.g., on the UCR~\cite{dau_bagnall} datasets), they do not consider to transform embeddings. All such methods have previously been benchmarked on OnHW datasets, and \cite{ott_tist} showed a benefit of small CNNs combined with bidirectional LSTMs (BiLSTMs) on OnHW recognition.

\paragraph{Domain Adaptation for Time-Series Classification.} Research on MTS-specific DA is rare. The domain adversarial neural network (DANN) learns domain-invariant features \cite{costa_akcay} and uses a time window approach to extract temporal information from time-series data for prognostics with LSTMs. Similar, the variational recurrent adversarial deep domain adaptation (VRADA) \cite{purushotham} learns domain-invariant temporal relationships based on a variational RNN (VRNN) \cite{chung_kastner} for MTS healthcare datasets. However, both of them cannot align the condition distribution well. Sparse associate structure alignment (SASA) \cite{cai_chen} exploits the sparse associative structure to mitigate the difficult domain-invariant extraction of time-series for offsets (change of time lags, Figure~\ref{image_da_time_series}). SASA generates a segment set to exclude the obstacle of offsets, extracts associative structure time-series data with time lags, and uses the structure alignment for knowledge transfer. The domain adaptation forecaster (DAF) \cite{jin_park} leverages statistical strengths from the source domain to improve the performance on the target domain for forecasting. The attention-based shared module with a domain discriminator across domains induces domain-invariant latent features and retrains domain-specific features. However, all those methods require learning latent features between domains, while applications on embedded devices require approaches to quickly adapt to a new domain at inference. Similar to our approach but for forecasting, Hu et al.~\cite{hu_tang} use pre-trained cross-domain time-series representations to augment the target domain. Wilson et al.~\cite{wilson_doppa} proposed a DA model with weak supervision and a domain classifier with a domain-invariant representation, while we use a target-specific representation.

\paragraph{Covariance- and Discrepancy-based Methods for DA} Procrustes analysis \cite{wang_mahadevan} uses the singular value decomposition for manifold alignment, but has a run time complexity of $O(N^3)$ for $N$ samples. Correlation alignment (CORAL) \cite{sun_feng, chen_chen, sun_saenko} is a simple method and can be used in a (non-)trainable manner for embedding alignment, but has not been used on time-series data before. We make use of CORAL for transformation computations and selection. Many DA methods are based on MMD \cite{long_cao} that minimizes the discrepancy of feature distributions, for example, joint MMD (JMMD) \cite{long_zhu_mmd} or transfer joint matching (TJM) \cite{long_wang}. Chen et al. \cite{chen_fu} proposed a higher-order moment matching (HoMM) technique and extend HoMM into a reproducing kernel Hilbert space (RKHS) \cite{gong, zhang_wang, duan, huang}. The first order HoMM is equivalent to MMD, and the second-order HoMM is equivalent to CORAL. Higher orders ($\geq 3$), however, result in long training times. These methods do not directly compute a transformation between source and target domains, but minimize the discrepancy at training time, and hence, we use these methods for transformation selection. Transfer component analysis (TCA) \cite{pan} learns transfer components in an RKHS using MMD. Joint distribution adaptation (JDA) \cite{long_wang2} reduces the difference between source and target domains in both the marginal and conditional distributions by a principled dimensionality reduction procedure based on MMD. This approach reduces the embedding dimensions and cannot be applied for cross-domain feature classification as given in our application.

\paragraph{Optimal Transport for DA} The earth mover's distance (EMD) and Sinkhorn transport \cite{courty_flamary, flamary_courty, ferradans} has rarely been used for time-series DA, but successfully for image DA. The Python Optimal Transport (POT) package \cite{pot_flamary_courty} has been used for adapting domains for the classification of satellite images \cite{tardy}. Images are considered as time-series as they are an ordered set that is re-sampled onto a regular time grid for consistent length ($<$ 30 images per time-series). For an overview, see \cite{fatras}. We provide a broad evaluation of EMD and Sinkhorn transport for time-series DA and show benefits over correlation techniques.

\paragraph{Embedding Distances.} Related work for comparing embeddings commonly use the Euclidean metric, but also correlation-based metrics are used \cite{ott_wacv, ott_ijcai}. Recent methods extend the canonical correlation analysis (CCA) \cite{ranjan, chapman} that learn linear projection matrices by maximizing pairwise correlation, but require long computing times. Metrics typically used for DA can also be utilized as distance metrics between embeddings (e.g., CORAL~\cite{sun_feng}, MMD~\cite{long_cao, long_zhu_mmd, chen_fu} and HoMM~\cite{chen_fu}). \cite{ott_ijcai} use the cross- and Pearson correlation \cite{pearson} for similarity computation. Our evaluation proposes performance comparisons of these methods for the time-series DA application.

\section{Methodology}
\label{chap_method}

We start with a formal definition of the notation for domain adaptation (DA) in Section~\ref{chap_notation}. We then give an overview of our method that consists of two parts: Optimizing a transformation from source to target domain and selecting the transformation at inference (Section~\ref{chap_method_overview}). An overview of all the introduced notation is additionally given in Appendix~\ref{sec_app_notation}.

\subsection{Notation}
\label{chap_notation}

\paragraph{MTS Classification.} An MTS $\mathbf{U} = \{\mathbf{u}_1,\ldots,\mathbf{u}_m\} \in \mathbb{R}^{m \times l}$ is an ordered sequence of $l \in \mathbb{N}$ streams with $\mathbf{u}_i = (u_{i,1},\ldots, u_{i,l}), i\in \{1,\ldots,m\}$, where $m \in \mathbb{N}$ is the length of the time-series. The MTS training set is a subset of the array $\mathcal{U} = \{\mathbf{U}_1,\ldots,\mathbf{U}_{n_U}\} \in \mathbb{R}^{n_U \times m \times l}$, where $n_U$ is the number of time-series. The aim of MTS classification is to predict an unknown class label $y \in \mathcal{Y}$ for a given MTS. We define the target domain dataset as $\mathcal{U}_T$. Given a smaller adaptation set of a source domain $\mathcal{U}_S$ with MTS $\mathbf{U}_S$, the goal of DA is to find an optimal transformation $\mathbf{T}$ of the representation of the latent embedding $f(\mathbf{U}_S)$ of the source domain to the representation of the latent embedding $f(\mathbf{U}_T)$ of the target domain such that the prediction of the unknown class label $y_S$ of the source domain is maximized. $f(\mathbf{U}_T) \in \mathbb{R}^{q_T \times w_T}$ and $f(\mathbf{U}_S) \in \mathbb{R}^{q_S \times w_S}$ are the latent target and source embeddings of the neural network.

\paragraph{Domain Adaptation.} A domain $\mathcal{D}$ consists of a feature space $\mathcal{X}$ with marginal probability $P(\mathcal{X})$. The task is defined by the label space $\mathcal{Y}$. The joint distribution is $P(\mathcal{X}, \mathcal{Y})$ and the conditional distribution is denoted as $P(\mathcal{Y} | \mathcal{X})$. When considering MTS classification, there is a source domain $\mathcal{D}_S = \{\mathcal{X}_S^i, \mathcal{Y}_S^i\}_{i=1}^{\mathcal{N}_S}$ of $\mathcal{N}_S$ labeled samples of $|\mathcal{Y}^i_S|$ categories, and a target domain $\mathcal{D}_T = \{\mathcal{X}_T^i, \mathcal{Y}_T^i\}_{i=1}^{\mathcal{N}_T}$ of $\mathcal{N}_T$ labeled samples of $|\mathcal{Y}_T^i|$ categories. Due to the difference of the two domains, the distributions are assumed to be different: $P(\mathcal{X}_S) \neq P(\mathcal{X}_T)$ and $P(\mathcal{Y}_S|\mathcal{X}_S) \neq P(\mathcal{Y}_T|\mathcal{X}_T)$ (see Figure~\ref{image_domain_adaptation}, left) \cite{zhang}. DA can mitigate the domain shift and improve the classification accuracy in the target domain (see Figure~\ref{image_domain_adaptation}, 2\textsuperscript{nd} and 4\textsuperscript{th}).

\subsection{Method Overview}
\label{chap_method_overview}

\begin{figure}[t!]
	\centering
    \includegraphics[width=1.0\linewidth]{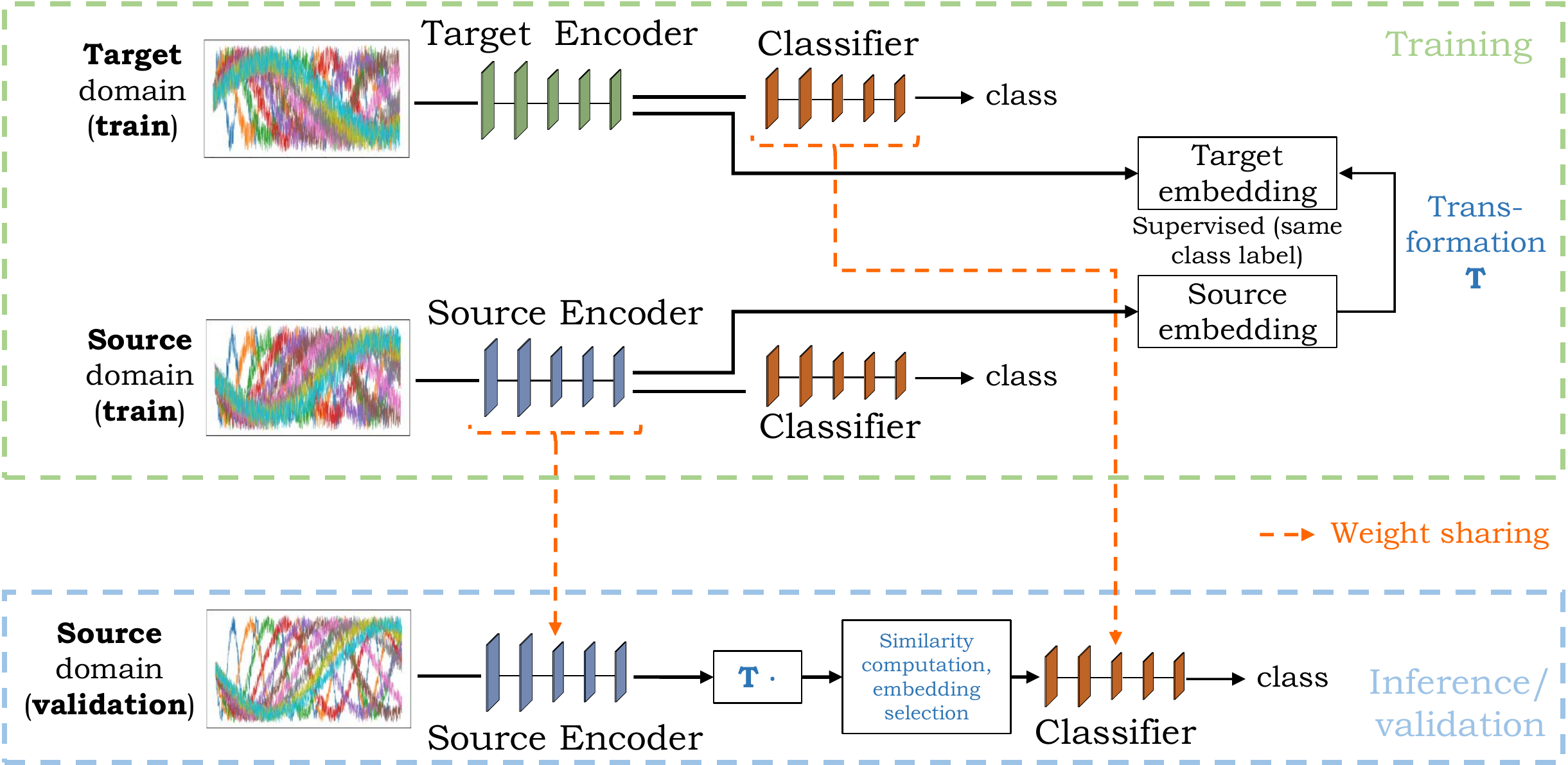}
    \caption{Method overview. We transform the feature embeddings of the source domain (training data, few samples) into feature embeddings of the target domain (training adaptation data, many samples) with optimal transport and correlation alignment. For inference, we evaluate the feature similarity.}
    \label{image_method_overview}
\end{figure}

An overview of our method for MTS classification is given in Figure~\ref{image_method_overview} and in Algorithm~\ref{algorithm_da}. We train a convolutional neural network with two BiLSTM layers on the target data, and train a different (but same) model on the (training) source data. Next, we search for the optimal transformation $\mathbf{T}$ to transform the feature embeddings of the source data onto the feature embeddings of the target data. As feature embeddings $f(\mathbf{U}_S)$ and $f(\mathbf{U}_T)$ we choose the output of the last convolutional layer before the two BiLSTMs of size $\mathbb{R}^{50 \times 30}$ and $\mathbb{R}^{19 \times 200}$. We obtain a separate transformation for each class. To find the optimal transformation, we evaluate different DA techniques, see Section~\ref{chap_emb_trans}. At inference time, we extract features of the (validation) source data with the source domain model, transform the embedding with each class-dependent transformation, and choose the transformation for the closest embedding to the target domain data. For source and target domain similarity computation, see Section~\ref{chap_trans_selection}. Lastly, we classify the transformed embedding with the target domain classifier with the cross-entropy (CE) loss.

\renewcommand{\algorithmicrequire}{\textbf{Input:}}
\renewcommand{\algorithmicensure}{\textbf{Output:}}
\begin{algorithm}[t!]
\caption{Domain adaptation: Source data transformation and transformation selection.}\label{algorithm_da}
\begin{algorithmic}[1]
\Require Target data $\mathcal{U}_{T_t}$, source data $\mathcal{U}_{S_t}$ and $\mathcal{U}_{S_v}$, and labels $\mathcal{Y}_{T_t}$, $\mathcal{Y}_{S_t}$ and $\mathcal{Y}_{S_v}$
\Ensure Class predictions for source validation data $\mathcal{U}_{S_v}$
\Function{DomainAdaptation}{$\mathcal{U}_{T_t}, \mathcal{U}_{S_t}, \mathcal{U}_{S_v}$}
\State Train target model with data $\mathcal{U}_{T_t}$ and labels $\mathcal{Y}_{T_t}$
\State Train source model with data $\mathcal{U}_{S_t}$ and labels $\mathcal{Y}_{S_t}$
\For{$i \gets 1$ to $|\mathcal{Y}_{S_t}|$}
    \State Compute transformation $\mathbf{T}_i$ between $\mathcal{U}_{T_t,i}$ and $\mathcal{U}_{S_t,i}$
    \\ \Comment See Section~\ref{chap_emb_trans}
\EndFor
\For{$i \gets 1$ to $|\mathcal{U}_{S_t}|$}
    \For{$k \gets 1$ to $|\mathcal{Y}_{S_v}|$}
        \State Apply transformation $\mathbf{T}_k$ to source data $\mathcal{U}_{S_v,i}$
        \State Compute similarity between $\mathcal{U}_{S_v,i}(\mathbf{T}_k)$ and $\mathcal{U}_{T_t}$
        \\ \Comment See Section~\ref{chap_trans_selection}
    \EndFor
    \State Select $\mathbf{T}$ with highest similarity of $\mathcal{U}_{S_v,i}(\mathbf{T})$ and $\mathcal{U}_{T_t}$
    \State Predict class label $y_{S_v}$ of transformed $\mathcal{U}_{S_v,i}(\mathbf{T})$
\EndFor
\EndFunction
\end{algorithmic}
\end{algorithm}

\subsubsection{Embedding Transformation}
\label{chap_emb_trans}

The goal of DA is to minimize the target domain error by bounding the source domain error and the discrepancy between them \cite{ben_david_blitzer, ben_david_crammer, blitzer_crammer}. DA approaches consider the target data for optimizing the source domain model and reduce the discrepancy between them as in Theorem~\ref{theorem_da}:
\begin{theorem}[Zhang, 2021 \cite{zhang}]
\label{theorem_da}
Let $\mathcal{H}$ be a hypothesis space. Given the target domain $\mathcal{D}_T$ and source domain $\mathcal{D}_S$, we have
\begin{equation}
    \forall h \in \mathcal{H}, \mathcal{R}_T(h) \leq \mathcal{R}_S(h) + d_{\mathcal{H}\Delta\mathcal{H}}(\mathcal{D}_S, \mathcal{D}_T) + \beta,
\end{equation}
where $\mathcal{R}_T(h)$ is the target domain error and $\mathcal{R}_S(h)$ is the source domain error. $d_{\mathcal{H}\Delta\mathcal{H}}$ is the discrepancy distance between $\mathcal{D}_S$ and $\mathcal{D}_T$ w.r.t. $\mathcal{H}$. Then, given the label functions $g_T$ and $g_S$ determined by the domain labels $\mathcal{Y}_T$ and $\mathcal{Y}_S$, the shared error $\beta$ is
\begin{equation}
    \beta = \arg \min_{h \in \mathcal{H}} \mathcal{R}_S(h^*, g_S) + \mathcal{R}_T(h^*, g_T),
\end{equation}
where $h^*$ is the ideal hypothesis.
\end{theorem}

\paragraph{Homogeneous Domain Adaptation.} DA models aim to find a minimal discrepancy distance $d_{\mathcal{H}\Delta\mathcal{H}}(\mathcal{D}_S, \mathcal{D}_T)$. For classifying time-series data, we restrict our methods to \textit{homogeneous} DA, where the feature space is the same ($\mathbf{U}_T = \mathbf{U}_S$) with the same feature dimensionality ($q_T = q_S$ and $w_T = w_S$) and interpolate the time-series to a pre-defined fixed length. In the following, we provide details on computing the optimal transformation.

\paragraph{Optimal Transport.} To mitigate the domain shift for our multivariate time-series application, we assume that the domain drift is due to an unknown, possibly nonlinear map of the input space $\mathbf{T}: \mathcal{D}_S \xrightarrow{} \mathcal{D}_T$ that preserves the conditional distribution $P_S\big(y|f(\mathbf{U}_S)\big) = P_T\big(y|\mathbf{T}(f(\mathbf{U}_S))\big)$ such that the label information is preserved \cite{courty_flamary}. Searching for $\mathbf{T}$ in the space of all possible transformations is intractable. Hence, $\mathbf{T}$ is chosen such that a transportation cost
\begin{equation}
    C(\mathbf{T}) = \int_{\mathcal{D}_S} c\big(f(\mathbf{U}), \mathbf{T}(f(\mathbf{U}))\big) d\mu(f(\mathbf{U})),
\end{equation}
is minimized, where $c: \mathcal{D}_T \times \mathcal{D}_S \xrightarrow{} \mathbb{R}^{+}$ is a distance function over the metric space $\mathcal{D}$ \cite{courty_flamary}. The optimal transportation problem is 
\begin{equation}
    \mathbf{T}_0 = \arg \min_{\mathbf{T}} = \int_{\mathcal{D}_S} c\big(f(\mathbf{U}), \mathbf{T}(f(\mathbf{U}))\big)d\mu(f(\mathbf{U})).
\end{equation}
This is also known as the Kantorovitch formulation \cite{kantorovitch} that allows to search a general coupling $\alpha \in \Theta$ by the \textit{transportation plan} \cite{santambrogio}:
\begin{equation}
    \alpha_0 = \arg \min_{\alpha \in \Theta} \int_{\mathcal{D}_T \times \mathcal{D}_S} c\big(f(\mathbf{U}_T), f(\mathbf{U}_S)\big)d\alpha\big(f(\mathbf{U}_T), f(\mathbf{U}_S)\big),
\end{equation}
where $\Theta$ is a set of all probabilistic couplings $\Theta \in P(\mathcal{D}_T \times \mathcal{D}_S)$ with marginals $\mu_T$ and $\mu_S$. Then, the \textit{Wasserstein distance} of order $p$ between $\mu_T$ and $\mu_S$ can be defined as
\begin{equation}
    \resizebox{.99\linewidth}{!}{$
    \displaystyle
    W_p(\mu_T, \mu_S) := \Big(\inf_{\alpha \in \Theta} \int_{\mathcal{D}_T \times \mathcal{D}_S} d\big(f(\mathbf{U}_T), f(\mathbf{U}_S)\big)^p d\alpha\big(f(\mathbf{U}_T), f(\mathbf{U}_S)\big)\Big)^{\frac{1}{p}},
    $}
\end{equation}
where $d$ is a distance metric \cite{courty_flamary} as the cost function:
\begin{equation}
    c\big(f(\mathbf{U}_T), f(\mathbf{U}_S)\big) = d\big(f(\mathbf{U}_T), f(\mathbf{U}_S)\big)^p.
\end{equation}
We apply the earth mover's distance (EMD), the EMD with Laplacian regularization \cite{flamary_courty}, and Sinkhorn transport \cite{altschuler, marco_cuturi} (with $L_{p}L_{1}$ and $L_{1}L_{2}$ class regularization of 0.5) implemented by the Python Optimal Transport (POT) package \cite{pot_flamary_courty} between our source and target domain samples. Selecting a proper cost function is crucial for the effectiveness of the adaptation of source to target domain. Typically, the $L_2$-based metric is used, but other cost functions are also possible, e.g., norm-based metrics \cite{villani}, metrics based on Riemannian distances over a manifold \cite{villani}, metrics used as a loss function \cite{cuturi_avis}, or concave cost functions \cite{ferradans}. We evaluate 18 different distance metrics in Section~\ref{chap_eval_da}. For comparison, we use correlation alignment \cite{sun_feng} (for more information, see Appendix~\ref{sec_app_coral}). 

\subsubsection{Transformation Selection for Inference}
\label{chap_trans_selection}

At inference we aim to select the best class-specific transformation (while the class label of the sample is not known). Hence, we compute the transformation of the validation source domain embeddings for each class, and select the transformation for which the embeddings of the target $f(\mathbf{U}_T)$ and transformed source $\mathbf{T}(f(\mathbf{U}_S))$ domains have the smallest cost $c\big(\mathbf{T}(f(\mathbf{U}_S)), f(\mathbf{U}_T)\big)$. We use common similarity metrics such as the cross correlation (CC) $d_{\text{CC}}$ and Pearson correlation (PC)~\cite{pearson} $d_{\text{PC}}$. We also compute the MMD~\cite{long_cao} by
\begin{equation}
    d_{\text{MMD}}(\mathcal{D}_T^C, \mathcal{D}_S^C) = \Bigg |\Bigg |\frac{1}{\mathcal{N}_T} \sum_{i=1}^{\mathcal{N}_T}f\big(\mathbf{U}_T^{C,i}\big) - \frac{1}{\mathcal{N}_S} \sum_{j=1}^{\mathcal{N}_S}\mathbf{T}\big(f(\mathbf{U}_S^{C,j}\big)\big)\Bigg |\Bigg |_{\mathcal{H}}^2,
\end{equation}
for class $C$ and the RKHS~$\mathcal{H}$. We compare to HoMM~\cite{chen_fu} of order 3 and kMMD~\cite{long_zhu_mmd} (which is equivalent to the kernelized HoMM of order 1). We also make use of different CORAL metrics (standard, Stein~\cite{cherian_sra} and Jeff~\cite{moakher_batchelor} CORAL based on symmetrized Bregman divergences \cite{harandi_salzmann}) for embedding comparisons.

\section{Experimental Setup}
\label{chap_experiments}

We apply our DA technique to two time-series datasets: generated time-series (Section~\ref{chap_exp_synth}), and OnHW recognition (Section~\ref{chap_exp_onhw}).

\subsection{Synthetic Time-Series Classification}
\label{chap_exp_synth}

\begin{figure}[t!]
	\centering
	\begin{minipage}[b]{0.32\linewidth}
        \centering
    	\includegraphics[trim=14 13 10 11, clip, width=1.0\linewidth]{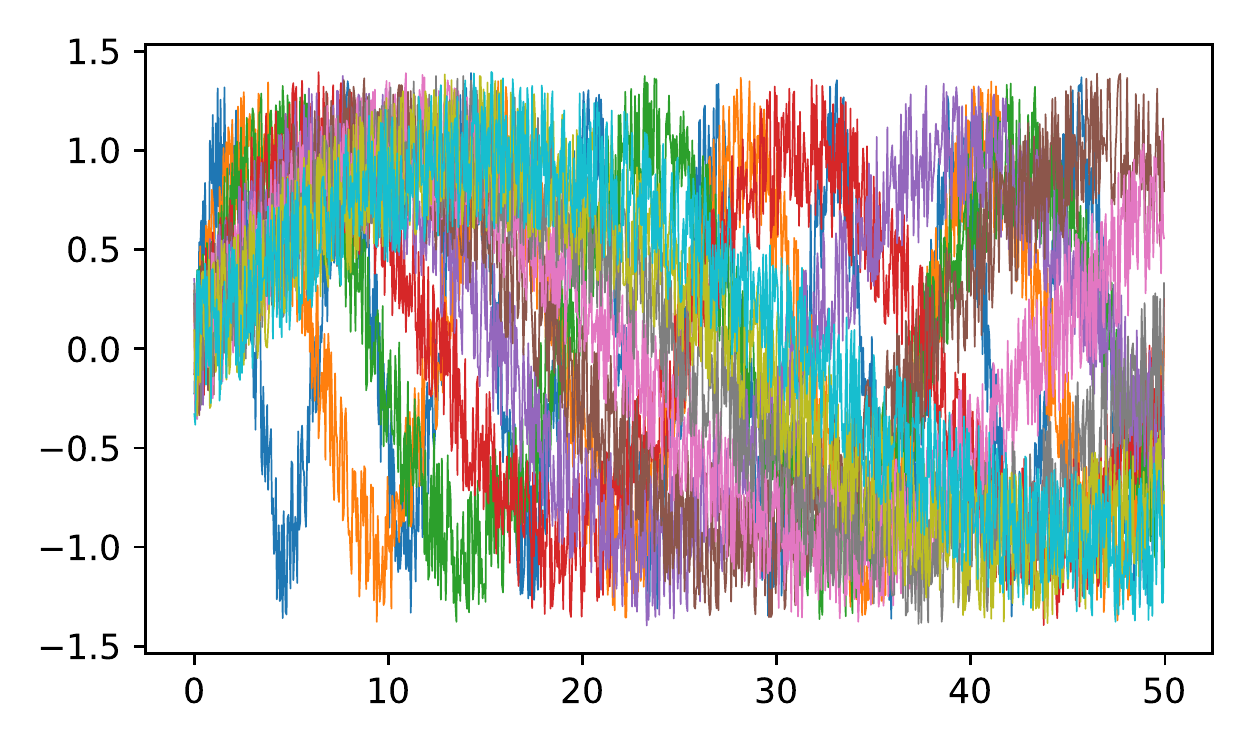}
    	\subcaption{Target domain (training) data $\mathcal{U}_{T_t}$ with noise $b = 0.4$.}
    	\label{image_sinus_a}
    \end{minipage}
    \hfill
	\begin{minipage}[b]{0.32\linewidth}
        \centering
    	\includegraphics[trim=14 13 10 11, clip, width=1.0\linewidth]{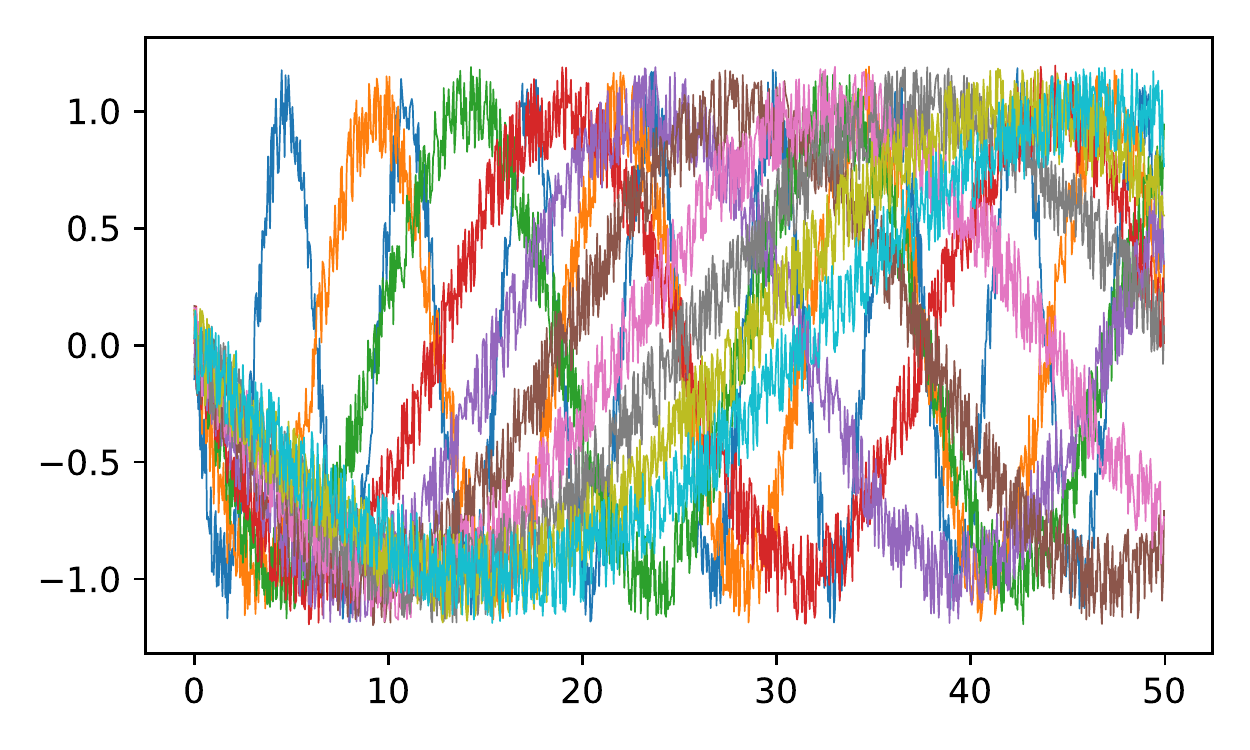}
    	\subcaption{Source domain (training) data $\mathcal{U}_{S_t}$ with noise $b = 0.2$.}
    	\label{image_sinus_b}
    \end{minipage}
    \hfill
	\begin{minipage}[b]{0.32\linewidth}
        \centering
    	\includegraphics[trim=14 13 10 11, clip, width=1.0\linewidth]{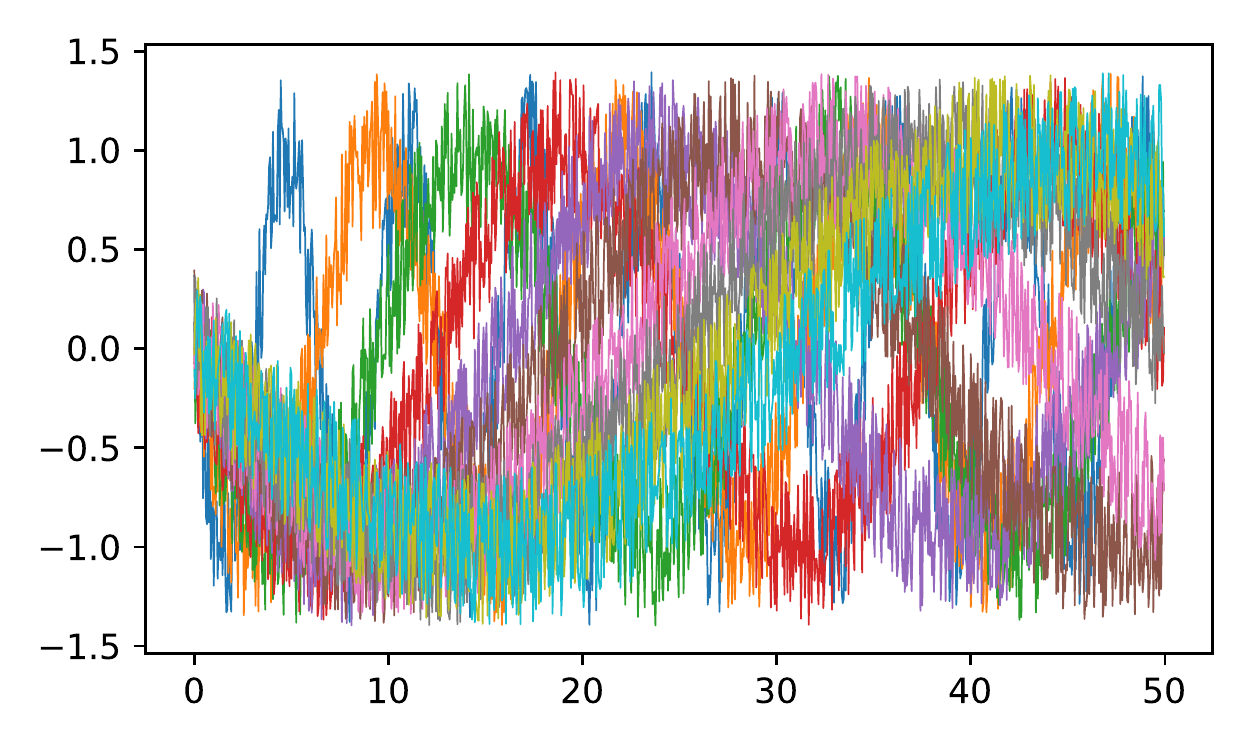}
    	\subcaption{Source domain (validation) data $\mathcal{U}_{S_v}$ with noise $b = 0.4$.}
    	\label{image_sinus_c}
    \end{minipage}
    \vspace{-0.1cm}
    \caption{Time-series of the generated sinusoidal dataset.}
    \label{image_sinus_cos}
\end{figure}

We first investigate the efficacy of our method to transform synthetically generated univariate time-series data. For this, we generate sinusoidal signal data of 200 time steps with different frequencies for 10 classes. We add noise from a continuous uniform distribution $U(a,b)$ with $a=0.0$ and $b \in B = \{0.0, 0.1, 0.2, \ldots, 1.9\}$ for the target domain dataset (Figure~\ref{image_sinus_a}), and train a CNN+BiLSTM. For the source domain (training) dataset, we flip the sign of the generated time-series and add uniform $U(a,b/2)$ noise (Figure~\ref{image_sinus_b}). We then validate the adapted model using the flipped dataset with added uniform noise for $b\in B$ values (Figure~\ref{image_sinus_c}). This allows us to evaluate the time-series adaptation for different noise ratios. 

\subsection{Online Handwriting (OnHW) Recognition}
\label{chap_exp_onhw}

\begin{figure}[t!]
	\centering
	\begin{minipage}[b]{0.487\linewidth}
        \centering
    	\includegraphics[trim=14 14 9 9, clip, width=1.0\linewidth]{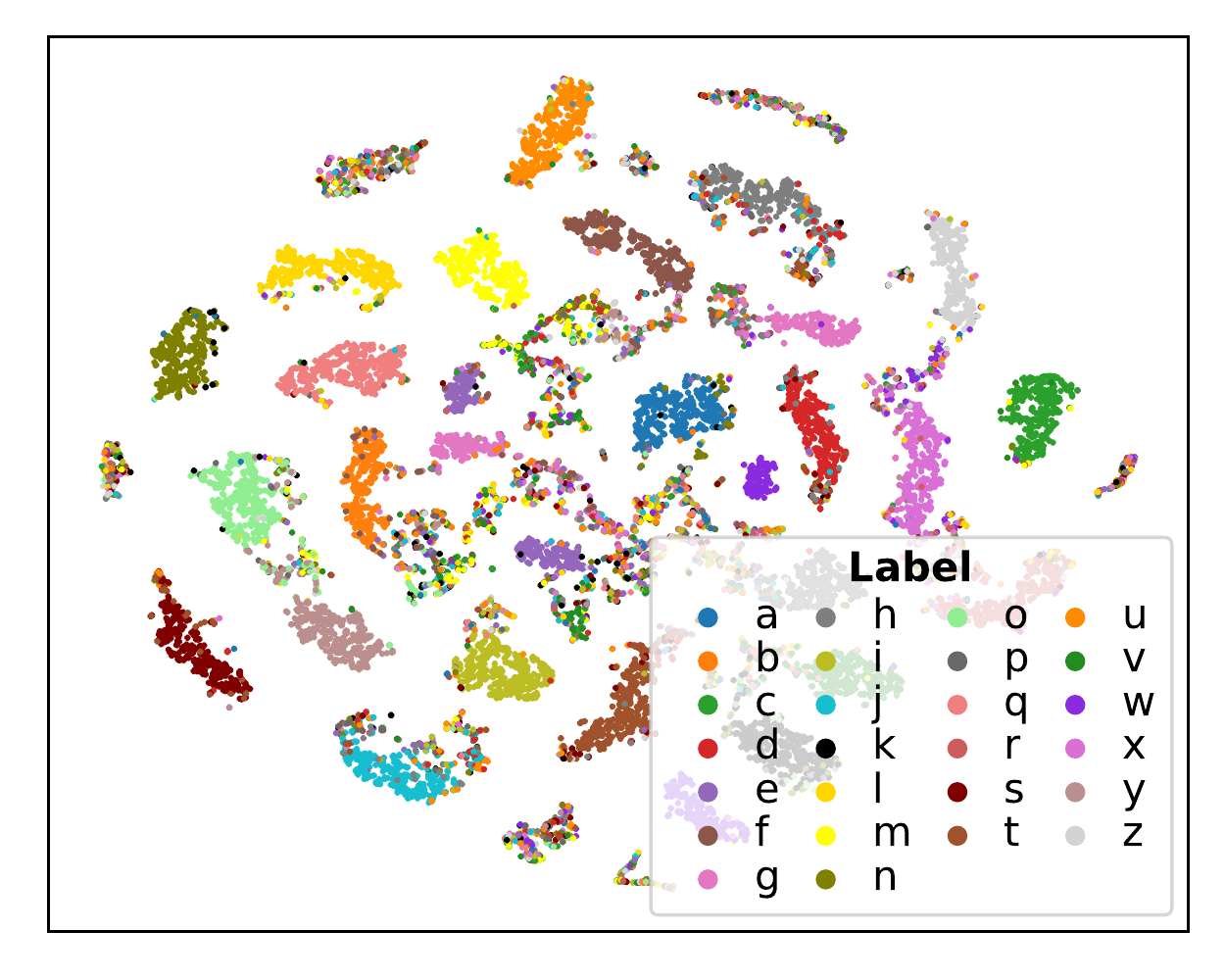}
    	\subcaption{Visualization of the 26 lower letters of the right-handed target domain OnHW-chars dataset.}
    	\label{image_emb_1}
    \end{minipage}
    \hfill
	\begin{minipage}[b]{0.487\linewidth}
        \centering
    	\includegraphics[trim=14 14 9 9, clip, width=1.0\linewidth]{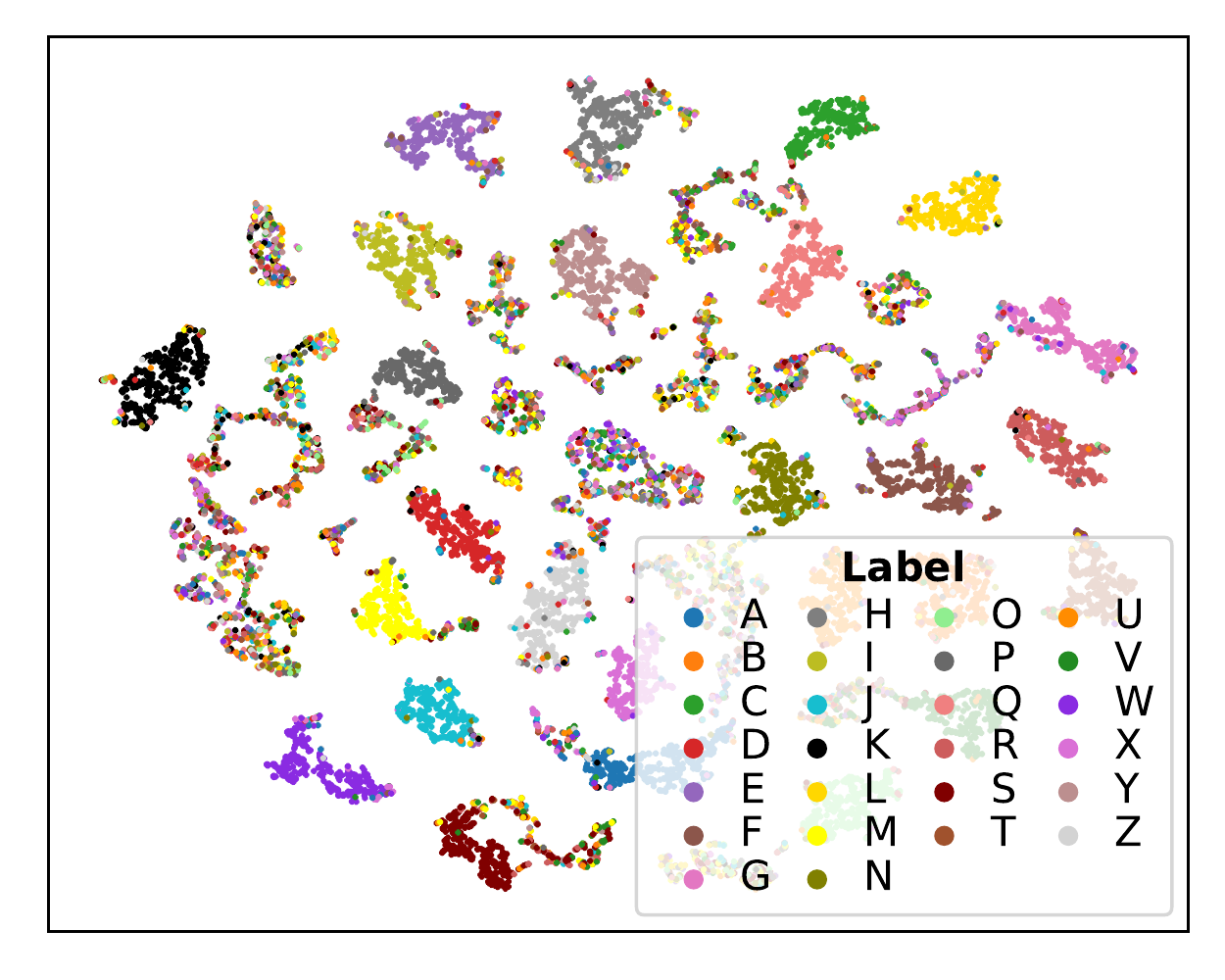}
    	\subcaption{Visualization of the 26 upper letters of the right-handed target domain OnHW-chars dataset.}
    	\label{image_emb_2}
    \end{minipage}
    \hfill
	\begin{minipage}[b]{0.487\linewidth}
        \centering
    	\includegraphics[trim=14 14 9 3, clip, width=1.0\linewidth]{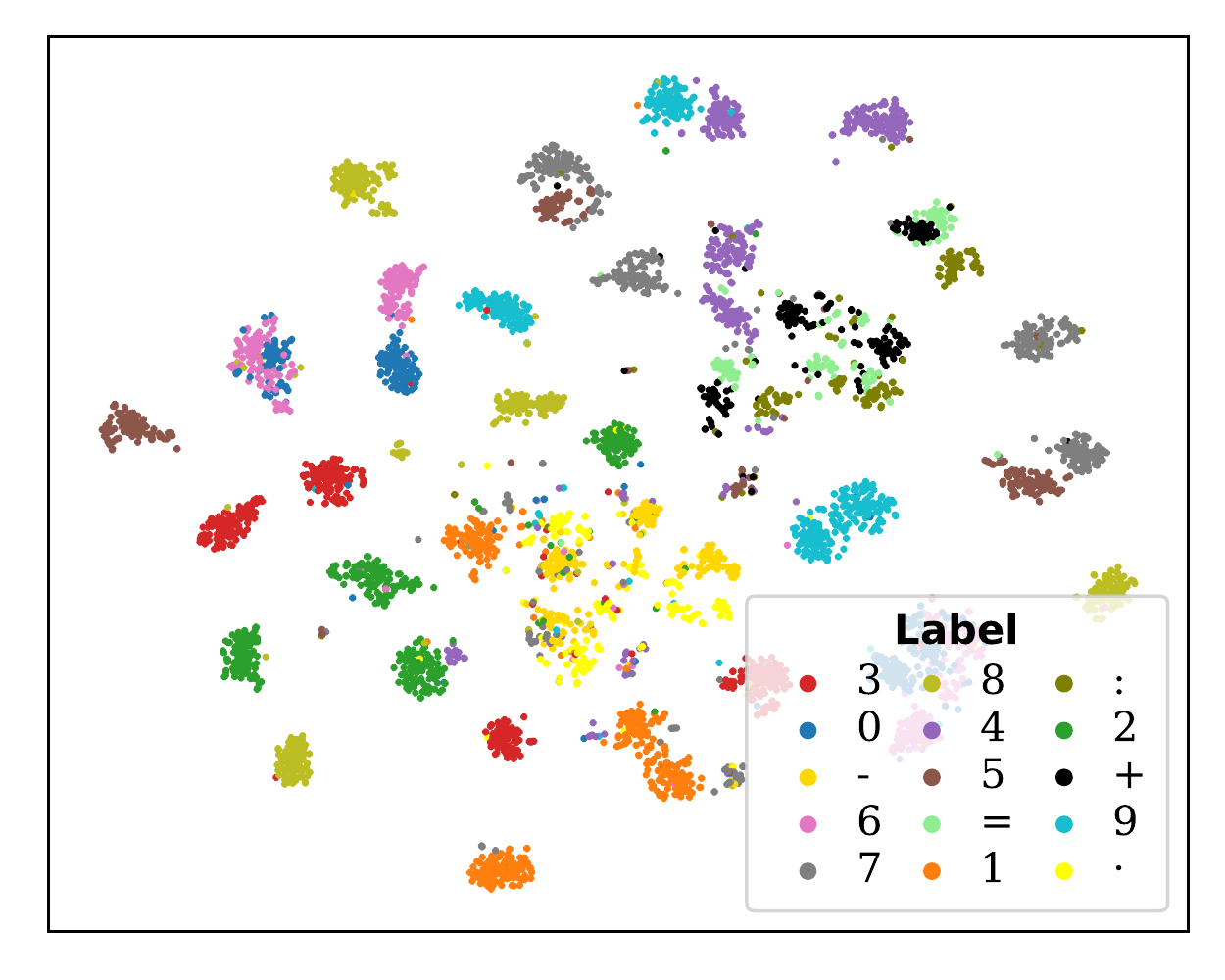}
    	\subcaption{Visualization of the 15 classes of right-handed target domain split OnHW-equations.}
    	\label{image_emb_3}
    \end{minipage}
    \hfill
	\begin{minipage}[b]{0.487\linewidth}
        \centering
    	\includegraphics[trim=14 14 9 3, clip, width=1.0\linewidth]{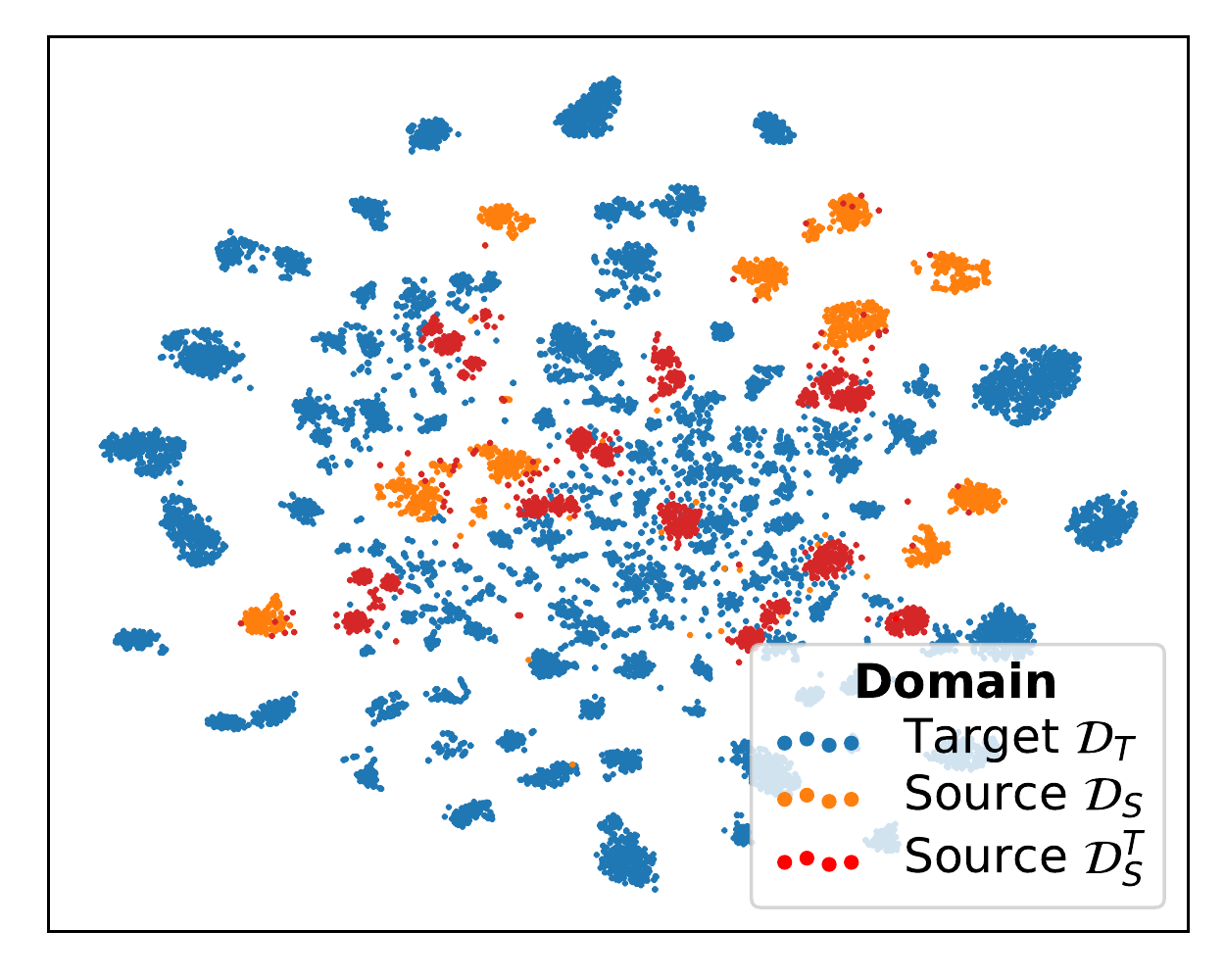}
    	\subcaption{Visualization of the target $\mathcal{U}_T$, source $\mathcal{U}_S$ and transformed source $\mathcal{U}_S^T$ domain features.}
    	\label{image_emb_4}
    \end{minipage}
    \vspace{-0.2cm}
    \caption{Embedding $f(\mathbf{U})$ visualization of OnHW-chars (a and b) and split OnHW-equations (c and d) using t-SNE.}
    \label{image_emb_figure}
    \vspace{-0.2cm}
\end{figure}

\begin{table*}[t!]
\begin{center}
\setlength{\tabcolsep}{3.7pt}
    \caption{Evaluation results (CRR in \%, mean and standard deviation) for transfer learning techniques on the OnHW-symbols and split OnHW-equations datasets \cite{ott_tist} (averaged over four left-handed writers) and on the OnHW-chars \cite{ott} dataset (averaged over nine left-handed writers) based on the CNN+BiLSTM architecture.}
    \label{table_post_training}
    \vspace{-0.1cm}
    \small \begin{tabular}{ p{0.5cm} | p{0.5cm} | p{0.5cm} | p{0.5cm} | p{0.5cm} | p{0.5cm} }
    & \multicolumn{1}{c|}{\textbf{OnHW-}} & \multicolumn{1}{c|}{\textbf{split OnHW-}} & \multicolumn{3}{c}{\textbf{OnHW-chars \cite{ott}}} \\
    \multicolumn{1}{c|}{\textbf{Method}} & \multicolumn{1}{c|}{\textbf{symbols \cite{ott_tist}}} & \multicolumn{1}{c|}{\textbf{equations \cite{ott_tist}}} & \multicolumn{1}{c|}{\textbf{lower}} & \multicolumn{1}{c|}{\textbf{upper}} & \multicolumn{1}{c}{\textbf{combined}} \\ \hline
    \multicolumn{1}{l|}{\textbf{Baseline ($\mathcal{U}_{S_v}$ in Target Model)}} & \multicolumn{1}{l|}{\:\: 19.18} & \multicolumn{1}{l|}{\:\:\:\:\: 33.52 } & \multicolumn{1}{l|}{\textbf{45.80}} & \multicolumn{1}{l|}{\textbf{45.97}} & \multicolumn{1}{l}{\: 25.19} \\
    \multicolumn{1}{l|}{\textbf{Without Transformation}} & \multicolumn{1}{r|}{\textbf{36.92} $\pm$ \, \footnotesize 8.71} & \multicolumn{1}{r|}{47.06 $\pm$ \footnotesize 27.13} & \multicolumn{1}{r|}{3.60 $\pm$ \, \footnotesize 2.35} & \multicolumn{1}{r|}{3.87 $\pm$ \, \footnotesize 5.80} & \multicolumn{1}{r}{3.03 $\pm$ \, \footnotesize 2.44} \\
    \multicolumn{1}{l|}{\textbf{Post-Training (full)}} & \multicolumn{1}{r|}{27.49 $\pm$ \footnotesize 14.29} & \multicolumn{1}{r|}{\textbf{90.53} $\pm$ \, \footnotesize 6.12} & \multicolumn{1}{r|}{35.85 $\pm$ \footnotesize 30.33} & \multicolumn{1}{r|}{31.15 $\pm$ \footnotesize 30.20} & \multicolumn{1}{r}{\textbf{50.22} $\pm$ \footnotesize 19.74} \\
    \multicolumn{1}{l|}{\textbf{Post-Training (middle)}} & \multicolumn{1}{r|}{19.23 $\pm$ \footnotesize 16.00} & \multicolumn{1}{r|}{88.00 $\pm$ \, \footnotesize 6.51} & \multicolumn{1}{r|}{28.39 $\pm$ \footnotesize 23.54} & \multicolumn{1}{r|}{21.42 $\pm$ \footnotesize 23.92} & \multicolumn{1}{r}{41.76 $\pm$ \footnotesize 18.81} \\
    \multicolumn{1}{l|}{\textbf{Post-Training (last)}} & \multicolumn{1}{r|}{22.15 $\pm$ \footnotesize 20.47} & \multicolumn{1}{r|}{78.87 $\pm$ \footnotesize 10.89} & \multicolumn{1}{r|}{16.17 $\pm$ \footnotesize 14.31} & \multicolumn{1}{r|}{10.75 $\pm$ \footnotesize 14.99} & \multicolumn{1}{r}{19.88 $\pm$ \footnotesize 14.20} \\
    \multicolumn{1}{l|}{\textbf{Layer Adapting}} & \multicolumn{1}{r|}{6.87 $\pm$ \, \footnotesize 2.45} & \multicolumn{1}{r|}{21.53 $\pm$ \, \footnotesize 4.60} & \multicolumn{1}{r|}{2.67 $\pm$ \, \footnotesize 2.04} & \multicolumn{1}{r|}{4.87 $\pm$ \, \footnotesize 2.87} & \multicolumn{1}{r}{2.10 $\pm$ \, \footnotesize 2.56} \\
    \end{tabular}
\end{center}
\end{table*}

OnHW recognition typically uses time in association with different types of spatio-temporal signals. The data contains information about the displacement of certain input devices \cite{plamondon_srihari}. OnHW recognition from sensor-enhanced pens uses data from inertial measurement units to capture the pen movement. The pen in \cite{ott} uses two accelerometers, one gyroscope, one magnetometer, 3 axes each, and one force sensor at 100\,Hz. One sample of size $m \times l$ represents an MTS of $m$ time steps from $l=13$ sensor channels. We make use of three character-based datasets: The \textit{OnHW-chars} \cite{ott} dataset contains 31,275 samples of small and capital characters (52 label classes) from 119 right-handed writers. The \textit{OnHW-symbols} \cite{ott_tist} dataset contains 2,326 samples of numbers and symbols (15 label classes) from 27 right-handed writers, and the \textit{split OnHW-equations} \cite{ott_tist} dataset contains 39,643 numbers and symbols from 55 right-handed writers obtained from 10,713 equations. For these datasets, 80/20 train/validation splits are available for writer-dependent (WD) and writer-independent (WI) classification tasks. We define these datasets as our target domain $\mathcal{U}_T$. Usually, observations from left-handed writers in OnHW datasets is scarce \cite{klass_lorenz}. As we want OnHW recognition to work equally well for left-handed writers as for right-handed writers, we use the smaller available left-handed datasets as source domain $\mathcal{U}_S$ and split it into an adaptation (training) set $\mathcal{U}_{S_t}$ and a validation set $\mathcal{U}_{S_v}$. For an overview, see Appendix~\ref{sec_app_datasets}.

Figure~\ref{image_emb_figure} visualizes the $19 \times 200$ dimensional feature embeddings of the CNN+BiLSTM model for the OnHW-chars and split OnHW-equations datasets. We use the t-SNE method \cite{maaten} with an initial dimension 3,800, perplexity of 30, an initial momentum of 0.5, and a final momentum of 0.8. Figure~\ref{image_emb_1} and \ref{image_emb_2} visualizes 26 lower and 26 uppercase class labels of OnHW-chars. Several samples are in clusters associated with a different class and hence wrongly classified. Figure~\ref{image_emb_3} differentiates between all 15 class labels for the split OnHW-equations dataset. Here, we can clearly see that the labels \texttt{'+'}, \texttt{'='} and \texttt{':'} are close in their low-dimensional embedding. Figure~\ref{image_emb_4} shows the low-dimensional embedding of the right-handed target domain $\mathcal{D}_T$ (blue) and the left-handed source domain $\mathcal{D}_{S}$ (orange). It is notable that the features of both domains have different distributions. After the transformation, the left-handed source domain $\mathcal{D}_{S}^T$ (red) is closer to the target domain.

\section{Experimental Results}
\label{chap_results}

\begin{figure}[t!]
	\centering
    \includegraphics[width=1.0\linewidth]{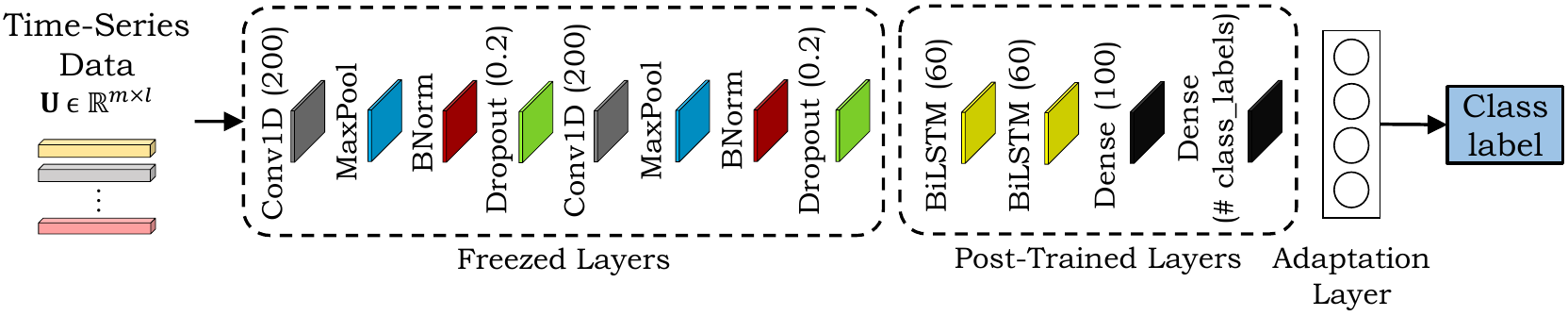}
    \vspace{-0.4cm}
    \caption{Network architecture with transfer learning by freezing the first layers and fine-tuning the last layers, or only post-training an additional adaptation layer. The network is pre-trained on the target domain $\mathcal{U}_{T_t}$, and adapted on the source domain training set $\mathcal{U}_{S_t}$ and validated on the set $\mathcal{U}_{S_v}$.}
    \label{image_post_training_methods}
\end{figure}

\paragraph{Hardware and Training Setup.} For all experiments we use Nvidia Tesla V100-SXM2 GPUs with 32 GB VRAM equipped with Core Xeon CPUs and 192 GB RAM. We use the vanilla Adam optimizer with a learning rate of $10^{-4}$. We use the character recognition rate (CRR) in \% as time-series classification evaluation metric. Details on the architecture, are proposed in Appendix~\ref{sec_app_architecture}.

\subsection{Limitations of Transfer Learning (TL)}
\label{chap_lim_transfer_learn}

Fine-tuning an existing model on each source domain can be data- and parameter-inefficient. Under the assumption that the adaptation dataset is large enough, fine-tuning leads to a better performance than adapting the domains by feature-based transfer as the model can overfit on the source data \cite{howard_ruder}. Hence, we apply different TL techniques and compare to DA techniques. Figure~\ref{image_post_training_methods} shows our network architecture and different TL techniques. First, we adapt the whole model by fine-tuning on each source domain dataset from the pre-trained network on the target domain dataset. Second, we freeze the first layers and only post-train the spatio-temporal layers (two BiLSTMs and two dense layers). This layer freezing leads to a faster training than full post-training. Third, we apply a structurally similar method to the Adapter by Houlsby et al.~\cite{houlsby}: We freeze all previously pre-trained layers, add an adaptation layer (a standard \textit{dense} layer) at the end, and only train the additional layer. The \textit{dense} layer has $|\mathcal{Y}|$ units. This yields a compact and fast trainable model by adding only a few trainable parameters per task.

\begin{figure}[t!]
	\centering
    \includegraphics[trim=12 13 10 11, clip, width=0.6\linewidth]{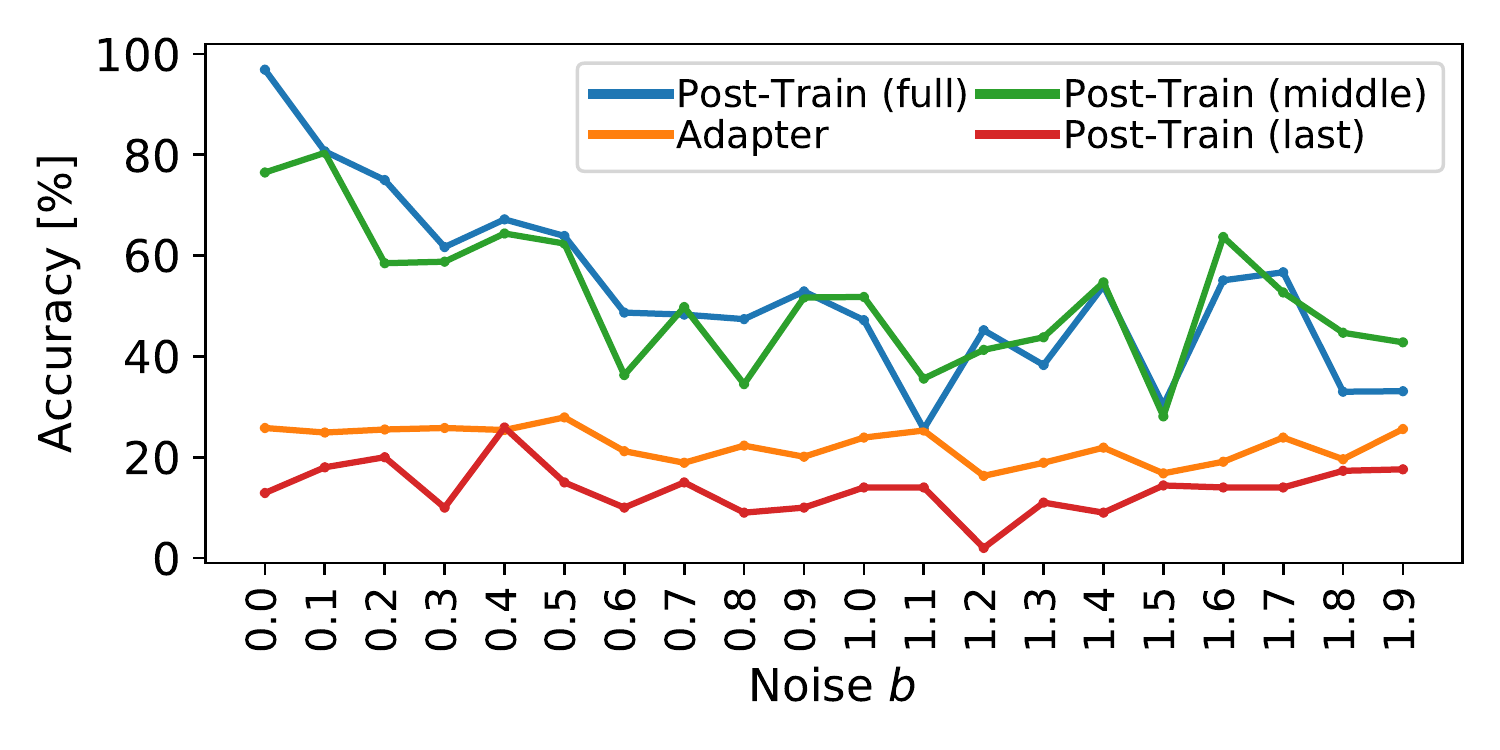}
    \vspace{-0.1cm}
    \caption{Evaluation of transfer learning on the generated dataset for noise parameters between $b=0.0$ and $b=1.9$.}
    \label{image_sinus_post_train}
\end{figure}

\begin{figure*}[t!]
	\begin{minipage}[b]{0.196\linewidth}
        \centering
    	\includegraphics[trim=12 13 9 11, clip, width=1.0\linewidth]{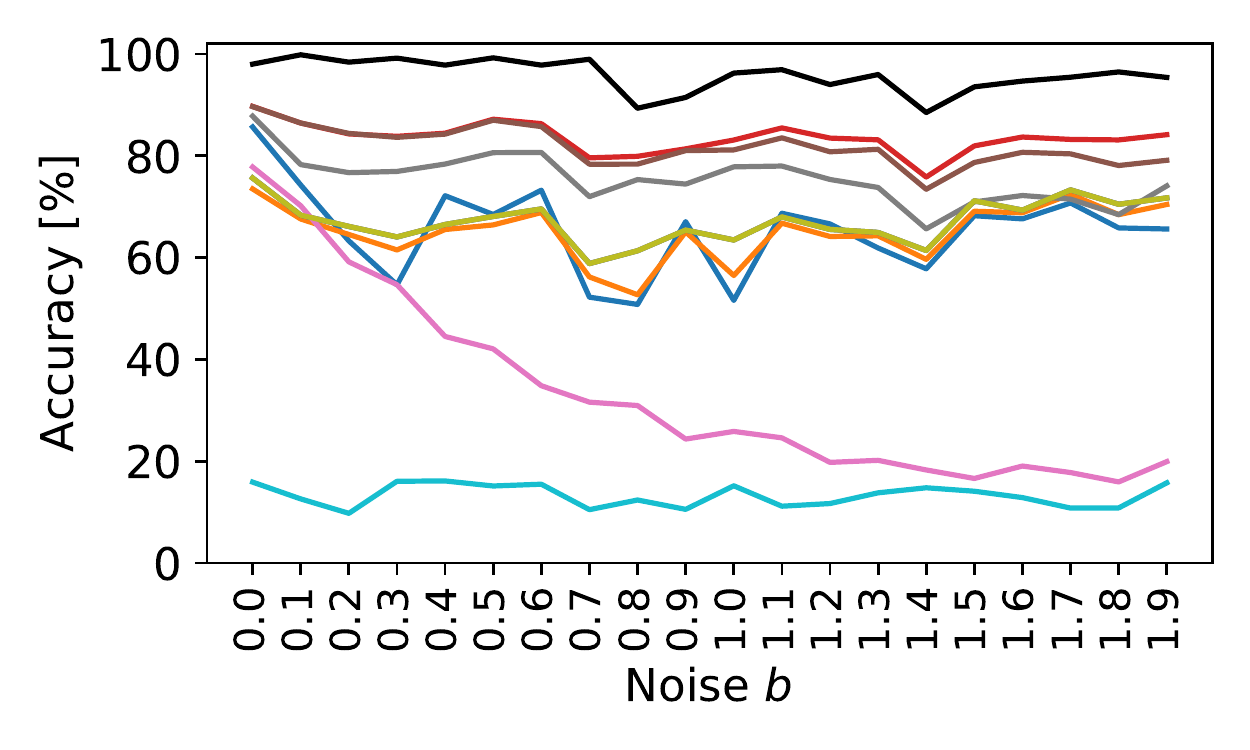}
    	\subcaption{EMD.}
    	\label{image_eval_sinus_a}
    \end{minipage}
    \hfill
	\begin{minipage}[b]{0.196\linewidth}
        \centering
    	\includegraphics[trim=12 13 9 11, clip, width=1.0\linewidth]{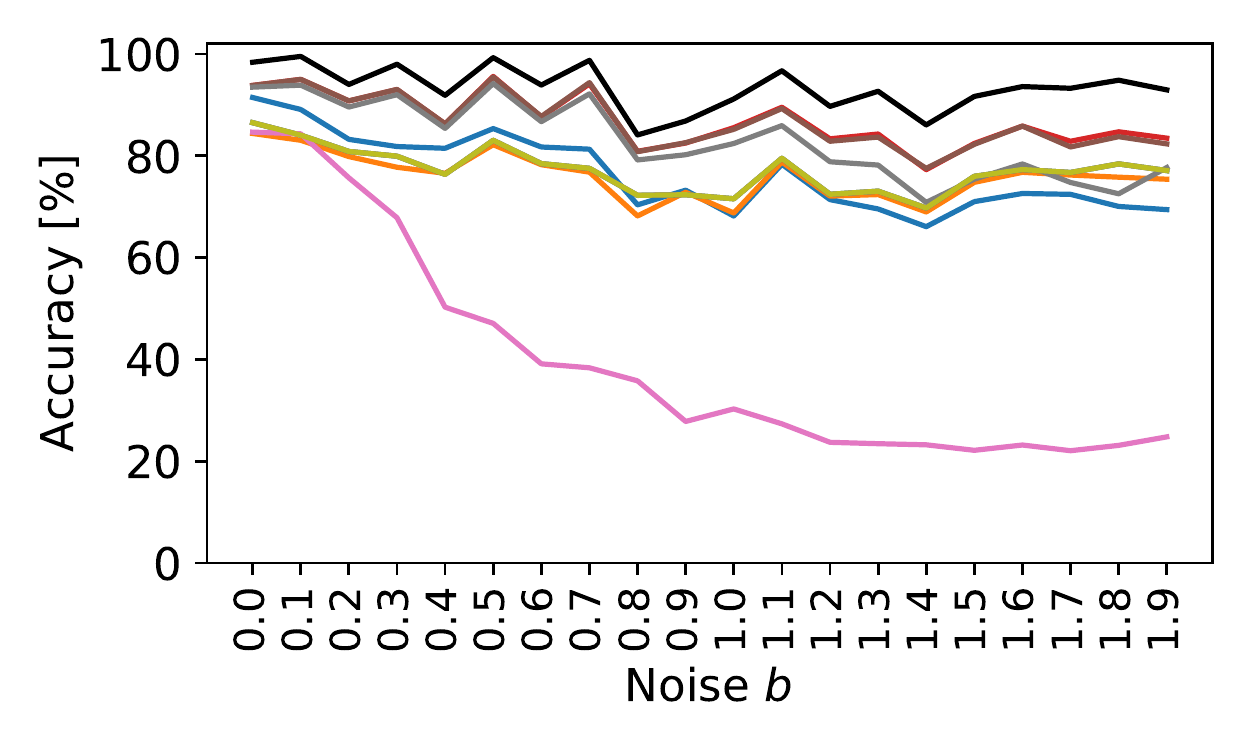}
    	\subcaption{SEMD.}
    	\label{image_eval_sinus_b}
    \end{minipage}
    \hfill
	\begin{minipage}[b]{0.196\linewidth}
        \centering
    	\includegraphics[trim=12 13 9 11, clip, width=1.0\linewidth]{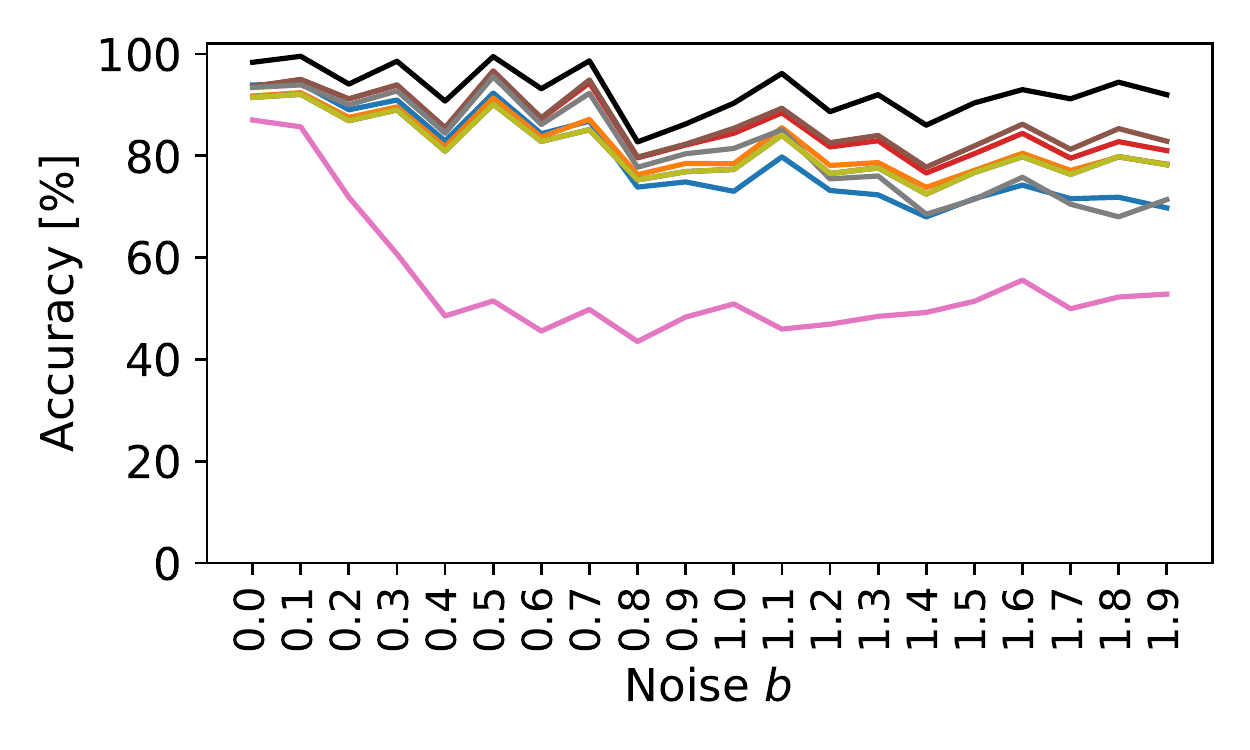}
    	\subcaption{Sinkhorn.}
    	\label{image_eval_sinus_c}
    \end{minipage}
    \hfill
	\begin{minipage}[b]{0.196\linewidth}
        \centering
    	\includegraphics[trim=12 13 9 11, clip, width=1.0\linewidth]{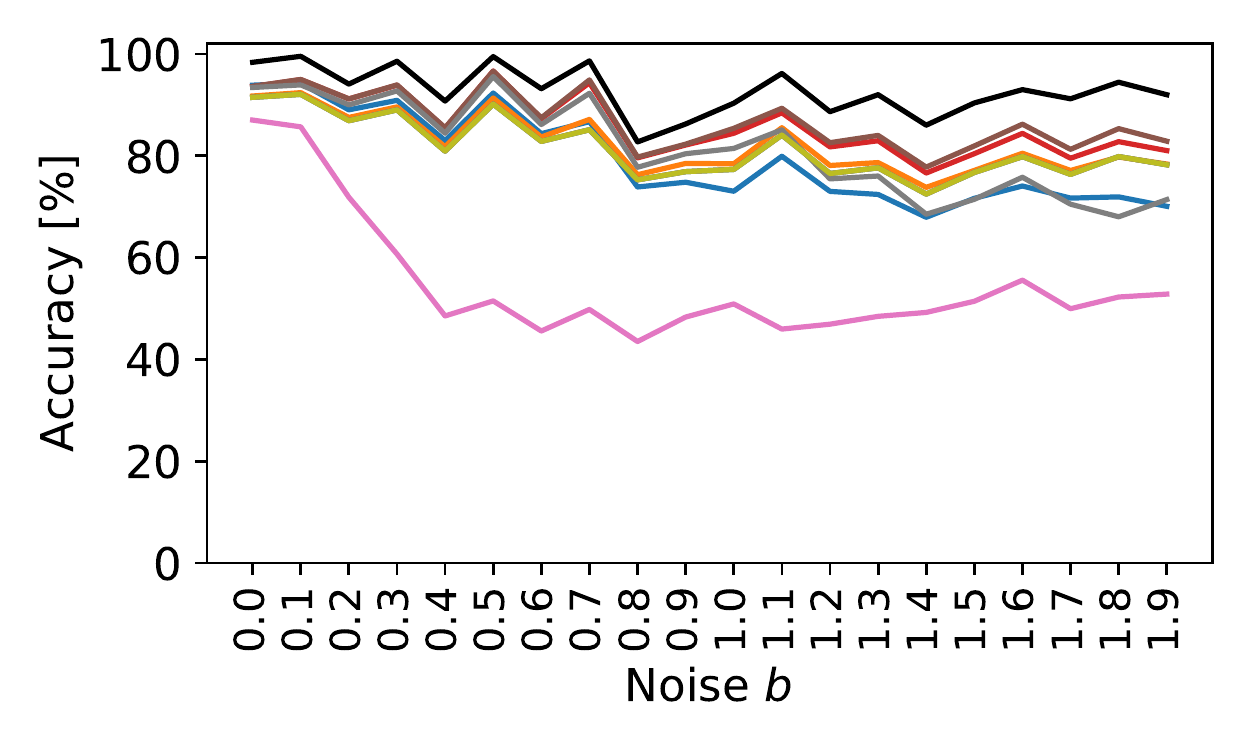}
    	\subcaption{Sinkhorn ($L_{p}L_{1}$).}
    	\label{image_eval_sinus_d}
    \end{minipage}
    \hfill
	\begin{minipage}[b]{0.196\linewidth}
        \centering
    	\includegraphics[trim=12 13 9 11, clip, width=1.0\linewidth]{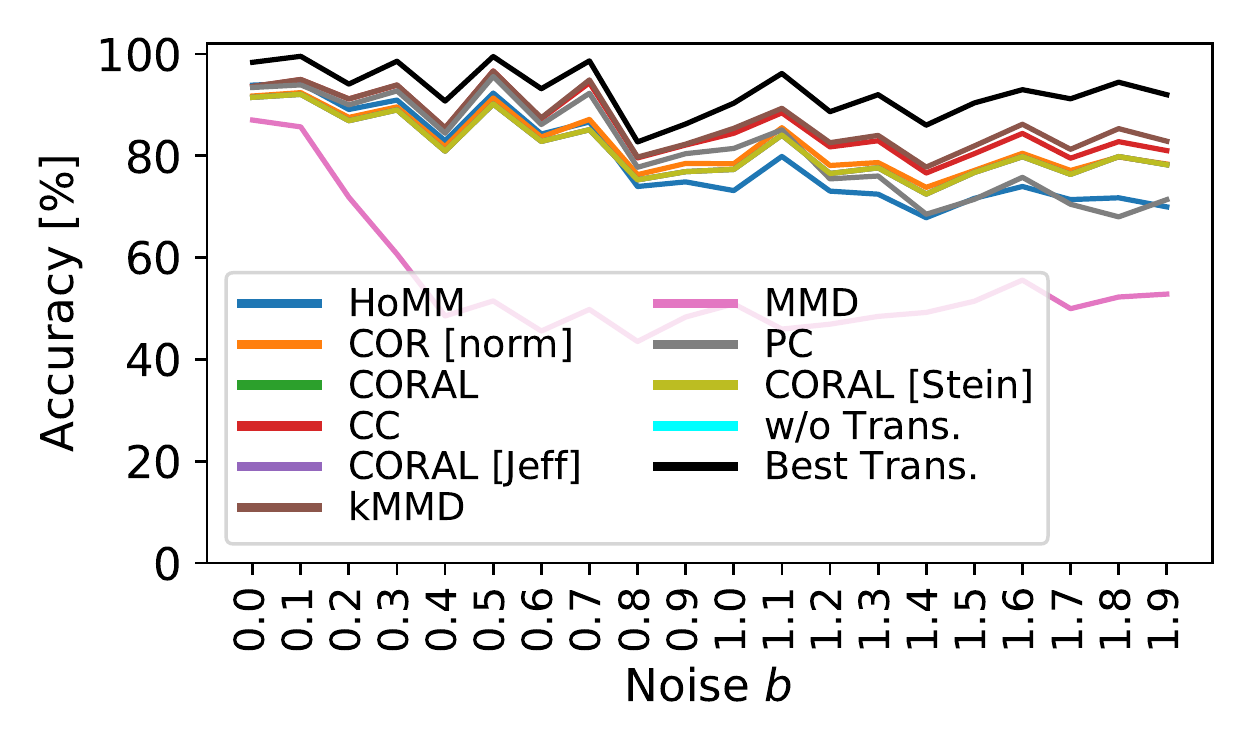}
    	\subcaption{Sinkhorn ($L_{1}L_{2}$).}
    	\label{image_eval_sinus_e}
    \end{minipage}
    \vspace{-0.3cm}
    \caption{Results for optimal transport methods and transformation selection on the synthetic dataset (for $b$ between $0.0$ and $1.9$). For better readability we depict the approach without transformation only in (a) as it is independent of optimal transport.}
    \label{image_eval_sinus}
\end{figure*}

\begin{table*}[t!]
\begin{center}
\setlength{\tabcolsep}{3.8pt}
    \caption{Evaluation results (CRR in \%) for the left- and right-handed writer OnHW-symbols and split OnHW-equations datasets \cite{ott_tist} for different MTS classification techniques. $\mathcal{U}_{T_v}$ are right-handed writer datasets, and $\mathcal{U}_{S_v}$ are left-handed writer datasets.}
    \label{table_baseline_symbols}
    \vspace{-0.1cm}
    \small \begin{tabular}{ p{0.5cm} | p{0.5cm} | p{0.5cm} | p{0.5cm} | p{0.5cm} | p{0.5cm} | p{0.5cm} | p{0.5cm} | p{0.5cm} | p{0.5cm} | p{0.5cm} }
    & \multicolumn{5}{c|}{\textbf{OnHW-symbols \cite{ott_tist}}} & \multicolumn{5}{c}{\textbf{split OnHW-equations \cite{ott_tist}}} \\
    & \multicolumn{2}{c|}{\textbf{Right-handed}} & \multicolumn{2}{c|}{\textbf{Left-handed}} & \multicolumn{1}{c|}{\textbf{L in R}} & \multicolumn{2}{c|}{\textbf{Right-handed}} & \multicolumn{2}{c|}{\textbf{Left-handed}} & \multicolumn{1}{c}{\textbf{L in R}} \\
    \multicolumn{1}{c|}{\textbf{Method}} & \multicolumn{1}{c|}{\textbf{WD}} & \multicolumn{1}{c|}{\textbf{WI}} & \multicolumn{1}{c|}{\textbf{WD}} & \multicolumn{1}{c|}{\textbf{WI}} & \multicolumn{1}{c|}{\textbf{WI}} & \multicolumn{1}{c|}{\textbf{WD}} & \multicolumn{1}{c|}{\textbf{WI}} & \multicolumn{1}{c|}{\textbf{WD}} & \multicolumn{1}{c|}{\textbf{WI}} & \multicolumn{1}{c}{\textbf{WI}} \\ \hline
    \multicolumn{1}{l|}{\textbf{CNN+BiLSTM \cite{ott_tist}}} & \multicolumn{1}{r|}{\textbf{96.20}} & \multicolumn{1}{r|}{\textbf{79.51}} & \multicolumn{1}{r|}{\textbf{92.00}} & \multicolumn{1}{r|}{\textbf{54.00}} & \multicolumn{1}{c|}{\textbf{19.18}} & \multicolumn{1}{r|}{\textbf{95.70}} & \multicolumn{1}{r|}{83.88} & \multicolumn{1}{r|}{\textbf{92.00}} & \multicolumn{1}{r|}{\textbf{51.50}} & \multicolumn{1}{c}{\textbf{33.52}} \\
    \multicolumn{1}{l|}{\textbf{LSTM-FCN \cite{karim_majumdar}}} & \multicolumn{1}{r|}{92.39} & \multicolumn{1}{r|}{73.32} & \multicolumn{1}{r|}{75.34} & \multicolumn{1}{r|}{41.40} & \multicolumn{1}{c|}{-} & \multicolumn{1}{r|}{93.95} & \multicolumn{1}{r|}{81.47} & \multicolumn{1}{r|}{88.56} & \multicolumn{1}{r|}{47.56} & \multicolumn{1}{c}{-} \\
    \multicolumn{1}{l|}{\textbf{ResCNN ($nf=64$) \cite{zou_wang}}} & \multicolumn{1}{r|}{92.23} & \multicolumn{1}{r|}{77.41} & \multicolumn{1}{r|}{80.82} & \multicolumn{1}{r|}{47.87} & \multicolumn{1}{c|}{-} & \multicolumn{1}{r|}{94.58} & \multicolumn{1}{r|}{80.95} & \multicolumn{1}{r|}{89.59} & \multicolumn{1}{r|}{40.45} & \multicolumn{1}{c}{-} \\
    \multicolumn{1}{l|}{\textbf{ResNet ($nf=64$) \cite{wang_yan_oates}}} & \multicolumn{1}{r|}{94.50} & \multicolumn{1}{r|}{77.41} & \multicolumn{1}{r|}{80.82} & \multicolumn{1}{r|}{47.87} & \multicolumn{1}{c|}{-} & \multicolumn{1}{r|}{94.68} & \multicolumn{1}{r|}{83.45} & \multicolumn{1}{r|}{89.20} & \multicolumn{1}{r|}{39.21} & \multicolumn{1}{c}{-} \\
    \multicolumn{1}{l|}{\textbf{XResNet50 \cite{he_zhang}}} & \multicolumn{1}{r|}{93.66} & \multicolumn{1}{r|}{74.47} & \multicolumn{1}{r|}{78.08} & \multicolumn{1}{r|}{47.87} & \multicolumn{1}{c|}{-} & \multicolumn{1}{r|}{94.63} & \multicolumn{1}{r|}{81.74} & \multicolumn{1}{r|}{89.67} & \multicolumn{1}{r|}{45.15} & \multicolumn{1}{c}{-} \\
    \multicolumn{1}{l|}{\textbf{XceptionTime ($nf=16$) \cite{rahimian_zabihi}}} & \multicolumn{1}{r|}{91.54} & \multicolumn{1}{r|}{72.34} & \multicolumn{1}{r|}{75.34} & \multicolumn{1}{r|}{40.43} & \multicolumn{1}{c|}{-} & \multicolumn{1}{r|}{94.03} & \multicolumn{1}{r|}{82.24} & \multicolumn{1}{r|}{88.72} & \multicolumn{1}{r|}{50.73} & \multicolumn{1}{c}{-} \\
    \multicolumn{1}{l|}{\textbf{InceptionTime ($nf=64$, $depth=12$) \cite{fawaz_lucas}}} & \multicolumn{1}{r|}{91.97} & \multicolumn{1}{r|}{76.92} & \multicolumn{1}{r|}{80.82} & \multicolumn{1}{r|}{46.81} & \multicolumn{1}{c|}{-} & \multicolumn{1}{r|}{94.87} & \multicolumn{1}{r|}{\textbf{84.35}} & \multicolumn{1}{r|}{88.48} & \multicolumn{1}{r|}{44.15} & \multicolumn{1}{c}{-} \\
    \multicolumn{1}{l|}{\textbf{TST \cite{zerveas_jayaraman}}} & \multicolumn{1}{r|}{91.12} & \multicolumn{1}{r|}{71.85} & \multicolumn{1}{r|}{78.08} & \multicolumn{1}{r|}{51.06} & \multicolumn{1}{c|}{-} & \multicolumn{1}{r|}{93.07} & \multicolumn{1}{r|}{80.40} & \multicolumn{1}{r|}{87.61} & \multicolumn{1}{r|}{47.27} & \multicolumn{1}{c}{-} \\
    \end{tabular}
\end{center}
\end{table*}

\paragraph{Sinusoidal Dataset Evaluation.} We adapt the pre-trained models (on the source domain datasets) with the target domain datasets by post-training the models provided in Figure~\ref{image_post_training_methods}. Figure~\ref{image_sinus_post_train} shows the results averaged over 10 trainings for all noise parameters $b\in B$. Post-training the full model and freezing the first layers while post-training the last layers yields the highest TL results. Post-training only the last layer or adapting an additional layer results in low classification accuracies below 20\%. In general, TL performs poorly when we see significant changes between the domains (i.e., $b>0.5$). In particular, all results are lower than the results achieved by our DA techniques (see Section~\ref{chap_eval_sin}, Figure~\ref{image_eval_sinus}).

\paragraph{OnHW Recognition Evaluation.} Table~\ref{table_post_training} shows TL results for all OnHW datasets. Without transformation, the models fail in the classification tasks, which proves the existence of a domain shift in the data \cite{klass_lorenz}. As the OnHW-symbols dataset is rather small, all TL techniques cannot adapt to a specific writer. The larger OnHW-chars dataset leads to better results. TL only yields good results on the split OnHW-equations dataset, while full model post-training outperforms training only specific layers or adapting an additional layer. Again, TL is limited in its efficacy while our DA approach shows promising results (see Section~\ref{chap_eval_da}).

\begin{table*}[t!]
\begin{center}
\setlength{\tabcolsep}{3.2pt}
    \caption{Evaluation results (CRR in \%) for the left-handed ($\mathcal{U}_{S_v}$) and right-handed ($\mathcal{U}_{T_v}$) writer OnHW-chars \cite{ott} datasets for different time-series classification techniques with same parameters as in Table~\ref{table_baseline_symbols}.}
    \label{table_baseline_chars}
    \vspace{-0.1cm}
    \small \begin{tabular}{ p{0.5cm} | p{0.5cm} | p{0.5cm} | p{0.5cm} | p{0.5cm} | p{0.5cm} | p{0.5cm} | p{0.5cm} | p{0.5cm} | p{0.5cm} | p{0.5cm} | p{0.5cm} | p{0.5cm} | p{0.5cm} | p{0.5cm} | p{0.5cm} }
    & \multicolumn{6}{c|}{\textbf{Right-handed ($\mathcal{U}_{T_v}$)}} & \multicolumn{6}{c|}{\textbf{Left-handed ($\mathcal{U}_{S_v}$)}} & \multicolumn{3}{c}{\textbf{ $\mathcal{U}_{S_v}$ in Target Model}} \\
    & \multicolumn{2}{c|}{\textbf{Lower}} & \multicolumn{2}{c|}{\textbf{Upper}} & \multicolumn{2}{c|}{\textbf{Combined}} & \multicolumn{2}{c|}{\textbf{Lower}} & \multicolumn{2}{c|}{\textbf{Upper}} & \multicolumn{2}{c|}{\textbf{Combined}} & \multicolumn{1}{c|}{\textbf{Lower}} & \multicolumn{1}{c|}{\textbf{Upper}} & \multicolumn{1}{c}{\textbf{Combined}} \\
    \multicolumn{1}{c|}{\textbf{Method}} & \multicolumn{1}{c|}{\textbf{WD}} & \multicolumn{1}{c|}{\textbf{WI}} & \multicolumn{1}{c|}{\textbf{WD}} & \multicolumn{1}{c|}{\textbf{WI}} & \multicolumn{1}{c|}{\textbf{WD}} & \multicolumn{1}{c|}{\textbf{WI}} & \multicolumn{1}{c|}{\textbf{WD}} & \multicolumn{1}{c|}{\textbf{WI}} & \multicolumn{1}{c|}{\textbf{WD}} & \multicolumn{1}{c|}{\textbf{WI}} & \multicolumn{1}{c|}{\textbf{WD}} & \multicolumn{1}{c|}{\textbf{WI}} & \multicolumn{1}{c|}{\textbf{WI}} & \multicolumn{1}{c|}{\textbf{WI}} & \multicolumn{1}{c}{\textbf{WI}} \\ \hline
    \multicolumn{1}{l|}{\textbf{CNN+BiLSTM \cite{ott_tist}}} & \multicolumn{1}{r|}{\textbf{88.85}} & \multicolumn{1}{r|}{\textbf{79.48}} & \multicolumn{1}{r|}{\textbf{92.15}} & \multicolumn{1}{r|}{\textbf{85.60}} & \multicolumn{1}{r|}{\textbf{78.17}} & \multicolumn{1}{r|}{\textbf{68.06}} & \multicolumn{1}{r|}{\textbf{94.70}} & \multicolumn{1}{c|}{\textbf{43.60}} & \multicolumn{1}{r|}{\textbf{91.90}} & \multicolumn{1}{c|}{\textbf{43.62}} & \multicolumn{1}{r|}{\textbf{82.80}} & \multicolumn{1}{c|}{\textbf{32.00}} & \multicolumn{1}{c|}{\textbf{45.80}} & \multicolumn{1}{c|}{\textbf{45.97}} & \multicolumn{1}{c}{\textbf{25.19}} \\
    \multicolumn{1}{l|}{\textbf{LSTM-FCN \cite{karim_majumdar}}} & \multicolumn{1}{r|}{81.43} & \multicolumn{1}{r|}{71.41} & \multicolumn{1}{r|}{85.43} & \multicolumn{1}{r|}{77.07} & \multicolumn{1}{r|}{67.34} & \multicolumn{1}{r|}{57.93} & \multicolumn{1}{r|}{70.55} & \multicolumn{1}{r|}{34.06} & \multicolumn{1}{r|}{72.50} & \multicolumn{1}{r|}{29.27} & \multicolumn{1}{r|}{61.02} & \multicolumn{1}{r|}{22.68} & \multicolumn{1}{c|}{-} & \multicolumn{1}{c|}{-} & \multicolumn{1}{c}{-} \\
    \multicolumn{1}{l|}{\textbf{ResCNN ($nf=64$) \cite{zou_wang}}} & \multicolumn{1}{r|}{82.52} & \multicolumn{1}{r|}{72.00} & \multicolumn{1}{r|}{86.91} & \multicolumn{1}{r|}{78.64} & \multicolumn{1}{r|}{67.55} & \multicolumn{1}{r|}{58.67} & \multicolumn{1}{r|}{80.00} & \multicolumn{1}{r|}{38.78} & \multicolumn{1}{r|}{80.63} & \multicolumn{1}{r|}{29.79} & \multicolumn{1}{r|}{65.39} & \multicolumn{1}{r|}{26.21} & \multicolumn{1}{c|}{-} & \multicolumn{1}{c|}{-} & \multicolumn{1}{c}{-} \\
    \multicolumn{1}{l|}{\textbf{ResNet ($nf=64$) \cite{wang_yan_oates}}} & \multicolumn{1}{r|}{83.01} & \multicolumn{1}{r|}{71.93} & \multicolumn{1}{r|}{86.41} & \multicolumn{1}{r|}{78.03} & \multicolumn{1}{r|}{68.56} & \multicolumn{1}{r|}{58.74} & \multicolumn{1}{r|}{81.01} & \multicolumn{1}{r|}{40.24} & \multicolumn{1}{r|}{82.95} & \multicolumn{1}{r|}{30.12} & \multicolumn{1}{r|}{66.95} & \multicolumn{1}{r|}{26.17} & \multicolumn{1}{c|}{-} & \multicolumn{1}{c|}{-} & \multicolumn{1}{c}{-} \\
    \multicolumn{1}{l|}{\textbf{XResNet50 \cite{he_zhang}}} & \multicolumn{1}{r|}{80.99} & \multicolumn{1}{r|}{69.14} & \multicolumn{1}{r|}{86.05} & \multicolumn{1}{r|}{76.69} & \multicolumn{1}{r|}{64.98} & \multicolumn{1}{r|}{54.38} & \multicolumn{1}{r|}{74.86} & \multicolumn{1}{r|}{31.24} & \multicolumn{1}{r|}{76.43} & \multicolumn{1}{r|}{28.35} & \multicolumn{1}{r|}{60.80} & \multicolumn{1}{r|}{18.38} & \multicolumn{1}{c|}{-} & \multicolumn{1}{c|}{-} & \multicolumn{1}{c}{-} \\
    \multicolumn{1}{l|}{\textbf{XceptionTime ($nf=16$) \cite{rahimian_zabihi}}} & \multicolumn{1}{r|}{81.41} & \multicolumn{1}{r|}{70.76} & \multicolumn{1}{r|}{85.94} & \multicolumn{1}{r|}{78.23} & \multicolumn{1}{r|}{66.70} & \multicolumn{1}{r|}{56.92} & \multicolumn{1}{r|}{75.41} & \multicolumn{1}{r|}{40.08} & \multicolumn{1}{r|}{79.20} & \multicolumn{1}{r|}{30.66} & \multicolumn{1}{r|}{63.92} & \multicolumn{1}{r|}{25.91} & \multicolumn{1}{c|}{-} & \multicolumn{1}{c|}{-} & \multicolumn{1}{c}{-} \\
    \multicolumn{1}{l|}{\textbf{InceptionTime ($64, 12$) \cite{fawaz_lucas}}} & \multicolumn{1}{r|}{84.14} & \multicolumn{1}{r|}{75.28} & \multicolumn{1}{r|}{87.80} & \multicolumn{1}{r|}{81.62} & \multicolumn{1}{r|}{70.43} & \multicolumn{1}{r|}{61.68} & \multicolumn{1}{r|}{79.08} & \multicolumn{1}{r|}{43.12} & \multicolumn{1}{r|}{81.25} & \multicolumn{1}{r|}{36.48} & \multicolumn{1}{r|}{65.12} & \multicolumn{1}{r|}{29.35} & \multicolumn{1}{c|}{-} & \multicolumn{1}{c|}{-} & \multicolumn{1}{c}{-} \\
    \multicolumn{1}{l|}{\textbf{TST \cite{zerveas_jayaraman}}} & \multicolumn{1}{r|}{80.10} & \multicolumn{1}{r|}{70.75} & \multicolumn{1}{r|}{84.81} & \multicolumn{1}{r|}{78.34} & \multicolumn{1}{r|}{66.12} & \multicolumn{1}{r|}{57.56} & \multicolumn{1}{r|}{77.43} & \multicolumn{1}{r|}{41.27} & \multicolumn{1}{r|}{79.11} & \multicolumn{1}{r|}{29.86} & \multicolumn{1}{r|}{63.39} & \multicolumn{1}{r|}{26.83} & \multicolumn{1}{c|}{-} & \multicolumn{1}{c|}{-} & \multicolumn{1}{c}{-} \\
    \end{tabular}
\end{center}
\end{table*}

\subsection{Evaluation of Sinusoidal Data}
\label{chap_eval_sin}

We train each sinusoidal dataset 10 times, and present results of mean and standard deviation. As a gold standard we apply the transformation based on the known label class, which we define as the \textit{upper bound} for transformation selection, and define the \textit{lower bound} by the classification without the use of any transformation.

Figure~\ref{image_eval_sinus} presents results for the five optimal transport techniques. The classification accuracy notably drops for higher noise rates ($b > 0.7$). Without transformation (\textit{lower bound}) the accuracy is below 20\% (see Figure~\ref{image_eval_sinus_a}, cyan). The \textit{upper bound} (black line) yields an accuracy between 80\% and 100\%. MMD is outperformed by all transformation selection techniques as order 1 is not suitable. For the remaining techniques, the Laplacian regularization improves EMD results, while regularizing Sinkhorn ($L_{p}L_{1}$ and $L_{1}L_{2}$) does not yield better results compared to standard Sinkhorn. Consistently, CC and kMMD outperform CORAL which yields higher accuracies than HoMM of order three and Pearson correlation. For the evaluation of feature embeddings, we refer to Appendix~\ref{sec_app_embeddings}.

\newcommand\tabrotate[1]{\rotatebox{90}{#1\hspace{\tabcolsep}}}
\begin{table*}[t!]
\begin{center}
\setlength{\tabcolsep}{3.5pt}
    \caption{Evaluation results (CRR in \%, mean and standard deviation) for different transformation techniques and transformation selections for all OnHW \cite{ott, ott_tist} datasets based on the CNN+BiLSTM architecture. J = Jeff, S = Stein.}
    \label{table_da_results_all}
    \vspace{-0.2cm}
    \small \begin{tabular}{ p{0.5cm} | p{0.5cm} | p{0.5cm} | p{0.5cm} | p{0.5cm} | p{0.5cm} | p{0.5cm} | p{0.5cm} | p{0.5cm} | p{0.5cm} || p{0.5cm} }
    \multicolumn{1}{c|}{\textbf{Dataset}} & \multicolumn{1}{c|}{\textbf{Method}} & \multicolumn{1}{c|}{\textbf{CC~\cite{pearson}}} & \multicolumn{1}{c|}{\textbf{PC~\cite{pearson}}} & \multicolumn{1}{c|}{\textbf{MMD~\cite{long_cao}}} & \multicolumn{1}{c|}{\textbf{kMMD~\cite{long_zhu_mmd}}} & \multicolumn{1}{c|}{\textbf{HoMM~\cite{chen_fu}}} & \multicolumn{1}{c|}{\textbf{CORAL}} & \multicolumn{1}{c|}{\textbf{CORAL (J)}} & \multicolumn{1}{c||}{\textbf{CORAL (S)}} & \multicolumn{1}{c}{\textbf{w/ T.}} \\ \hline
    & \multicolumn{1}{l|}{\textbf{EMD~\cite{ferradans}}} & \multicolumn{1}{r|}{70.03$\pm$\footnotesize \: 8.4} & \multicolumn{1}{r|}{59.14$\pm$\footnotesize 13.4} & \multicolumn{1}{r|}{61.55$\pm$\footnotesize 18.2} & \multicolumn{1}{r|}{85.02$\pm$\footnotesize 6.9} & \multicolumn{1}{r|}{67.11$\pm$\footnotesize 6.8} & \multicolumn{1}{r|}{75.29$\pm$\footnotesize 16.0} & \multicolumn{1}{r|}{82.24$\pm$\footnotesize 7.1} & \multicolumn{1}{r||}{80.85$\pm$\footnotesize 8.3} & \multicolumn{1}{r}{89.18$\pm$\footnotesize 10.8} \\
    & \multicolumn{1}{l|}{\textbf{SEMD~\cite{flamary_courty}}} & \multicolumn{1}{r|}{70.03$\pm$\footnotesize \: 8.4} & \multicolumn{1}{r|}{59.14$\pm$\footnotesize 13.4} & \multicolumn{1}{r|}{61.55$\pm$\footnotesize 18.2} & \multicolumn{1}{r|}{85.02$\pm$\footnotesize 6.9} & \multicolumn{1}{r|}{65.72$\pm$\footnotesize 8.3} & \multicolumn{1}{r|}{75.29$\pm$\footnotesize 16.0} & \multicolumn{1}{r|}{82.24$\pm$\footnotesize 7.2} & \multicolumn{1}{r||}{80.85$\pm$\footnotesize 8.4} & \multicolumn{1}{r}{89.18$\pm$\footnotesize 10.8} \\
    \multicolumn{1}{c|}{\textbf{OnHW-}} & \multicolumn{1}{l|}{\textbf{Sinkhorn~\cite{courty_flamary}}} & \multicolumn{1}{r|}{64.62$\pm$\footnotesize 19.2} & \multicolumn{1}{r|}{51.97$\pm$\footnotesize \: 5.2} & \multicolumn{1}{r|}{67.25$\pm$\footnotesize \: 8.1} & \multicolumn{1}{r|}{\textbf{85.09}$\pm$\footnotesize 7.7} & \multicolumn{1}{r|}{70.03$\pm$\footnotesize 7.4} & \multicolumn{1}{r|}{78.14$\pm$\footnotesize \: 6.5} & \multicolumn{1}{r|}{80.92$\pm$\footnotesize 8.0} & \multicolumn{1}{r||}{82.31$\pm$\footnotesize 7.8} & \multicolumn{1}{r}{93.35$\pm$\footnotesize \: 8.6} \\
    \multicolumn{1}{c|}{\textbf{symbols}} & \multicolumn{1}{l|}{\textbf{Sink.} ($L_{p}L_{1}$)} & \multicolumn{1}{r|}{64.62$\pm$\footnotesize 19.2} & \multicolumn{1}{r|}{51.97$\pm$\footnotesize \: 5.2} & \multicolumn{1}{r|}{67.25$\pm$\footnotesize \: 8.1} & \multicolumn{1}{r|}{\textbf{85.09}$\pm$\footnotesize 7.7} & \multicolumn{1}{r|}{69.96$\pm$\footnotesize 5.6} & \multicolumn{1}{r|}{78.14$\pm$\footnotesize \: 6.5} & \multicolumn{1}{r|}{80.92$\pm$\footnotesize 8.0} & \multicolumn{1}{r||}{82.31$\pm$\footnotesize 7.8} & \multicolumn{1}{r}{93.35$\pm$\footnotesize \: 8.6} \\
    & \multicolumn{1}{l|}{\textbf{Sink.} ($L_{1}L_{2}$)} & \multicolumn{1}{r|}{64.62$\pm$\footnotesize 19.2} & \multicolumn{1}{r|}{51.97$\pm$\footnotesize \: 5.2} & \multicolumn{1}{r|}{67.25$\pm$\footnotesize \: 8.1} & \multicolumn{1}{r|}{\textbf{85.09}$\pm$\footnotesize 7.7} & \multicolumn{1}{r|}{68.64$\pm$\footnotesize 7.3} & \multicolumn{1}{r|}{78.14$\pm$\footnotesize \: 6.5} & \multicolumn{1}{r|}{80.92$\pm$\footnotesize 8.0} & \multicolumn{1}{r||}{82.31$\pm$\footnotesize 7.8} & \multicolumn{1}{r}{93.35$\pm$\footnotesize \: 8.6} \\
    & \multicolumn{1}{l|}{\textbf{CORAL~\cite{sun_feng}}} & \multicolumn{1}{r|}{5.48$\pm$\footnotesize \: 6.8} & \multicolumn{1}{r|}{2.78$\pm$\footnotesize \: 2.8} & \multicolumn{1}{r|}{5.49$\pm$\footnotesize \: 0.1} & \multicolumn{1}{r|}{6.80$\pm$\footnotesize 2.2} & \multicolumn{1}{r|}{5.49$\pm$\footnotesize 0.1} & \multicolumn{1}{r|}{9.66$\pm$\footnotesize \: 7.3} & \multicolumn{1}{r|}{5.49$\pm$\footnotesize 0.1} & \multicolumn{1}{r||}{4.17$\pm$\footnotesize 2.4} & \multicolumn{1}{r}{39.77$\pm$\footnotesize 10.0} \\ \hline
    & \multicolumn{1}{l|}{\textbf{EMD~\cite{ferradans}}} & \multicolumn{1}{r|}{68.90$\pm$\footnotesize 14.4} & \multicolumn{1}{r|}{56.16$\pm$\footnotesize 20.3} & \multicolumn{1}{r|}{59.33$\pm$\footnotesize 14.9} & \multicolumn{1}{r|}{79.00$\pm$\footnotesize 11.7} & \multicolumn{1}{r|}{53.87$\pm$\footnotesize 10.5} & \multicolumn{1}{r|}{73.61$\pm$\footnotesize 11.4} & \multicolumn{1}{r|}{77.39$\pm$\footnotesize 11.6} & \multicolumn{1}{r||}{72.09$\pm$\footnotesize 13.2} & \multicolumn{1}{r}{85.78$\pm$\footnotesize \: 7.7} \\
    \multicolumn{1}{c|}{\textbf{split}} & \multicolumn{1}{l|}{\textbf{SEMD~\cite{flamary_courty}}} & \multicolumn{1}{r|}{68.90$\pm$\footnotesize 14.4} & \multicolumn{1}{r|}{56.16$\pm$\footnotesize 20.3} & \multicolumn{1}{r|}{59.33$\pm$\footnotesize 14.9} & \multicolumn{1}{r|}{79.00$\pm$\footnotesize 11.7} & \multicolumn{1}{r|}{53.05$\pm$\footnotesize 10.0} & \multicolumn{1}{r|}{73.61$\pm$\footnotesize 11.4} & \multicolumn{1}{r|}{77.39$\pm$\footnotesize 11.6} & \multicolumn{1}{r||}{72.09$\pm$\footnotesize 13.2} & \multicolumn{1}{r}{85.78$\pm$\footnotesize \: 7.7} \\
    \multicolumn{1}{c|}{\textbf{OnHW-}} & \multicolumn{1}{l|}{\textbf{Sinkhorn~\cite{courty_flamary}}} & \multicolumn{1}{r|}{69.69$\pm$\footnotesize 17.1} & \multicolumn{1}{r|}{53.31$\pm$\footnotesize 16.8} & \multicolumn{1}{r|}{67.12$\pm$\footnotesize \: 9.9} & \multicolumn{1}{r|}{\textbf{84.03}$\pm$\footnotesize \: 9.4} & \multicolumn{1}{r|}{62.37$\pm$\footnotesize 13.3} & \multicolumn{1}{r|}{79.97$\pm$\footnotesize 10.8} & \multicolumn{1}{r|}{82.24$\pm$\footnotesize 10.3} & \multicolumn{1}{r||}{76.12$\pm$\footnotesize 13.0} & \multicolumn{1}{r}{90.26$\pm$\footnotesize \: 6.3} \\
    \multicolumn{1}{c|}{\textbf{equations}} & \multicolumn{1}{l|}{\textbf{Sink.} ($L_{p}L_{1}$)} & \multicolumn{1}{r|}{69.69$\pm$\footnotesize 17.1} & \multicolumn{1}{r|}{53.31$\pm$\footnotesize 16.8} & \multicolumn{1}{r|}{67.12$\pm$\footnotesize \: 9.9} & \multicolumn{1}{r|}{\textbf{84.03}$\pm$\footnotesize \: 9.4} & \multicolumn{1}{r|}{61.86$\pm$\footnotesize 13.2} & \multicolumn{1}{r|}{79.97$\pm$\footnotesize 10.8} & \multicolumn{1}{r|}{82.24$\pm$\footnotesize 10.3} & \multicolumn{1}{r||}{76.12$\pm$\footnotesize 13.0} & \multicolumn{1}{r}{90.26$\pm$\footnotesize \: 6.3} \\
    & \multicolumn{1}{l|}{\textbf{Sink.} ($L_{1}L_{2}$)} & \multicolumn{1}{r|}{69.69$\pm$\footnotesize 17.1} & \multicolumn{1}{r|}{53.33$\pm$\footnotesize 16.8} & \multicolumn{1}{r|}{67.12$\pm$\footnotesize \: 9.9} & \multicolumn{1}{r|}{\textbf{84.03}$\pm$\footnotesize \: 9.4} & \multicolumn{1}{r|}{62.29$\pm$\footnotesize 13.4} & \multicolumn{1}{r|}{79.97$\pm$\footnotesize 10.8} & \multicolumn{1}{r|}{82.24$\pm$\footnotesize 10.3} & \multicolumn{1}{r||}{76.12$\pm$\footnotesize 13.0} & \multicolumn{1}{r}{90.26$\pm$\footnotesize \: 6.3} \\
    & \multicolumn{1}{l|}{\textbf{CORAL~\cite{sun_feng}}} & \multicolumn{1}{r|}{22.19$\pm$\footnotesize 12.2} & \multicolumn{1}{r|}{15.82$\pm$\footnotesize 12.6} & \multicolumn{1}{r|}{15.05$\pm$\footnotesize \: 8.4} & \multicolumn{1}{r|}{19.63$\pm$\footnotesize 13.0} & \multicolumn{1}{r|}{19.01$\pm$\footnotesize \: 9.0} & \multicolumn{1}{r|}{18.27$\pm$\footnotesize 10.8} & \multicolumn{1}{r|}{18.18$\pm$\footnotesize 11.2} & \multicolumn{1}{r||}{17.72$\pm$\footnotesize 11.8} & \multicolumn{1}{r}{30.34$\pm$\footnotesize 11.0} \\ \hline
    & \multicolumn{1}{l|}{\textbf{EMD~\cite{ferradans}}} & \multicolumn{1}{r|}{68.41$\pm$\footnotesize 13.7} & \multicolumn{1}{r|}{54.27$\pm$\footnotesize \: 8.6} & \multicolumn{1}{r|}{33.06$\pm$\footnotesize 13.7} & \multicolumn{1}{r|}{77.72$\pm$\footnotesize 11.0} & \multicolumn{1}{r|}{50.51$\pm$\footnotesize \: 9.5} & \multicolumn{1}{r|}{64.25$\pm$\footnotesize 10.0} & \multicolumn{1}{r|}{80.08$\pm$\footnotesize 9.5} & \multicolumn{1}{r||}{79.41$\pm$\footnotesize 9.5} & \multicolumn{1}{r}{98.19$\pm$\footnotesize 2.1} \\
    & \multicolumn{1}{l|}{\textbf{SEMD~\cite{flamary_courty}}} & \multicolumn{1}{r|}{68.41$\pm$\footnotesize 13.7} & \multicolumn{1}{r|}{54.27$\pm$\footnotesize \: 8.6} & \multicolumn{1}{r|}{33.06$\pm$\footnotesize 13.7} & \multicolumn{1}{r|}{77.72$\pm$\footnotesize 11.0} & \multicolumn{1}{r|}{50.57$\pm$\footnotesize \: 9.5} & \multicolumn{1}{r|}{64.25$\pm$\footnotesize 10.0} & \multicolumn{1}{r|}{80.08$\pm$\footnotesize 9.5} & \multicolumn{1}{r||}{79.41$\pm$\footnotesize 9.5} & \multicolumn{1}{r}{98.19$\pm$\footnotesize 2.1} \\
    \multicolumn{1}{c|}{\textbf{OnHW-}} & \multicolumn{1}{l|}{\textbf{Sinkhorn~\cite{courty_flamary}}} & \multicolumn{1}{r|}{81.56$\pm$\footnotesize 12.3} & \multicolumn{1}{r|}{69.27$\pm$\footnotesize 12.6} & \multicolumn{1}{r|}{51.54$\pm$\footnotesize 16.6} & \multicolumn{1}{r|}{83.55$\pm$\footnotesize 10.1} & \multicolumn{1}{r|}{67.31$\pm$\footnotesize 10.5} & \multicolumn{1}{r|}{77.36$\pm$\footnotesize 11.3} & \multicolumn{1}{r|}{82.33$\pm$\footnotesize 9.5} & \multicolumn{1}{r||}{\textbf{85.22}$\pm$\footnotesize 8.2} & \multicolumn{1}{r}{100.00$\pm$\footnotesize 0.0} \\
    \multicolumn{1}{c|}{\textbf{chars}} & \multicolumn{1}{l|}{\textbf{Sink.} ($L_{p}L_{1}$)} & \multicolumn{1}{r|}{81.56$\pm$\footnotesize 12.3} & \multicolumn{1}{r|}{69.27$\pm$\footnotesize 12.6} & \multicolumn{1}{r|}{51.54$\pm$\footnotesize 16.6} & \multicolumn{1}{r|}{83.55$\pm$\footnotesize 10.1} & \multicolumn{1}{r|}{66.88$\pm$\footnotesize 11.1} & \multicolumn{1}{r|}{77.36$\pm$\footnotesize 11.3} & \multicolumn{1}{r|}{82.33$\pm$\footnotesize 9.5} & \multicolumn{1}{r||}{\textbf{85.22}$\pm$\footnotesize 8.2} & \multicolumn{1}{r}{100.00$\pm$\footnotesize 0.0} \\
    \multicolumn{1}{c|}{\textbf{(lower)}} & \multicolumn{1}{l|}{\textbf{Sink.} ($L_{1}L_{2}$)} & \multicolumn{1}{r|}{81.56$\pm$\footnotesize 12.3} & \multicolumn{1}{r|}{69.27$\pm$\footnotesize 12.6} & \multicolumn{1}{r|}{51.54$\pm$\footnotesize 16.6} & \multicolumn{1}{r|}{83.55$\pm$\footnotesize 10.1} & \multicolumn{1}{r|}{67.96$\pm$\footnotesize 10.0} & \multicolumn{1}{r|}{77.36$\pm$\footnotesize 11.3} & \multicolumn{1}{r|}{82.32$\pm$\footnotesize 9.5} & \multicolumn{1}{r||}{\textbf{85.22}$\pm$\footnotesize 8.2} & \multicolumn{1}{r}{100.00$\pm$\footnotesize 0.0} \\
    & \multicolumn{1}{l|}{\textbf{CORAL~\cite{sun_feng}}} & \multicolumn{1}{r|}{4.56$\pm$\footnotesize \: 4.1} & \multicolumn{1}{r|}{8.65$\pm$\footnotesize \: 6.3} & \multicolumn{1}{r|}{4.28$\pm$\footnotesize \: 0.4} & \multicolumn{1}{r|}{3.87$\pm$\footnotesize \: 3.9} & \multicolumn{1}{r|}{4.28$\pm$\footnotesize \: 0.4} & \multicolumn{1}{r|}{4.28$\pm$\footnotesize \: 0.4} & \multicolumn{1}{r|}{4.28$\pm$\footnotesize 0.4} & \multicolumn{1}{r||}{7.40$\pm$\footnotesize 4.3} & \multicolumn{1}{r}{47.13$\pm$\footnotesize 9.2} \\ \hline
    & \multicolumn{1}{l|}{\textbf{EMD~\cite{ferradans}}} & \multicolumn{1}{r|}{62.98$\pm$\footnotesize 18.1} & \multicolumn{1}{r|}{53.37$\pm$\footnotesize 15.1} & \multicolumn{1}{r|}{33.93$\pm$\footnotesize 10.2} & \multicolumn{1}{r|}{67.94$\pm$\footnotesize 13.1} & \multicolumn{1}{r|}{44.06$\pm$\footnotesize 12.0} & \multicolumn{1}{r|}{60.70$\pm$\footnotesize 14.0} & \multicolumn{1}{r|}{79.87$\pm$\footnotesize 15.3} & \multicolumn{1}{r||}{76.29$\pm$\footnotesize 16.0} & \multicolumn{1}{r}{99.31$\pm$\footnotesize \: 2.0} \\
    & \multicolumn{1}{l|}{\textbf{SEMD~\cite{flamary_courty}}} & \multicolumn{1}{r|}{62.98$\pm$\footnotesize 18.1} & \multicolumn{1}{r|}{53.37$\pm$\footnotesize 15.1} & \multicolumn{1}{r|}{33.93$\pm$\footnotesize 10.2} & \multicolumn{1}{r|}{67.94$\pm$\footnotesize 13.1} & \multicolumn{1}{r|}{42.67$\pm$\footnotesize 12.3} & \multicolumn{1}{r|}{60.70$\pm$\footnotesize 14.0} & \multicolumn{1}{r|}{79.87$\pm$\footnotesize 15.3} & \multicolumn{1}{r||}{76.29$\pm$\footnotesize 16.0} & \multicolumn{1}{r}{99.31$\pm$\footnotesize \: 2.0} \\
    \multicolumn{1}{c|}{\textbf{OnHW-}} & \multicolumn{1}{l|}{\textbf{Sinkhorn~\cite{courty_flamary}}} & \multicolumn{1}{r|}{78.82$\pm$\footnotesize 13.0} & \multicolumn{1}{r|}{73.32$\pm$\footnotesize 15.5} & \multicolumn{1}{r|}{55.82$\pm$\footnotesize 17.6} & \multicolumn{1}{r|}{82.25$\pm$\footnotesize 14.5} & \multicolumn{1}{r|}{57.57$\pm$\footnotesize 11.0} & \multicolumn{1}{r|}{74.56$\pm$\footnotesize 12.6} & \multicolumn{1}{r|}{80.96$\pm$\footnotesize 16.3} & \multicolumn{1}{r||}{78.25$\pm$\footnotesize 14.2} & \multicolumn{1}{r}{99.65$\pm$\footnotesize \: 1.0} \\
    \multicolumn{1}{c|}{\textbf{chars}} & \multicolumn{1}{l|}{\textbf{Sink.} ($L_{p}L_{1}$)} & \multicolumn{1}{r|}{78.82$\pm$\footnotesize 13.0} & \multicolumn{1}{r|}{73.32$\pm$\footnotesize 15.5} & \multicolumn{1}{r|}{55.82$\pm$\footnotesize 17.6} & \multicolumn{1}{r|}{\textbf{82.26}$\pm$\footnotesize 14.5} & \multicolumn{1}{r|}{57.57$\pm$\footnotesize 11.3} & \multicolumn{1}{r|}{74.56$\pm$\footnotesize 12.6} & \multicolumn{1}{r|}{80.96$\pm$\footnotesize 16.3} & \multicolumn{1}{r||}{78.25$\pm$\footnotesize 14.2} & \multicolumn{1}{r}{99.65$\pm$\footnotesize \: 1.0} \\
    \multicolumn{1}{c|}{\textbf{(upper)}} & \multicolumn{1}{l|}{\textbf{Sink.} ($L_{1}L_{2}$)} & \multicolumn{1}{r|}{78.82$\pm$\footnotesize 13.0} & \multicolumn{1}{r|}{73.32$\pm$\footnotesize 15.5} & \multicolumn{1}{r|}{55.82$\pm$\footnotesize 17.6} & \multicolumn{1}{r|}{\textbf{82.26}$\pm$\footnotesize 14.5} & \multicolumn{1}{r|}{57.34$\pm$\footnotesize 10.2} & \multicolumn{1}{r|}{74.56$\pm$\footnotesize 12.6} & \multicolumn{1}{r|}{80.96$\pm$\footnotesize 16.3} & \multicolumn{1}{r||}{78.25$\pm$\footnotesize 14.2} & \multicolumn{1}{r}{99.65$\pm$\footnotesize \: 1.0} \\
    & \multicolumn{1}{l|}{\textbf{CORAL~\cite{sun_feng}}} & \multicolumn{1}{r|}{7.03$\pm$\footnotesize \: 4.6} & \multicolumn{1}{r|}{21.51$\pm$\footnotesize 10.7} & \multicolumn{1}{r|}{4.26$\pm$\footnotesize \: 1.7} & \multicolumn{1}{r|}{6.27$\pm$\footnotesize \: 3.3} & \multicolumn{1}{r|}{4.26$\pm$\footnotesize \: 1.7} & \multicolumn{1}{r|}{4.26$\pm$\footnotesize \: 1.0} & \multicolumn{1}{r|}{4.26$\pm$\footnotesize \: 1.0} & \multicolumn{1}{r||}{11.11$\pm$\footnotesize \: 6.4} & \multicolumn{1}{r}{52.10$\pm$\footnotesize 16.5} \\ \hline
    & \multicolumn{1}{l|}{\textbf{EMD~\cite{ferradans}}} & \multicolumn{1}{r|}{62.18$\pm$\footnotesize 12.1} & \multicolumn{1}{r|}{51.79$\pm$\footnotesize 12.9} & \multicolumn{1}{r|}{31.57$\pm$\footnotesize 11.7} & \multicolumn{1}{r|}{69.62$\pm$\footnotesize \: 9.3} & \multicolumn{1}{r|}{45.21$\pm$\footnotesize 11.6} & \multicolumn{1}{r|}{57.79$\pm$\footnotesize 11.4} & \multicolumn{1}{r|}{69.56$\pm$\footnotesize 9.0} & \multicolumn{1}{r||}{63.96$\pm$\footnotesize 10.4} & \multicolumn{1}{r}{94.97$\pm$\footnotesize 3.4} \\
    & \multicolumn{1}{l|}{\textbf{SEMD~\cite{flamary_courty}}} & \multicolumn{1}{r|}{62.18$\pm$\footnotesize 12.1} & \multicolumn{1}{r|}{51.79$\pm$\footnotesize 12.9} & \multicolumn{1}{r|}{31.57$\pm$\footnotesize 11.7} & \multicolumn{1}{r|}{69.62$\pm$\footnotesize \: 9.3} & \multicolumn{1}{r|}{42.80$\pm$\footnotesize 12.2} & \multicolumn{1}{r|}{57.79$\pm$\footnotesize 11.4} & \multicolumn{1}{r|}{69.56$\pm$\footnotesize 9.1} & \multicolumn{1}{r||}{63.96$\pm$\footnotesize 10.4} & \multicolumn{1}{r}{94.98$\pm$\footnotesize 3.4} \\
    \multicolumn{1}{c|}{\textbf{OnHW-}} & \multicolumn{1}{l|}{\textbf{Sinkhorn~\cite{courty_flamary}}} & \multicolumn{1}{r|}{65.16$\pm$\footnotesize 12.1} & \multicolumn{1}{r|}{62.92$\pm$\footnotesize 10.4} & \multicolumn{1}{r|}{44.16$\pm$\footnotesize 14.7} & \multicolumn{1}{r|}{\textbf{73.87}$\pm$\footnotesize 14.7} & \multicolumn{1}{r|}{56.84$\pm$\footnotesize \: 9.6} & \multicolumn{1}{r|}{67.85$\pm$\footnotesize 10.5} & \multicolumn{1}{r|}{72.30$\pm$\footnotesize 9.1} & \multicolumn{1}{r||}{66.50$\pm$\footnotesize \: 7.7} & \multicolumn{1}{r}{96.25$\pm$\footnotesize 3.5} \\
    \multicolumn{1}{c|}{\textbf{chars}} & \multicolumn{1}{l|}{\textbf{Sink.} ($L_{p}L_{1}$)} & \multicolumn{1}{r|}{65.16$\pm$\footnotesize 12.1} & \multicolumn{1}{r|}{62.93$\pm$\footnotesize 10.4} & \multicolumn{1}{r|}{44.16$\pm$\footnotesize 14.7} & \multicolumn{1}{r|}{\textbf{73.87}$\pm$\footnotesize 11.1} & \multicolumn{1}{r|}{55.90$\pm$\footnotesize 10.1} & \multicolumn{1}{r|}{67.85$\pm$\footnotesize 10.5} & \multicolumn{1}{r|}{72.30$\pm$\footnotesize 9.1} & \multicolumn{1}{r||}{66.50$\pm$\footnotesize \: 7.7} & \multicolumn{1}{r}{96.25$\pm$\footnotesize 3.5} \\
    \multicolumn{1}{c|}{\textbf{(comb.)}} & \multicolumn{1}{l|}{\textbf{Sink.} ($L_{1}L_{2}$)} & \multicolumn{1}{r|}{65.16$\pm$\footnotesize 12.1} & \multicolumn{1}{r|}{62.92$\pm$\footnotesize 10.4} & \multicolumn{1}{r|}{44.16$\pm$\footnotesize 14.7} & \multicolumn{1}{r|}{\textbf{73.87}$\pm$\footnotesize 11.1} & \multicolumn{1}{r|}{56.48$\pm$\footnotesize \: 9.2} & \multicolumn{1}{r|}{67.85$\pm$\footnotesize 10.5} & \multicolumn{1}{r|}{72.30$\pm$\footnotesize 9.1} & \multicolumn{1}{r||}{66.50$\pm$\footnotesize \: 7.7} & \multicolumn{1}{r}{96.25$\pm$\footnotesize 3.5} \\
    & \multicolumn{1}{l|}{\textbf{CORAL~\cite{sun_feng}}} & \multicolumn{1}{r|}{3.35$\pm$\footnotesize \: 1.2} & \multicolumn{1}{r|}{4.11$\pm$\footnotesize \: 2.6} & \multicolumn{1}{r|}{0.87$\pm$\footnotesize \: 1.3} & \multicolumn{1}{r|}{3.03$\pm$\footnotesize \: 1.8} & \multicolumn{1}{r|}{0.87$\pm$\footnotesize \: 1.3} & \multicolumn{1}{r|}{0.87$\pm$\footnotesize \: 1.0} & \multicolumn{1}{r|}{0.87$\pm$\footnotesize 1.3} & \multicolumn{1}{r||}{3.24$\pm$\footnotesize \: 2.2} & \multicolumn{1}{r}{32.41$\pm$\footnotesize 8.7} \\
    \end{tabular}
    \vspace{-0.06cm}
\end{center}
\end{table*}

\subsection{Evaluation of OnHW Recognition}
\label{chap_eval_da}

\paragraph{Baseline Results.} We train all three OnHW right- and left-handed writer datasets for writer-dependent (WD) and writer-independent (WI) tasks. Architectures are taken from~\cite{ott_tist} (CNN+ BiLSTM) and the tsai toolbox \cite{tsai} (for all other models) \cite{karim_majumdar, zou_wang, wang_yan_oates, he_zhang, rahimian_zabihi, fawaz_lucas, zerveas_jayaraman}. Results of these comparisons are given in Tables~\ref{table_baseline_symbols} and \ref{table_baseline_chars}. For the OnHW-symbols dataset, the CNN+BiLSTM model outperforms all architectures, while for the split OnHW-equations dataset, InceptionTime~\cite{fawaz_lucas} outperforms the CNN+BiLSTM architecture on the right-handed WI task. On the OnHW-chars dataset, the accuracy of the CNN+BiLSTM model is notably higher compared to all other models, and we hence choose the CNN+BiLSTM model for further experiments. Simply classifying the left-handed writer samples with the model pre-trained on right-handed writer data yields low accuracies (19.18\% on the OnHW-symbols dataset and 33.52\% on the split OnHW-equations dataset) as left-handed writer samples are out-of-distribution with respect to the right-handed ones (see Figure~\ref{image_emb_4}). Hence, an efficient DA technique is necessary.

\paragraph{Evaluation of Domain Adaptation.} We first evaluate different distance metrices for optimal transport. The respective hyperparameter search results are given in Appendix~\ref{sec_app_searches}. Based on these findings, we choose the squared Euclidean metric for all further applications. Table~\ref{table_da_results_all} summarizes all results for DA pre-trained on the target domain data $\mathcal{U}_{T_t}$, adapted on the source domain data $\mathcal{U}_{S_t}$, and validated on $\mathcal{U}_{S_v}$. We train each left-handed writer separately and report average results with their respective standard deviation. For an evaluation for each left-handed writer, see Appendix~\ref{sec_app_writer_eval}. The last column in~Table~\ref{table_da_results_all} shows the \textit{upper bound} using the known transformation. EMD~\cite{ferradans} and SEMD~\cite{flamary_courty} perform similar, as well as Sinkhorn~\cite{courty_flamary} without and with ($L_{p}L_{1}$, $L_{1}L_{2}$) regularization. Sinkhorn transport consistently outperforms EMD (see also \cite{altschuler}). The model with CORAL~\cite{sun_feng} for transformation computation fails to classify the time-series data, even with known transformation selection. The MMD~\cite{long_cao} approach (of order 1) yields the lowest classification accuracy, and is notably be improved with CORAL~\cite{sun_feng} (of order 2). Increasing the order to 3 (HoMM~\cite{chen_fu}), decreases the accuracy as a higher number of iterations is required. The kernalized MMD (kMMD)~\cite{long_zhu_mmd} approach (of order 1) yields the highest classification accuracies. Kernalizing HoMM leads to extremely long runtimes. Jeff and Stein CORAL outperform the standard CORAL method as these are not dependent on its inverse \cite{harandi_salzmann}. Interestingly, CC performs better than the scale-invariant PC.

\paragraph{Runtimes} We demonstrate runtimes for transformation computation and selection methods exemplary on the OnHW-symbols dataset for one writer averaged over all samples. To find the optimal transformation, EMD (0.0033\textit{s}) and Sinkhorn (0.0042\textit{s}) are the fastest methods, while the regularization increases the computation time (SEMD: 1.8756\textit{s}, Sinkhorn $L_{p}L_{1}$: 0.025, $L_{1}L_{2}$: 0.0526\textit{s}). CORAL leads to extremely long runtimes of 405.7\textit{s}. Applying the transformation gives the following runtimes in ascending order: CC (0.013\textit{s}), MMD (0.013\textit{s}), PC (0.021\textit{s}), kMMD (0.024\textit{s}), CORAL (0.03\textit{s}), CORAL (S) (0.109\textit{s}), CORAL (J) (0.181\textit{s}), HoMM (2.844\textit{s}). We conclude that Sinkhorn with kMMD is the best trade-off between classification accuracy and runtime. 

\section{Conclusion}
\label{chap_summary}

We addressed DA for time-series classification by combining a large variety of DA techniques with transformation selection methods. We used optimal transport and correlation alignment techniques to transform features of a source domain into features of a target domain. At inference, we compared correlation metrics and methods based on MMD as embedding distance metrics to select the optimal transformation. A broad study on synthetic univariate time-series data and MTS OnHW datasets showed that Sinkhorn transport can outperform EMD and CORAL. The kernalized MMD metric yields the highest classification accuracies. Our DA approach yields higher accuracies than transfer learning on small adaptation datasets.

\section*{Acknowledgments}
Supported by the Federal Ministry of Education and Research (BMBF) of Germany by Grant No. 01IS18036A (David R\"ugamer) and by the research program Human-Computer-Interaction through the project ``Schreibtrainer'', Grant No. 16SV8228, as well as by the Bavarian Ministry for Economic Affairs, Infrastructure, Transport and Technology through the Center for Analytics-Data-Applications (ADA-Center) within the framework of ``BAYERN DIGITAL II''.

\bibliographystyle{ACM-Reference-Format}
\bibliography{ACMMM2022}

\clearpage
\appendix
\section{Appendices}
\label{sec_appendix}

\subsection{Notations}
\label{sec_app_notation}

\begin{table}[h!]
\begin{center}
\setlength{\tabcolsep}{3.7pt}
    \caption{Overview of notations used for our domain adaptation methodology.}
    \label{table_notations}
    \small \begin{tabular}{ p{0.5cm} | p{0.5cm} }
    \multicolumn{1}{c|}{\textbf{Notation}} & \multicolumn{1}{c}{\textbf{Description}} \\ \hline
    \multicolumn{2}{l}{\textbf{Multivariate Time-Series (MTS) Classification}} \\
    \multicolumn{1}{l|}{$\mathbf{U} = \{\mathbf{u}_1,\ldots,\mathbf{u}_m\} \in \mathbb{R}^{m \times l}$} & \multicolumn{1}{l}{An MTS (ordered sequence) of $l \in \mathbb{N}$ streams} \\
    \multicolumn{1}{l|}{$\mathbf{u}_i = (u_{i,1},\ldots, u_{i,l})$} & \multicolumn{1}{l}{A stream of an MTS} \\
    \multicolumn{1}{r|}{$i \in \{1,\ldots,m\}$} & \multicolumn{1}{l}{} \\
    \multicolumn{1}{l|}{$m \in \mathbb{N}$} & \multicolumn{1}{l}{Length of a time-series} \\
    \multicolumn{1}{l|}{$\mathcal{U} = \{\mathbf{U}_1,\ldots,\mathbf{U}_{n_U}\}$} & \multicolumn{1}{l}{Set of $n_U$ time-series with $\mathcal{U} \in \mathbb{R}^{n_U \times m \times l}$} \\
    \multicolumn{1}{l|}{$y \in \mathcal{Y}$} & \multicolumn{1}{l}{Unknown class label for a given MTS} \\
    \multicolumn{1}{l|}{$\mathcal{Y}$} & \multicolumn{1}{l}{Label space} \\
    \multicolumn{1}{l|}{$|\mathcal{Y}|$} & \multicolumn{1}{l}{Number of class labels} \\
    \multicolumn{1}{l|}{\textbf{Domain Adaptation}} & \\
    \multicolumn{1}{l|}{$\mathcal{U}_T$} & \multicolumn{1}{l}{Target domain dataset} \\
    \multicolumn{1}{l|}{$\mathcal{U}_S$} & \multicolumn{1}{l}{Source domain dataset} \\
    \multicolumn{1}{l|}{$\mathcal{U}_{T_t}$} & \multicolumn{1}{l}{Training subset of the target domain dataset} \\
    \multicolumn{1}{l|}{$\mathcal{U}_{T_v}$} & \multicolumn{1}{l}{Validation subset of the target domain dataset} \\
    \multicolumn{1}{l|}{$\mathcal{U}_{S_t}$} & \multicolumn{1}{l}{Training subset of the source domain dataset} \\
    \multicolumn{1}{l|}{$\mathcal{U}_{S_v}$} & \multicolumn{1}{l}{Validation subset of the source domain dataset} \\
    \multicolumn{1}{l|}{$\mathbf{U}_{T}$} & \multicolumn{1}{l}{MTS of the target domain dataset} \\
    \multicolumn{1}{l|}{$\mathbf{U}_{S}$} & \multicolumn{1}{l}{MTS of the source domain dataset} \\
    \multicolumn{1}{l|}{$\mathbf{U}_{T_t}$} & \multicolumn{1}{l}{Trainings subset of the target domain dataset} \\
    \multicolumn{1}{l|}{$\mathbf{U}_{T_v}$} & \multicolumn{1}{l}{Valdiation subset of the target domain dataset} \\
    \multicolumn{1}{l|}{$\mathbf{U}_{S_t}$} & \multicolumn{1}{l}{Trainings subset of the source domain dataset} \\
    \multicolumn{1}{l|}{$\mathbf{U}_{S_v}$} & \multicolumn{1}{l}{Valdiation subset of the source domain dataset} \\
    \multicolumn{1}{l|}{$\mu_T$} & \multicolumn{1}{l}{Mean of $\mathcal{U}_T$} \\
    \multicolumn{1}{l|}{$\mu_S$} & \multicolumn{1}{l}{Mean of $\mathcal{U}_S$} \\
    \multicolumn{1}{l|}{$\mathcal{C}_T$} & \multicolumn{1}{l}{Covariance matrix of $\mathcal{U}_T$} \\
    \multicolumn{1}{l|}{$\mathcal{C}_S$} & \multicolumn{1}{l}{Covariance matrix of $\mathcal{U}_S$} \\
    \multicolumn{1}{l|}{$\mathbf{T}: \mathcal{D}_S \xrightarrow{} \mathcal{D}_T$} & \multicolumn{1}{l}{Transformation} \\
    \multicolumn{1}{l|}{$f(\mathbf{U}_{T_t}) \in \mathbb{R}^{q_T \times w_T}$} & \multicolumn{1}{l}{Latent embedding of the training MTS} \\
    \multicolumn{1}{l|}{} & \multicolumn{1}{l}{of the target domain dataset} \\
    \multicolumn{1}{l|}{$f(\mathbf{U}_{S_t}) \in \mathbb{R}^{q_S \times w_S}$} & \multicolumn{1}{l}{Latent embedding of the training MTS} \\
    \multicolumn{1}{l|}{} & \multicolumn{1}{l}{of the source domain dataset} \\
    \multicolumn{1}{l|}{$q_*, w_*$} & \multicolumn{1}{l}{Size of the embedding $f(\mathbf{U}_{*_+}) \in \mathbb{R}^{q_* \times w_*}$} \\
    \multicolumn{1}{l|}{$\mathcal{X}$} & \multicolumn{1}{l}{Feature space} \\
    \multicolumn{1}{l|}{$P(\mathcal{X})$} & \multicolumn{1}{l}{Marginal probability of $\mathcal{X}$} \\
    \multicolumn{1}{l|}{$P(\mathcal{X}, \mathcal{Y})$} & \multicolumn{1}{l}{Joint distribution of $\mathcal{X}$ and $\mathcal{Y}$} \\
    \multicolumn{1}{l|}{$P(\mathcal{Y}|\mathcal{X})$} & \multicolumn{1}{l}{Conditional distribution between $\mathcal{X}$ and $\mathcal{Y}$} \\
    \multicolumn{1}{l|}{$\mathcal{D}$} & \multicolumn{1}{l}{Domain} \\
    \multicolumn{1}{l|}{$\mathcal{D}_T = \{\mathcal{X}_T^i, \mathcal{Y}_T^i\}_{i=1}^{\mathcal{N}_T}$} & \multicolumn{1}{l}{Target domain of $\mathcal{N}_T$ labeled samples} \\
    \multicolumn{1}{l|}{$\mathcal{D}_S = \{\mathcal{X}_S^i, \mathcal{Y}_S^i\}_{i=1}^{\mathcal{N}_S}$} & \multicolumn{1}{l}{Source domain of $\mathcal{N}_S$ labeled samples} \\
    \multicolumn{1}{l|}{$\mathcal{H}$} & \multicolumn{1}{l}{Hypothesis space} \\
    \multicolumn{1}{l|}{$\mathcal{R}_T$} & \multicolumn{1}{l}{Target domain error} \\
    \multicolumn{1}{l|}{$\mathcal{R}_S$} & \multicolumn{1}{l}{Source domain error} \\
    \multicolumn{1}{l|}{$d_{\mathcal{H}\Delta\mathcal{H}}$} & \multicolumn{1}{l}{Discrepancy distance between $\mathcal{D}_S$ and $\mathcal{D}_T$} \\
    \multicolumn{1}{l|}{} & \multicolumn{1}{l}{w.r.t. the hypothesis space $\mathcal{H}$} \\
    \multicolumn{1}{l|}{\textbf{Optimal Transport}} & \\
    \multicolumn{1}{l|}{$C(\mathbf{T})$} & \multicolumn{1}{l}{Transportation cost} \\
    \multicolumn{1}{l|}{$c: \mathcal{D}_T \times \mathcal{D}_S \xrightarrow{} \mathbb{R}^+$} & \multicolumn{1}{l}{Distance function} \\
    \multicolumn{1}{l|}{$\mathbf{T}_0$} & \multicolumn{1}{l}{Optimal transportation problem} \\
    \multicolumn{1}{l|}{$\alpha \in \Theta$} & \multicolumn{1}{l}{General coupling} \\
    \multicolumn{1}{l|}{$\Theta \in P(\mathcal{D}_T, \mathcal{D}_S)$} & \multicolumn{1}{l}{Set of all probabilistic couplings} \\
    \multicolumn{1}{l|}{$W_p$} & \multicolumn{1}{l}{Wasserstein distance of order $p$} \\
    \multicolumn{1}{l|}{$c(\mathbf{U}_T, \mathbf{U}_S)$} & \multicolumn{1}{l}{Cost function} \\
    \multicolumn{1}{l|}{$d(\mathbf{U}_T, \mathbf{U}_S)^p$} & \multicolumn{1}{l}{Distance function of order $p$} \\
    \end{tabular}
\end{center}
\end{table}

\subsection{Correlation Alignment (CORAL)}
\label{sec_app_coral}

The calculation of subspace-based methods is simple and efficient. Hence, we use CORAL~\cite{sun_feng}, which minimizes the domain shift by aligning the second-order statistics (i.e., the original feature distributions of source and target domains). Suppose $\mu_T$ and $\mu_S$ are the means of $f(\mathbf{U}_T)$ and $f(\mathbf{U}_S)$, and $\mathbf{C}_T$ and $\mathbf{C}_S$ are the covariance matrices. We normalize the features to have zero mean ($\mu_T$ = $\mu_S$ = 0). Then, CORAL minimizes the distance between $f(\mathbf{U}_T)$ and $f(\mathbf{U}_S)$ by a linear transformation $\mathbf{A}$ by
\begin{equation}
    \min_{\mathbf{A}} ||\mathbf{C}_{\hat{S}} - \mathbf{C}_T||_F^2 = \min_{\mathbf{A}} ||\mathbf{A}^{T}\mathbf{C}_{T}\mathbf{A} - \mathbf{C}_T||_F^2,
\end{equation}
where $||\cdot||_F^2$ is the squared Frobenius norm, and $\mathbf{C}_{\hat{S}}$ is the covariance of the transformed source features $f(\mathbf{U}_S)\mathbf{A}$ \cite{sun_feng}. The optimal solution of this problem is given by
\begin{equation}
    \mathbf{A}^* = \mathbf{P}_S {\mathbf{\Sigma}_S^+}^\frac{1}{2} \mathbf{P}_S^T {\mathbf{P}_T}_{[1:r]} {\mathbf{\Sigma}_T}_{[1:r]}^{\frac{1}{2}} {\mathbf{P}_T}_{[1:r]}^T,
\end{equation}
with $r = \min(r_{\mathbf{C}_S}, r_{\mathbf{C}_T})$, where $r_{\mathbf{C}_S}$ and $r_{\mathbf{C}_T}$ denote the rank of $\mathbf{C}_S$ and $\mathbf{C}_T$, and $\mathbf{\Sigma}^+$ is the Moore-Penrose pseudoinverse of $\mathbf{\Sigma}$. We use the singular value decomposition of a real matrix $\mathbf{Y}$ to compute the largest $r \leq r_Y$ singular values $\mathbf{\Sigma}_{Y_{[1:r]}}$, and left and right singular vectors $\mathbf{P}_{Y_{[1:r]}}$ and $\mathbf{V}_{Y_{[1:r]}}$ of $\mathbf{Y} = \mathbf{P}_Y \mathbf{\Sigma}_Y \mathbf{V}_Y$ of the real matrix $\mathbf{Y}$ of rank $r_Y$ \cite{cai_candes}.

\subsection{Datasets Overview}
\label{sec_app_datasets}

\begin{table}[b!]
\begin{center}
\setlength{\tabcolsep}{3.7pt}
    \caption{Overview of sample numbers of online handwriting (OnHW) recognition datasets for writer-dependent (WD) and writer-independent (WI) and right- and left-handed classification tasks. Top line: training. Bottom line: validation.}
    \label{table_app_onhw_1}
    \vspace{-0.2cm}
    \small \begin{tabular}{ p{0.5cm} | p{0.5cm} | p{0.5cm} | p{0.5cm} | p{0.5cm} | p{0.5cm} }
    \multicolumn{2}{c|}{\textbf{Dataset}} & \multicolumn{2}{c|}{\textbf{Right-handed}} & \multicolumn{2}{c}{\textbf{Left-handed}} \\
    \multicolumn{2}{l|}{} & \multicolumn{1}{c|}{\textbf{WD}} & \multicolumn{1}{c|}{\textbf{WI}} & \multicolumn{1}{c|}{\textbf{WD}} & \multicolumn{1}{c}{\textbf{WI}} \\ \hline
    \multicolumn{2}{l|}{\textbf{OnHW-symbols~\cite{ott_tist}}} & \multicolumn{1}{r|}{1,853} & \multicolumn{1}{r|}{1,715} & \multicolumn{1}{r|}{288} & \multicolumn{1}{r}{267} \\
    \multicolumn{2}{l|}{} & \multicolumn{1}{r|}{473} & \multicolumn{1}{r|}{611} & \multicolumn{1}{r|}{73} & \multicolumn{1}{r}{94} \\ \hline
    \multicolumn{2}{l|}{\textbf{Split OnHW-equations~\cite{ott_tist}}} & \multicolumn{1}{r|}{31,697} & \multicolumn{1}{r|}{30,408} & \multicolumn{1}{r|}{5,021} & \multicolumn{1}{r}{4,579} \\
    \multicolumn{2}{l|}{} & \multicolumn{1}{r|}{7,946} & \multicolumn{1}{r|}{9,235} & \multicolumn{1}{r|}{1,259} & \multicolumn{1}{r}{1,701} \\ \hline
    \multicolumn{1}{l|}{} & \multicolumn{1}{l|}{\textbf{lower}} & \multicolumn{1}{r|}{11,524} & \multicolumn{1}{r|}{11,647} & \multicolumn{1}{r|}{903} & \multicolumn{1}{r}{781} \\
    \multicolumn{1}{l|}{} & \multicolumn{1}{l|}{} & \multicolumn{1}{r|}{4,101} & \multicolumn{1}{r|}{3,978} & \multicolumn{1}{r|}{218} & \multicolumn{1}{r}{368} \\
    \multicolumn{1}{l|}{\textbf{OnHW-chars~\cite{ott}}} & \multicolumn{1}{l|}{\textbf{upper}} & \multicolumn{1}{r|}{11,542} & \multicolumn{1}{r|}{11,672} & \multicolumn{1}{r|}{925} & \multicolumn{1}{r}{757} \\
    \multicolumn{1}{l|}{} & \multicolumn{1}{l|}{} & \multicolumn{1}{r|}{4,108} & \multicolumn{1}{r|}{3,978} & \multicolumn{1}{r|}{224} & \multicolumn{1}{r}{364} \\
    \multicolumn{1}{l|}{} & \multicolumn{1}{l|}{\textbf{combined}} & \multicolumn{1}{r|}{23,066} & \multicolumn{1}{r|}{23,319} & \multicolumn{1}{r|}{1,821} & \multicolumn{1}{r}{1,538} \\
    \multicolumn{1}{l|}{} & \multicolumn{1}{l|}{} & \multicolumn{1}{r|}{8,209} & \multicolumn{1}{r|}{7,956} & \multicolumn{1}{r|}{449} & \multicolumn{1}{r}{732} \\
    \end{tabular}
\end{center}
\end{table}

\begin{table}[t!]
\begin{center}
\setlength{\tabcolsep}{3.7pt}
    \caption{Overview of number of samples for each left-handed writer for the OnHW datasets for train/validation splits. For OnHW-chars, we count for lower/upper/combined.}
    \label{table_app_onhw_2}
    \vspace{-0.2cm}
    \small \begin{tabular}{ p{0.5cm} | p{0.5cm} | p{0.5cm} | p{0.5cm} }
    \multicolumn{1}{c|}{\textbf{Dataset}} & \multicolumn{1}{c|}{\textbf{Writer}} & \multicolumn{1}{c|}{\textbf{Train} ($\mathcal{U}_{S_t}$)} & \multicolumn{1}{c}{\textbf{Val.} ($\mathcal{U}_{S_v}$)} \\ \hline
    \multicolumn{1}{l|}{\textbf{OnHW-symbols-L~\cite{ott_tist}}} & \multicolumn{1}{c|}{\textbf{1}} & \multicolumn{1}{r|}{71} & \multicolumn{1}{r}{18} \\
    & \multicolumn{1}{c|}{\textbf{2}} & \multicolumn{1}{r|}{72} & \multicolumn{1}{r}{18} \\
    & \multicolumn{1}{c|}{\textbf{3}} & \multicolumn{1}{r|}{70} & \multicolumn{1}{r}{18} \\
    & \multicolumn{1}{c|}{\textbf{4}} & \multicolumn{1}{r|}{75} & \multicolumn{1}{r}{19} \\ \hline
    \multicolumn{1}{l|}{\textbf{Split OnHW-}} & \multicolumn{1}{c|}{\textbf{1}} & \multicolumn{1}{r|}{1,299} & \multicolumn{1}{r}{327} \\
    \multicolumn{1}{l|}{\textbf{equations-L~\cite{ott_tist}}}& \multicolumn{1}{c|}{\textbf{2}} & \multicolumn{1}{r|}{1,067} & \multicolumn{1}{r}{267} \\
    & \multicolumn{1}{c|}{\textbf{3}} & \multicolumn{1}{r|}{1,295} & \multicolumn{1}{r}{324} \\
    & \multicolumn{1}{c|}{\textbf{4}} & \multicolumn{1}{r|}{1,360} & \multicolumn{1}{r}{341} \\ \hline
    \multicolumn{1}{l|}{\textbf{OnHW-chars-L~\cite{ott}}} & \multicolumn{1}{c|}{\textbf{1}} & \multicolumn{1}{r|}{78 /\,\,\, 89 / 167} & \multicolumn{1}{r}{26 / 15 /\,\,\, 41} \\
    \multicolumn{1}{l|}{} & \multicolumn{1}{c|}{\textbf{2}} & \multicolumn{1}{r|}{78 /\,\,\, 89 / 167} & \multicolumn{1}{r}{26 / 15 /\,\,\, 41} \\
    \multicolumn{1}{l|}{} & \multicolumn{1}{c|}{\textbf{3}} & \multicolumn{1}{r|}{78 /\,\,\, 70 / 148} & \multicolumn{1}{r}{26 / 11 /\,\,\, 37} \\
    \multicolumn{1}{l|}{} & \multicolumn{1}{c|}{\textbf{4}} & \multicolumn{1}{r|}{78 /\,\,\, 89 / 167} & \multicolumn{1}{r}{26 / 15 /\,\,\, 41} \\
    \multicolumn{1}{l|}{} & \multicolumn{1}{c|}{\textbf{5}} & \multicolumn{1}{r|}{79 /\,\,\, 89 / 168} & \multicolumn{1}{r}{26 / 15 /\,\,\, 41} \\
    \multicolumn{1}{l|}{} & \multicolumn{1}{c|}{\textbf{6}} & \multicolumn{1}{r|}{202 / 218 / 420} & \multicolumn{1}{r}{62 / 42 / 104} \\
    \multicolumn{1}{l|}{} & \multicolumn{1}{c|}{\textbf{7}} & \multicolumn{1}{r|}{78 /\,\,\, 89 / 167} & \multicolumn{1}{r}{26 / 15 /\,\,\, 41} \\
    \multicolumn{1}{l|}{} & \multicolumn{1}{c|}{\textbf{8}} & \multicolumn{1}{r|}{78 /\,\,\, 89 / 167} & \multicolumn{1}{r}{26 / 15 /\,\,\, 41} \\
    \multicolumn{1}{l|}{} & \multicolumn{1}{c|}{\textbf{9}} & \multicolumn{1}{r|}{120 / 130 / 250} & \multicolumn{1}{r}{360 / 26 /\,\,\, 386} \\
    \end{tabular}
\end{center}
\end{table}

Table~\ref{table_app_onhw_1} gives an overview of sample counts for the right-handed target domains for training $\mathcal{U}_{T_t}$ and validation $\mathcal{U}_{T_v}$. Results for these datasets are given in Table~\ref{table_baseline_symbols} and \ref{table_baseline_chars}. Here, the left-handed dataset comprise all writers. Table~\ref{table_app_onhw_2} shows the sample numbers for left-handed writers separated for each writer at an 80/20 training validation split. We propose counts for the source domains for training $\mathcal{U}_{S_t}$ and validation $\mathcal{U}_{S_v}$ data. Results are averaged over all writers (four for the OnHW-symbols and split OnHW-equations datasets, and nine for the OnHW-chars dataset) and are given in Table~\ref{table_da_results_all} for domain adaptation and in Table~\ref{table_post_training} for transfer learning.

\subsection{Feature Embeddings for the Sinusoidal Dataset}
\label{sec_app_embeddings}

\newcommand\widthsinus{0.196}
\begin{figure*}[t!]
	\centering
	\begin{minipage}[b]{\widthsinus\linewidth}
        \centering
    	\includegraphics[trim=14 13 10 10, clip, width=1.0\linewidth]{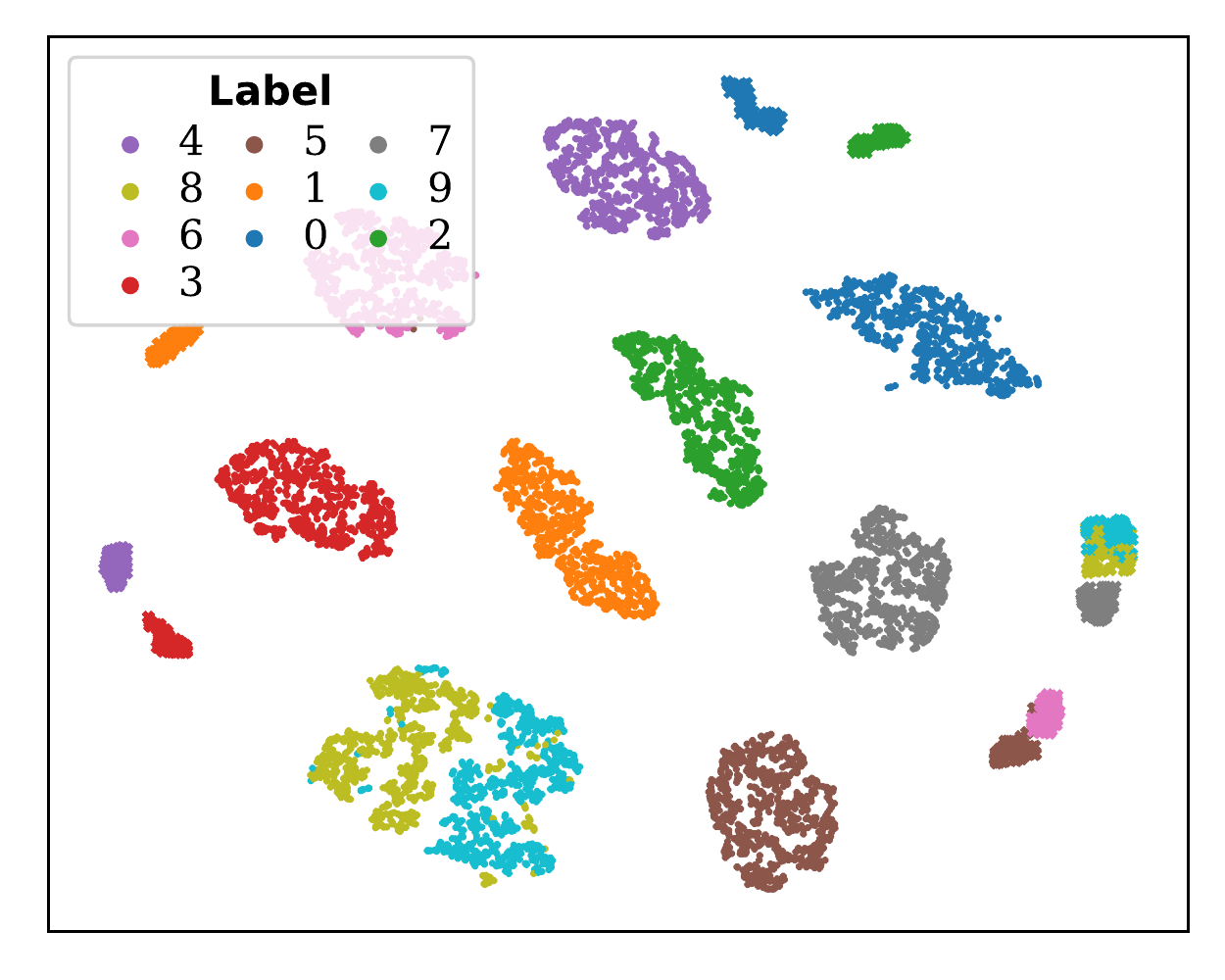}
    	\subcaption{$b=0.0$.}
    	\label{image_app_sin_cos_class_a}
    \end{minipage}
    \hfill
	\begin{minipage}[b]{\widthsinus\linewidth}
        \centering
    	\includegraphics[trim=14 13 10 10, clip, width=1.0\linewidth]{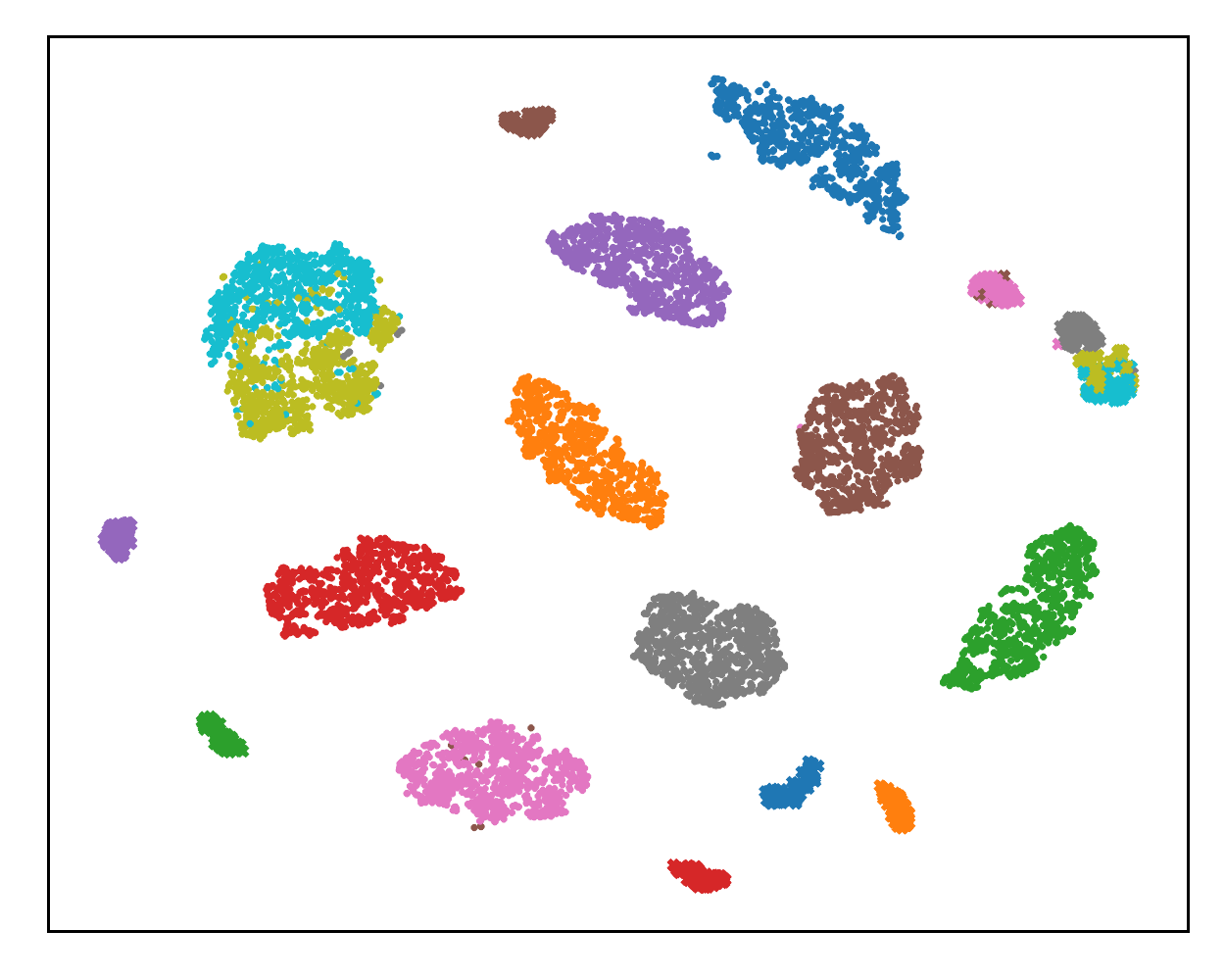}
    	\subcaption{$b=0.5$.}
    	\label{image_app_sin_cos_class_b}
    \end{minipage}
    \hfill
	\begin{minipage}[b]{\widthsinus\linewidth}
        \centering
    	\includegraphics[trim=14 13 10 10, clip, width=1.0\linewidth]{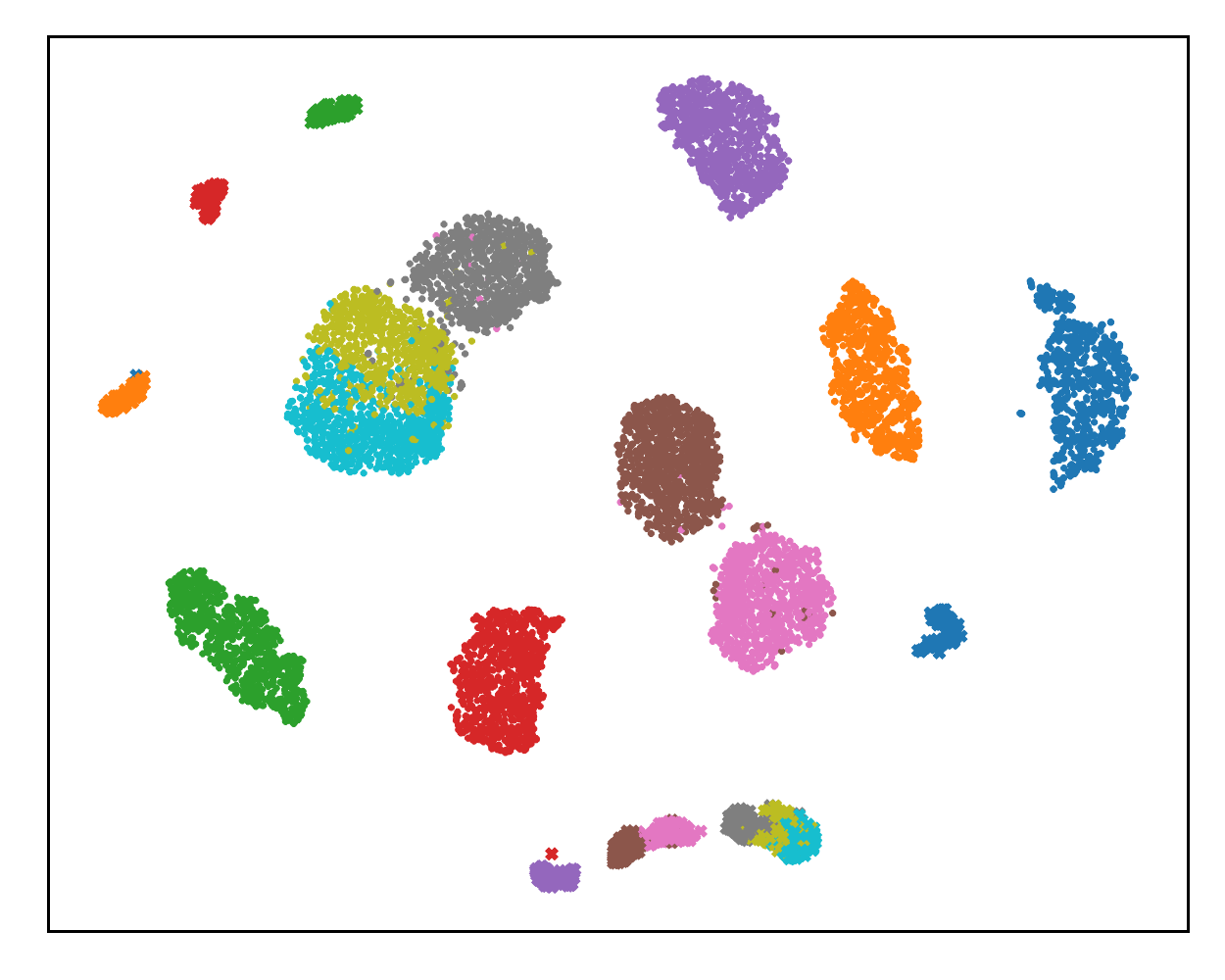}
    	\subcaption{$b=1.0$.}
    	\label{image_app_sin_cos_class_c}
    \end{minipage}
    \hfill
	\begin{minipage}[b]{\widthsinus\linewidth}
        \centering
    	\includegraphics[trim=14 13 10 10, clip, width=1.0\linewidth]{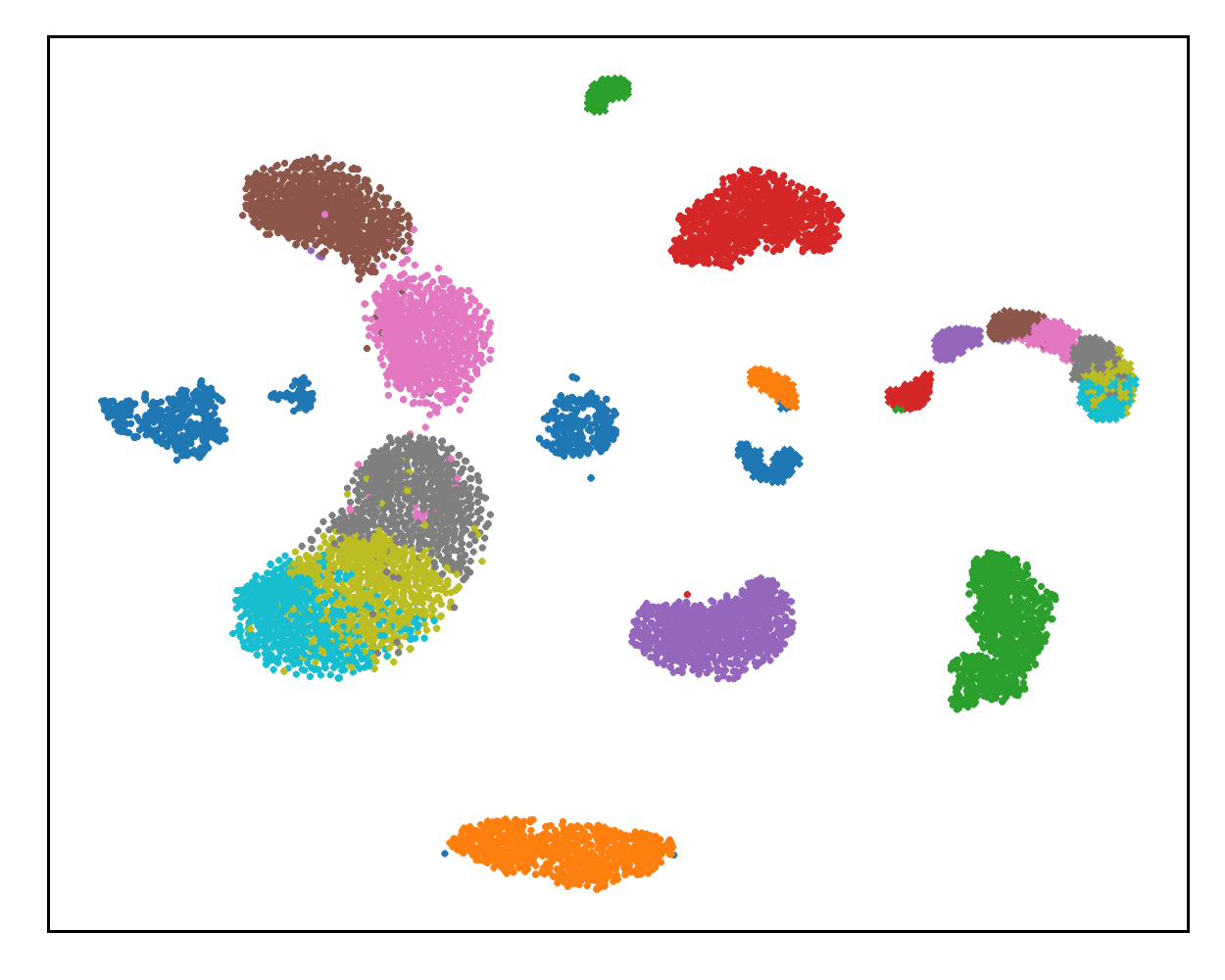}
    	\subcaption{$b=1.5$.}
    	\label{image_app_sin_cos_class_d}
    \end{minipage}
    \hfill
	\begin{minipage}[b]{\widthsinus\linewidth}
        \centering
    	\includegraphics[trim=14 13 10 10, clip, width=1.0\linewidth]{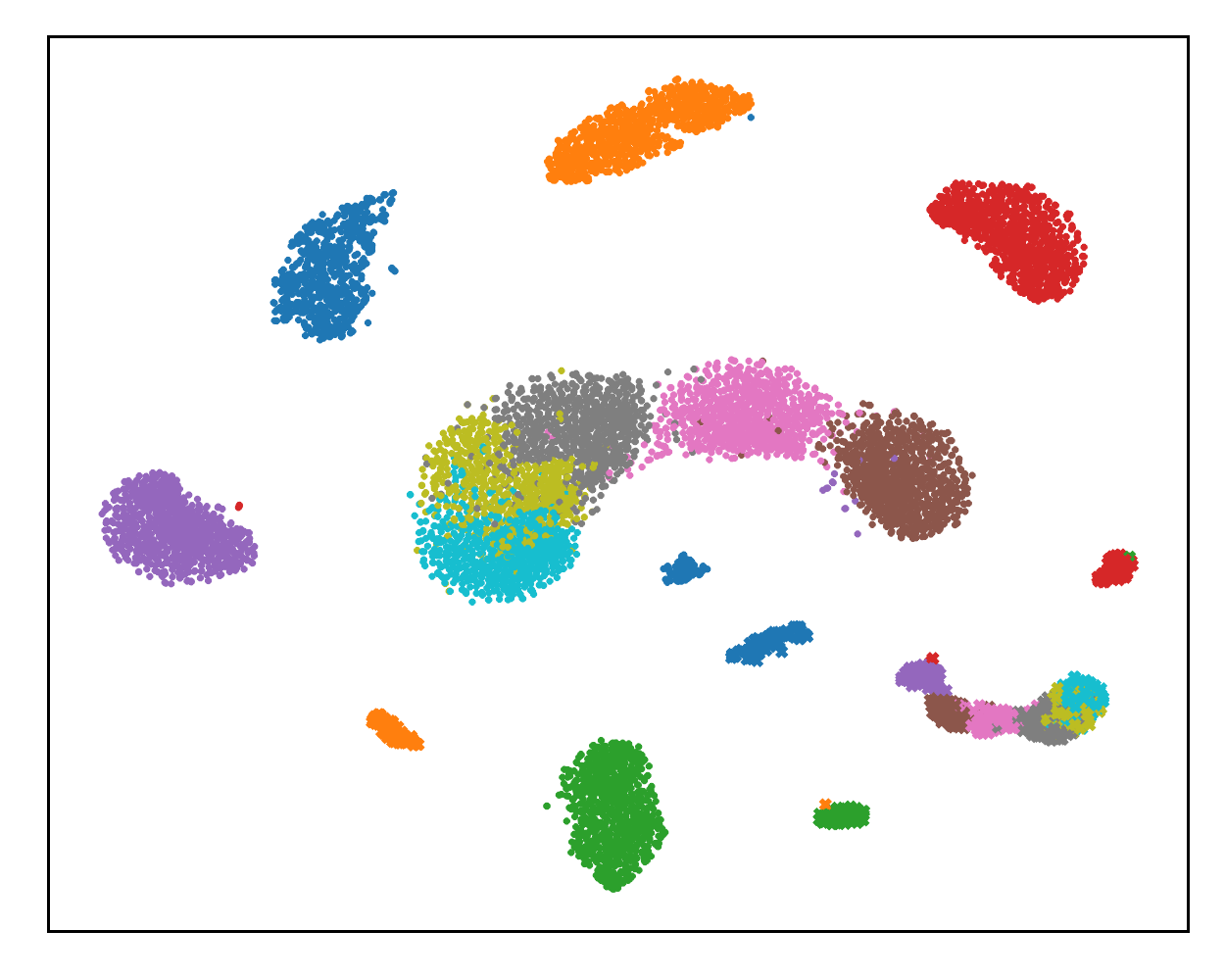}
    	\subcaption{$b=1.9$.}
    	\label{image_app_sin_cos_class_e}
    \end{minipage}
    \caption{Embedding visualization for the sinusoidal datasets for the target domain $\mathcal{U}_T$ and the source domain $\mathcal{U}_{S_v}$ with noise parameters $b \in B = \{0.0, 0.5, 1.0, 1.5, 1.9\}$. Marker $\cdot$: target domain embeddings. Marker $\times$: source domain validation embeddings.}
    \label{image_app_sin_cos_class}
\end{figure*}

\newcommand\widthimages{0.246}
\begin{figure*}[t!]
	\centering
	\begin{minipage}[b]{\widthimages\linewidth}
        \centering
    	\includegraphics[trim=12 12 10 10, clip, width=1.0\linewidth]{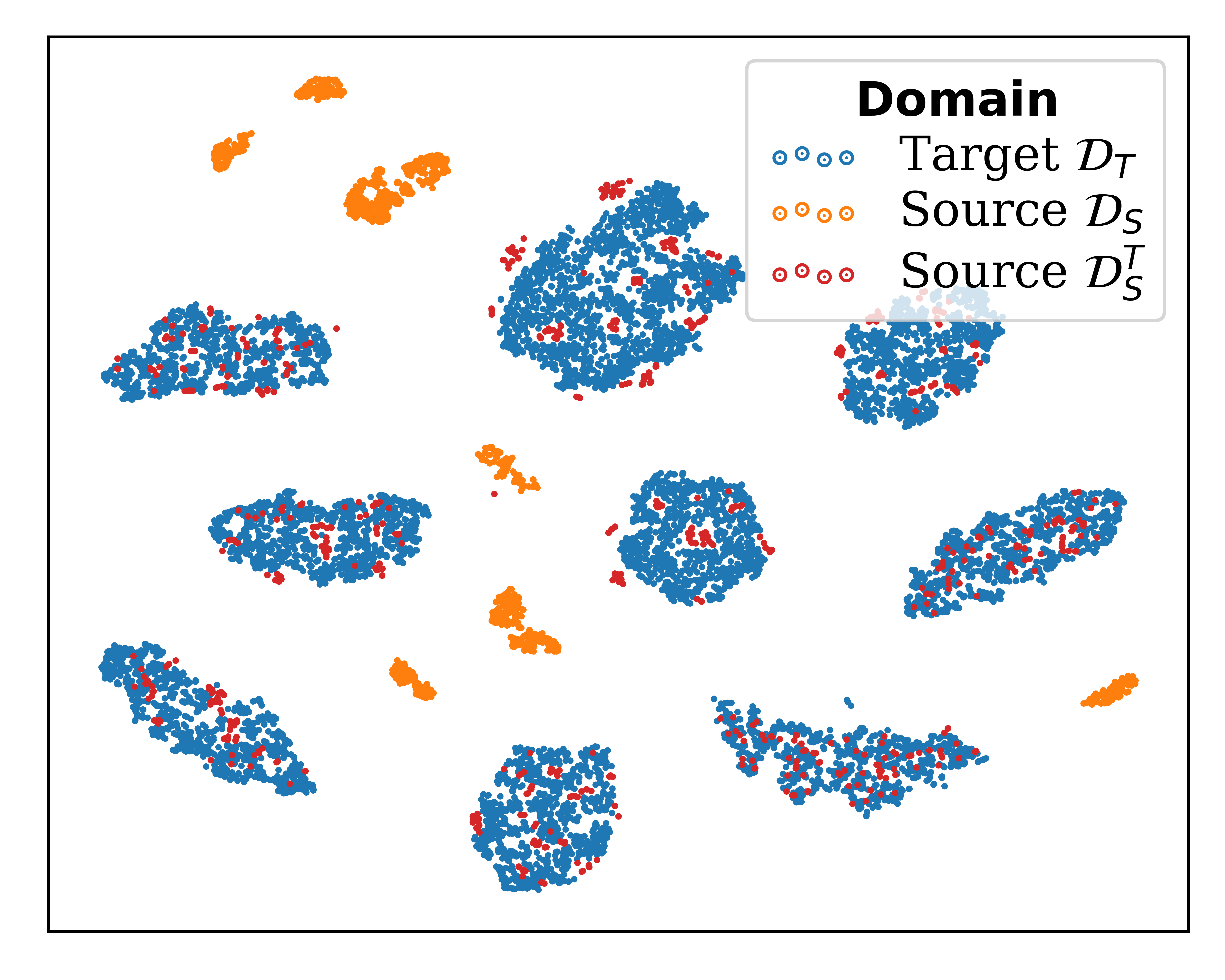}
    	\subcaption{EMD, best transformation.}
    	\label{image_app_emb_1}
    \end{minipage}
    \hfill
	\begin{minipage}[b]{\widthimages\linewidth}
        \centering
    	\includegraphics[trim=12 12 10 10, clip, width=1.0\linewidth]{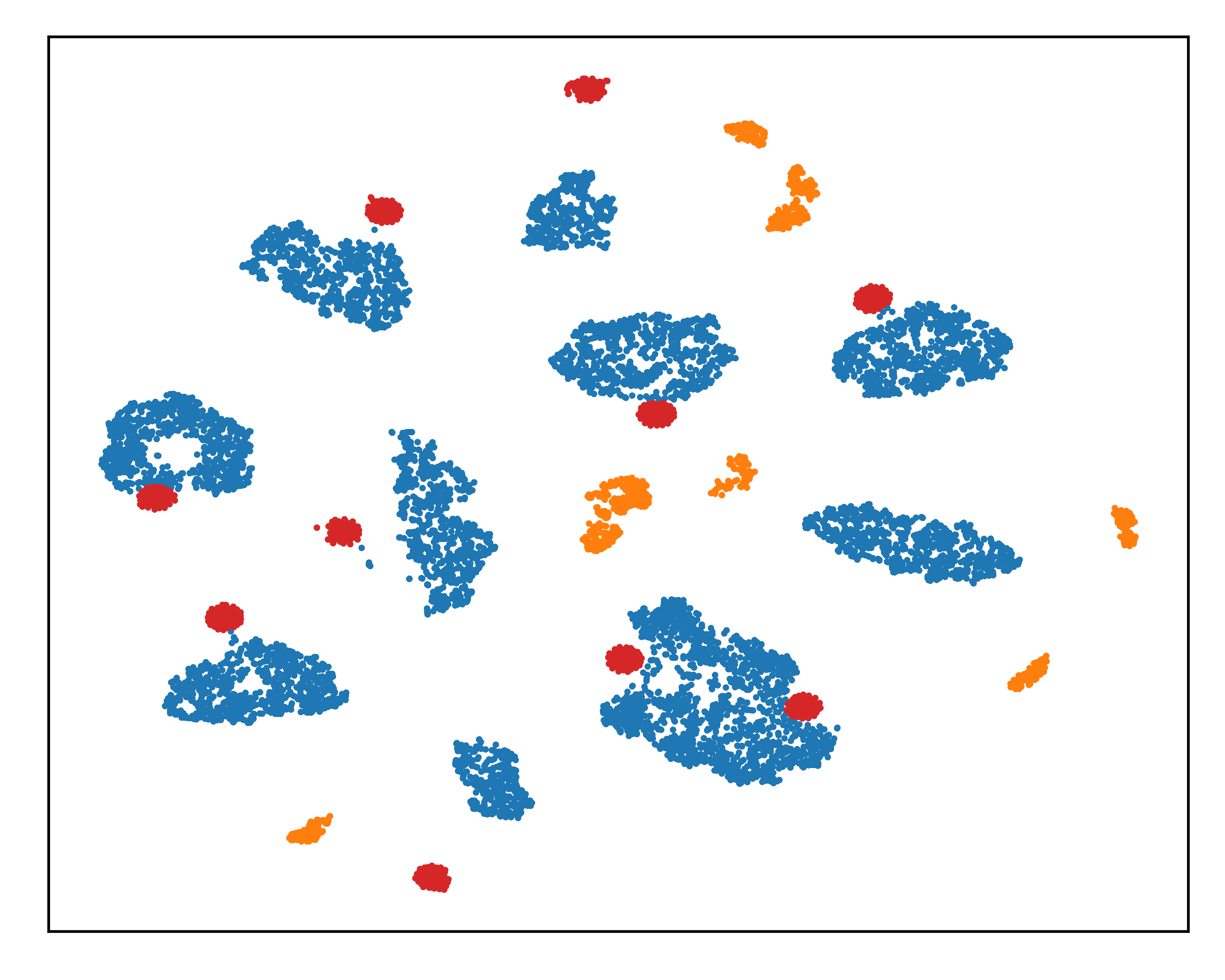}
    	\subcaption{Sinkhorn, best transformation.}
    	\label{image_app_emb_2}
    \end{minipage}
    \hfill
	\begin{minipage}[b]{\widthimages\linewidth}
        \centering
    	\includegraphics[trim=12 12 10 10, clip, width=1.0\linewidth]{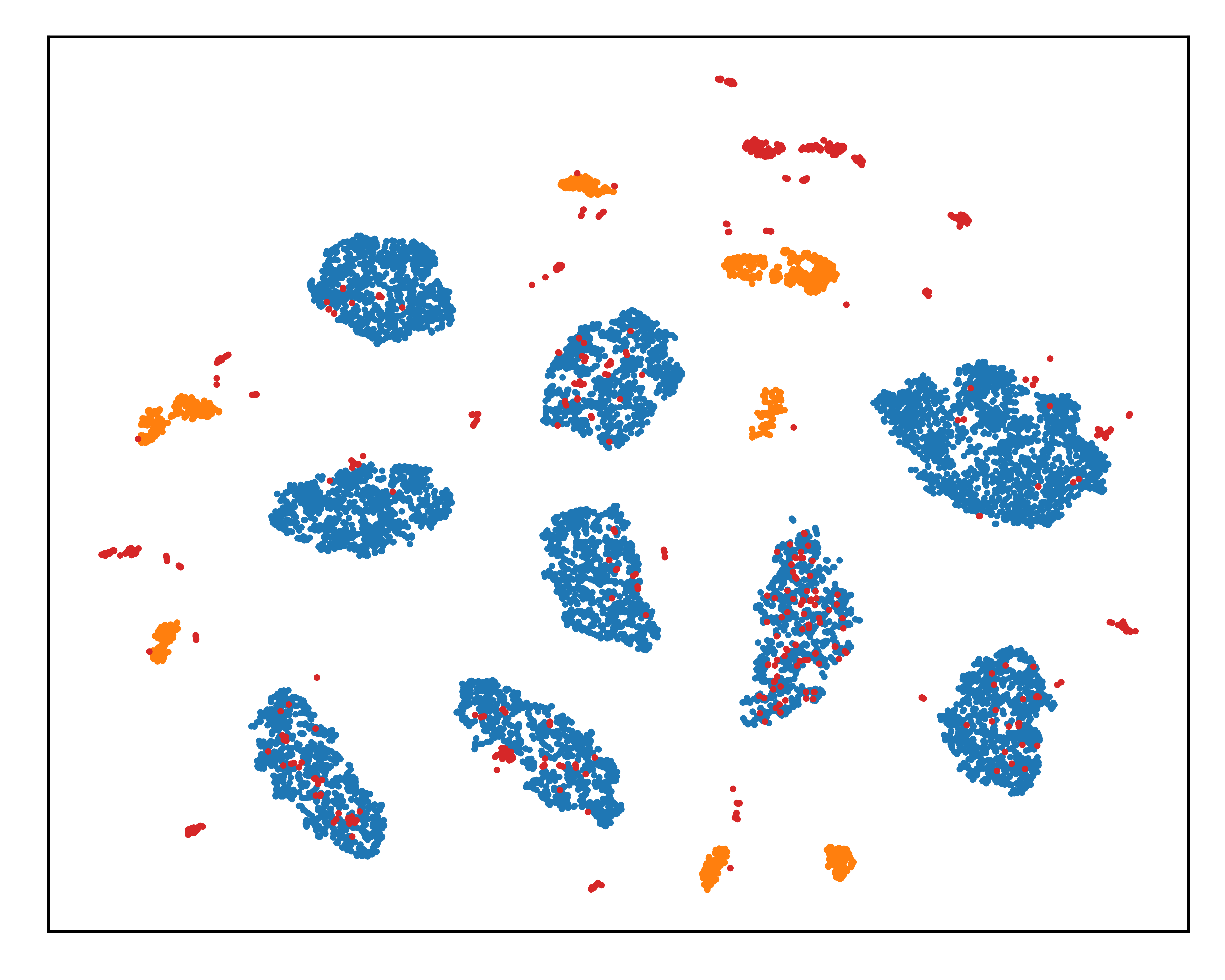}
    	\subcaption{EMD, MMD.}
    	\label{image_app_emb_3}
    \end{minipage}
    \hfill
	\begin{minipage}[b]{\widthimages\linewidth}
        \centering
    	\includegraphics[trim=12 12 10 10, clip, width=1.0\linewidth]{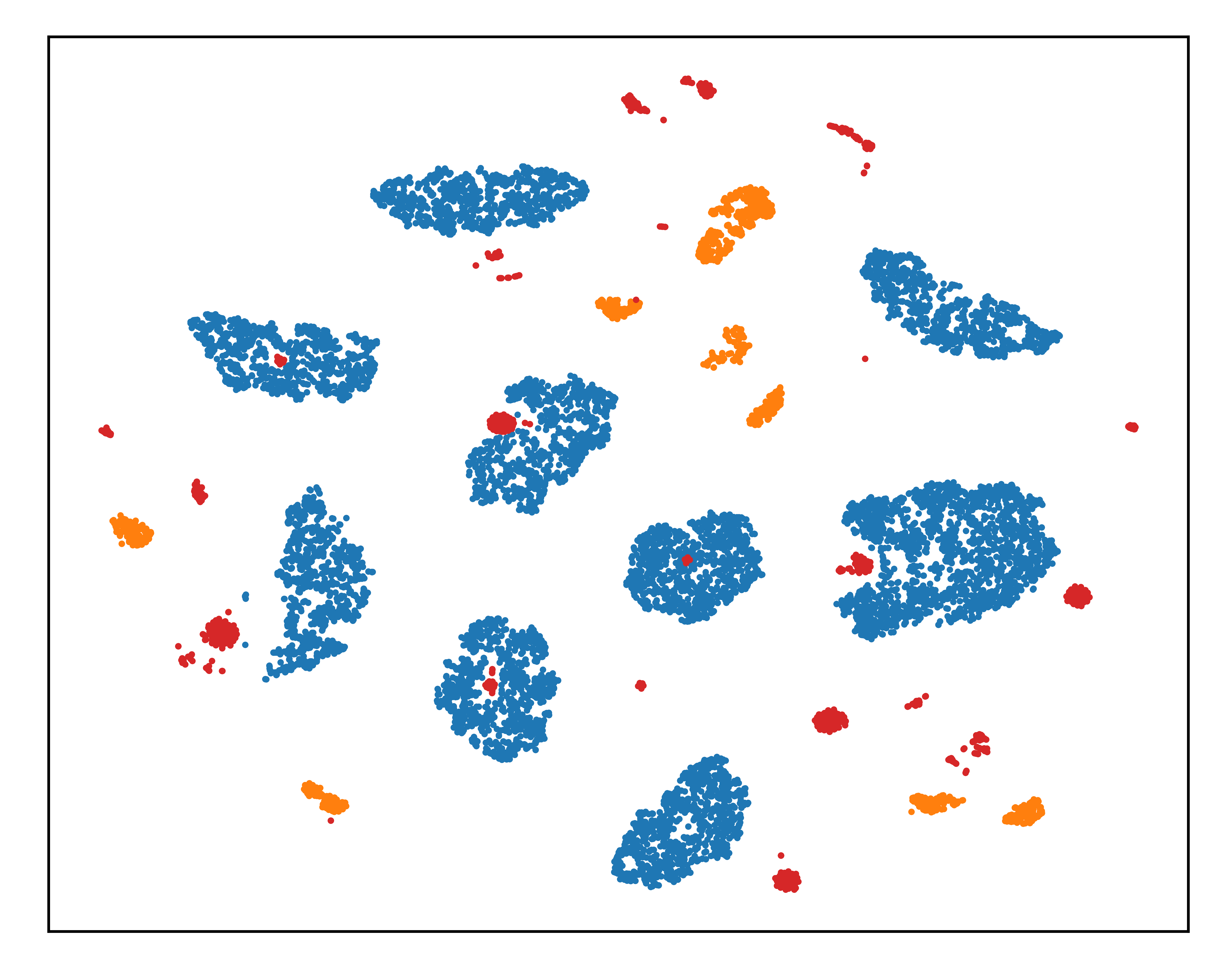}
    	\subcaption{Sinkhorn, MMD.}
    	\label{image_app_emb_4}
    \end{minipage}
    \caption{Embedding visualization for the target domain $\mathcal{U}_T$, source domain $\mathcal{U}_S$ and transformed source domain $\mathcal{U}_S^T$ for the sinusoidal datasets. We show embeddings for EMD and Sinkhorn transport without regularization. Evaluated are nine metrics for transformation selection: CC, PC, CORAL (standard, Jeff and Stein), correlation normalized, MMD, kMMD, HoMM, without transformation and best transformation. Noise is $b=0.5$. Note that we applied t-SNE for each plot separately that leads to different embeddings for the same data (i.e., target domain $\mathcal{D}_T$). Figure~\ref{image_app_emb_follow} follows.}
    \label{image_app_emb}
\end{figure*}

\begin{figure*}[t!]
	\centering
	\begin{minipage}[b]{\widthimages\linewidth}
        \centering
    	\includegraphics[trim=12 12 10 10, clip, width=1.0\linewidth]{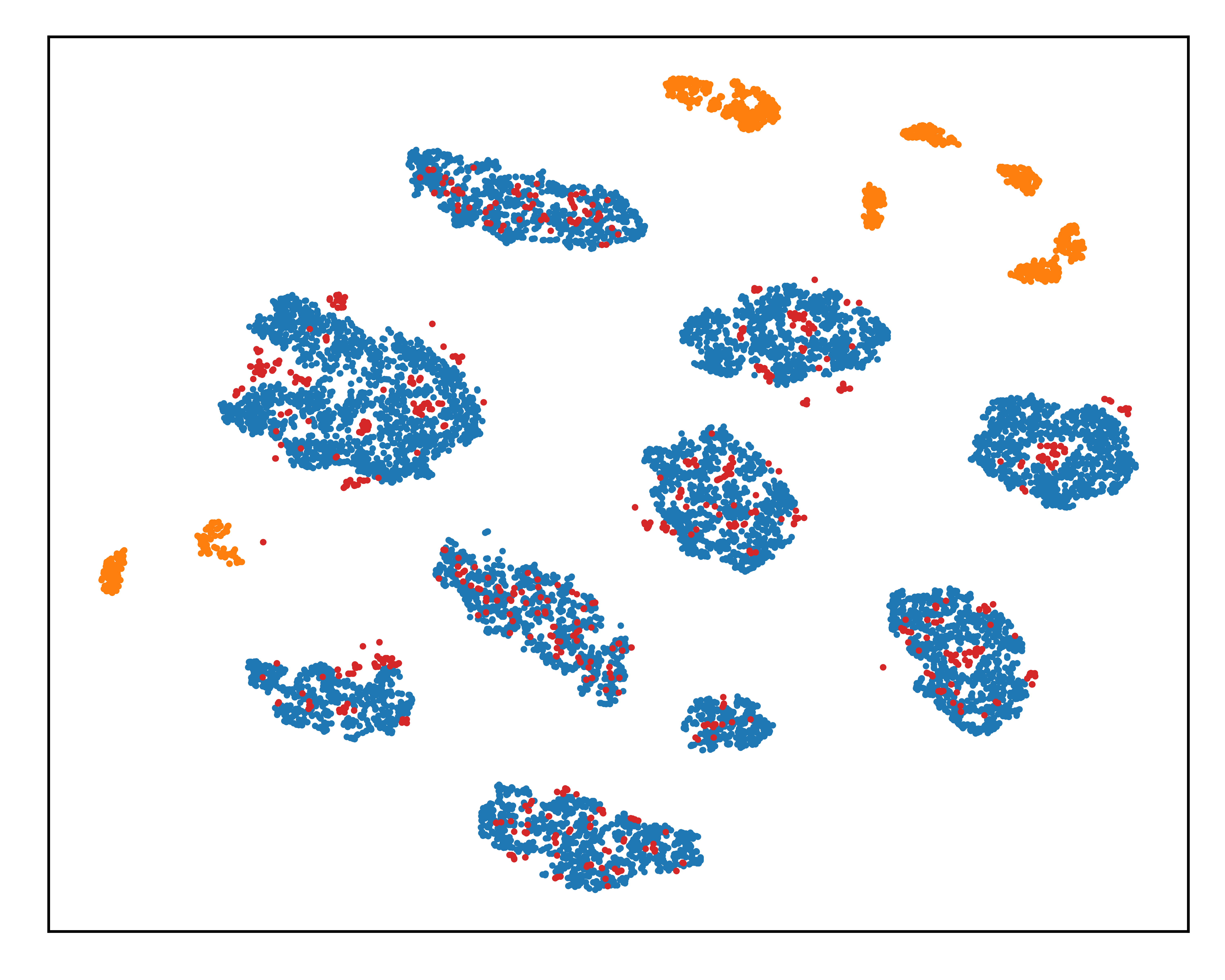}
    	\subcaption{EMD, kMMD.}
    	\label{image_app_emb_5}
    \end{minipage}
    \hfill
	\begin{minipage}[b]{\widthimages\linewidth}
        \centering
    	\includegraphics[trim=12 12 10 10, clip, width=1.0\linewidth]{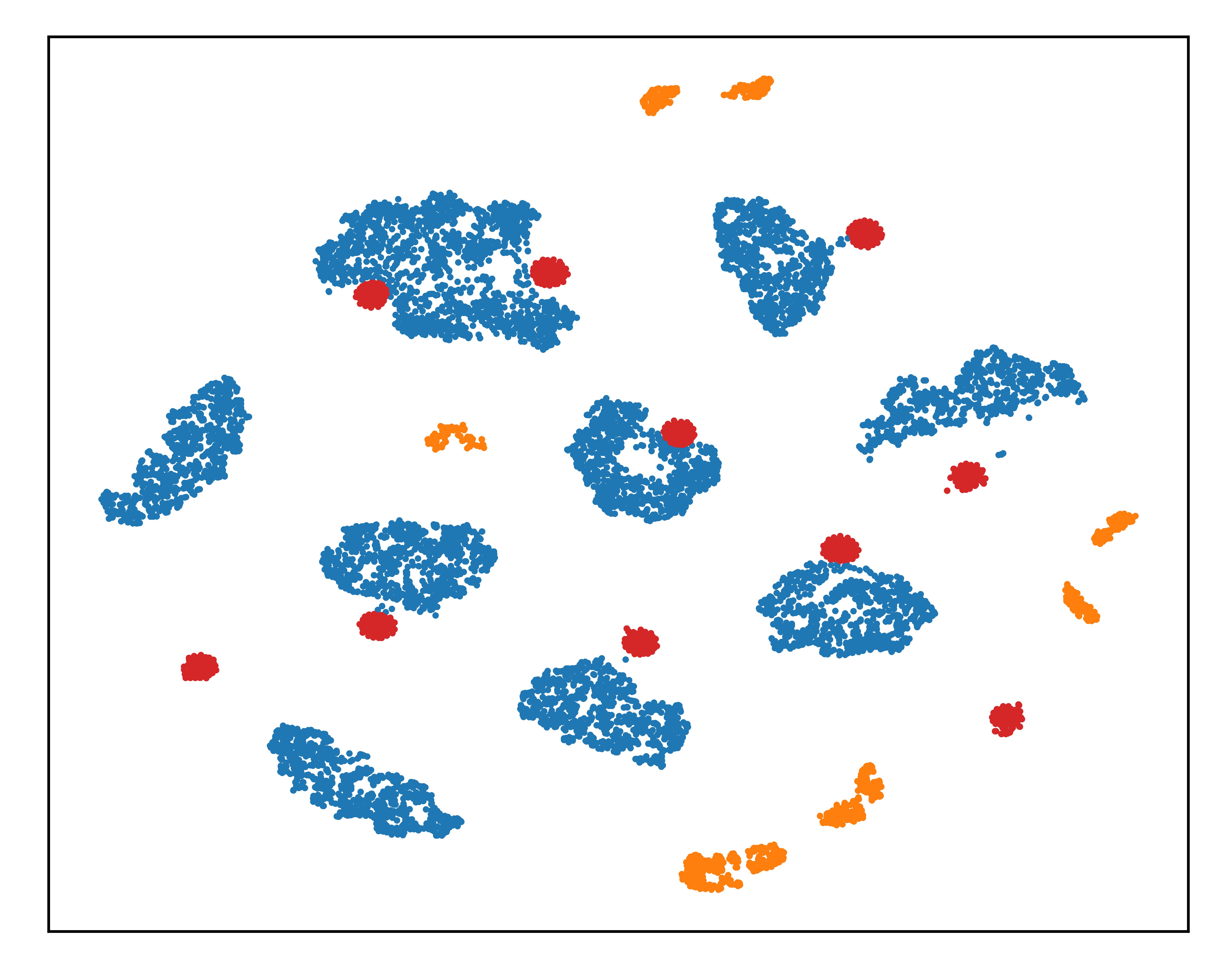}
    	\subcaption{Sinkhorn, kMMD.}
    	\label{image_app_emb_6}
    \end{minipage}
    \hfill
	\begin{minipage}[b]{\widthimages\linewidth}
        \centering
    	\includegraphics[trim=12 12 10 10, clip, width=1.0\linewidth]{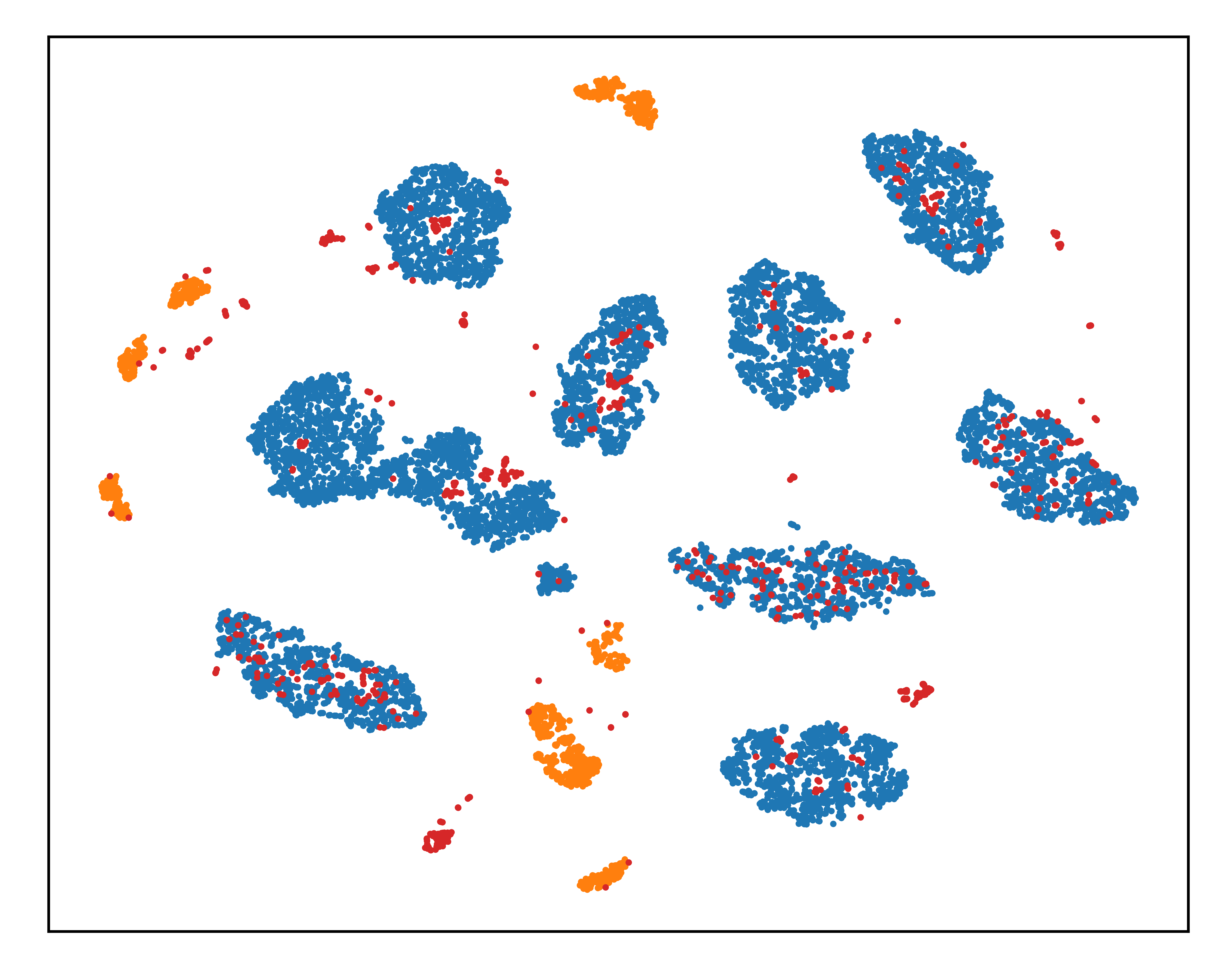}
    	\subcaption{EMD, HoMM.}
    	\label{image_app_emb_7}
    \end{minipage}
    \hfill
	\begin{minipage}[b]{\widthimages\linewidth}
        \centering
    	\includegraphics[trim=12 12 10 10, clip, width=1.0\linewidth]{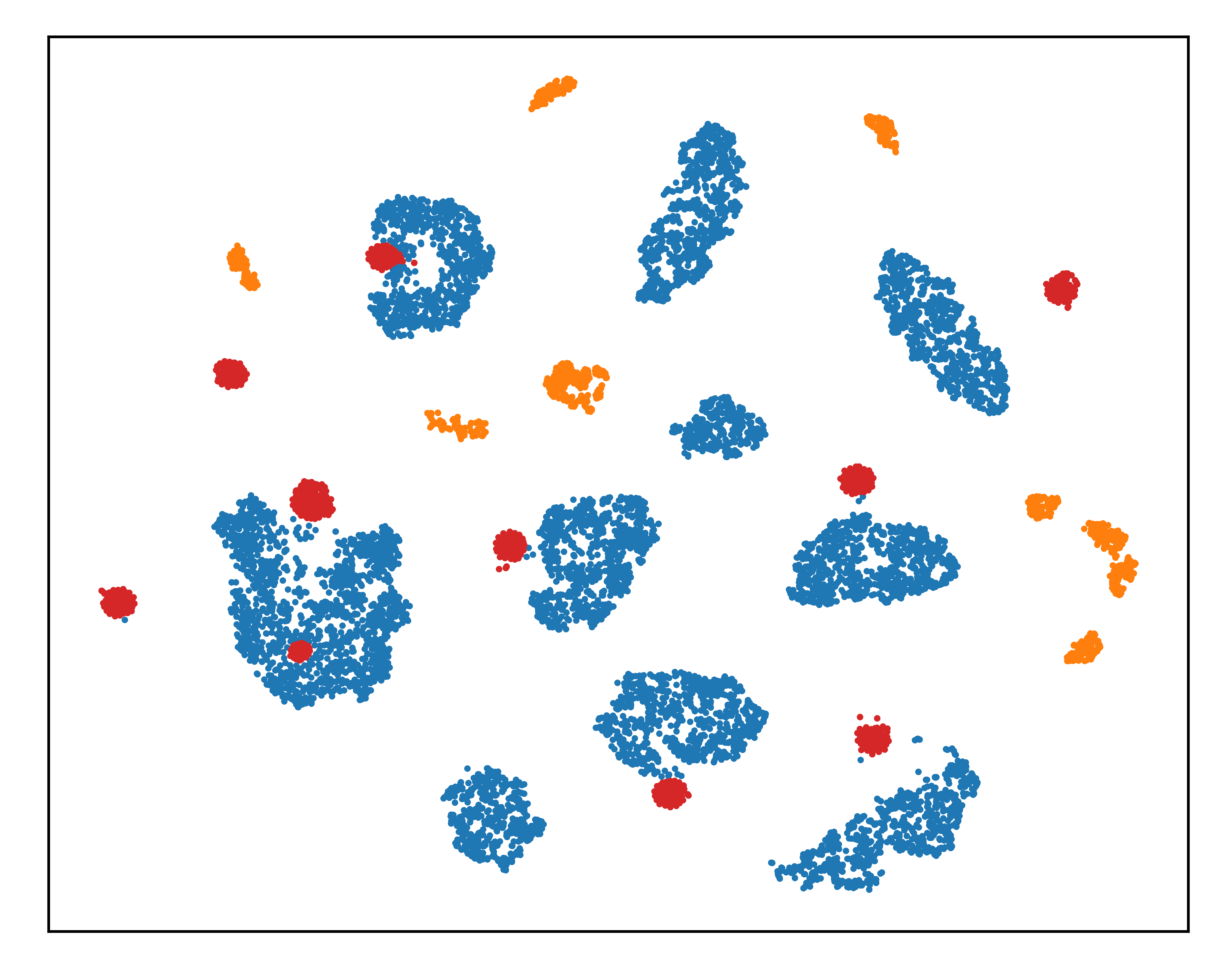}
    	\subcaption{Sinkhorn, HoMM.}
    	\label{image_app_emb_8}
    \end{minipage}
	\begin{minipage}[b]{\widthimages\linewidth}
        \centering
    	\includegraphics[trim=12 12 10 10, clip, width=1.0\linewidth]{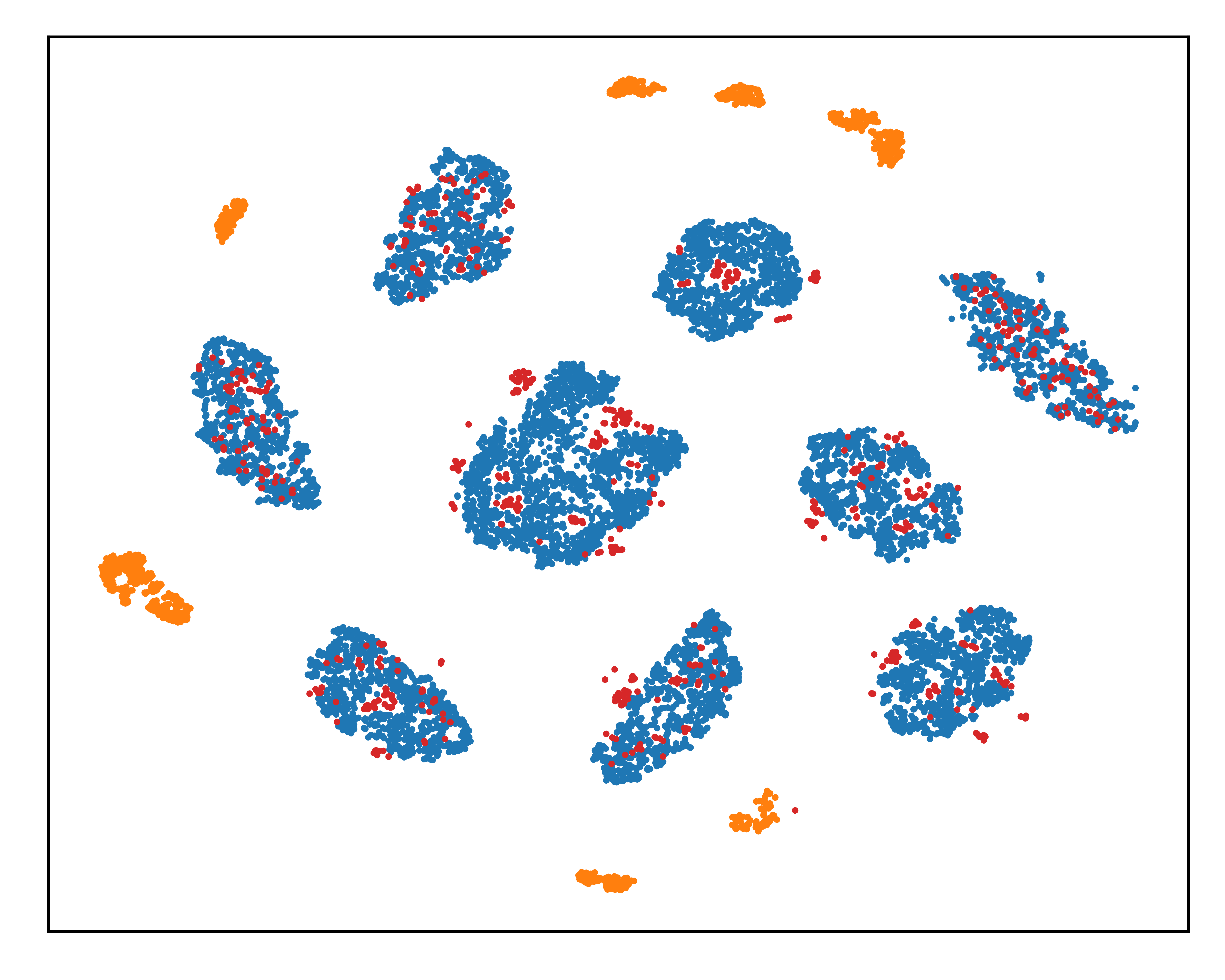}
    	\subcaption{EMD, CC.}
    	\label{image_app_emb_9}
    \end{minipage}
    \hfill
	\begin{minipage}[b]{\widthimages\linewidth}
        \centering
    	\includegraphics[trim=12 12 10 10, clip, width=1.0\linewidth]{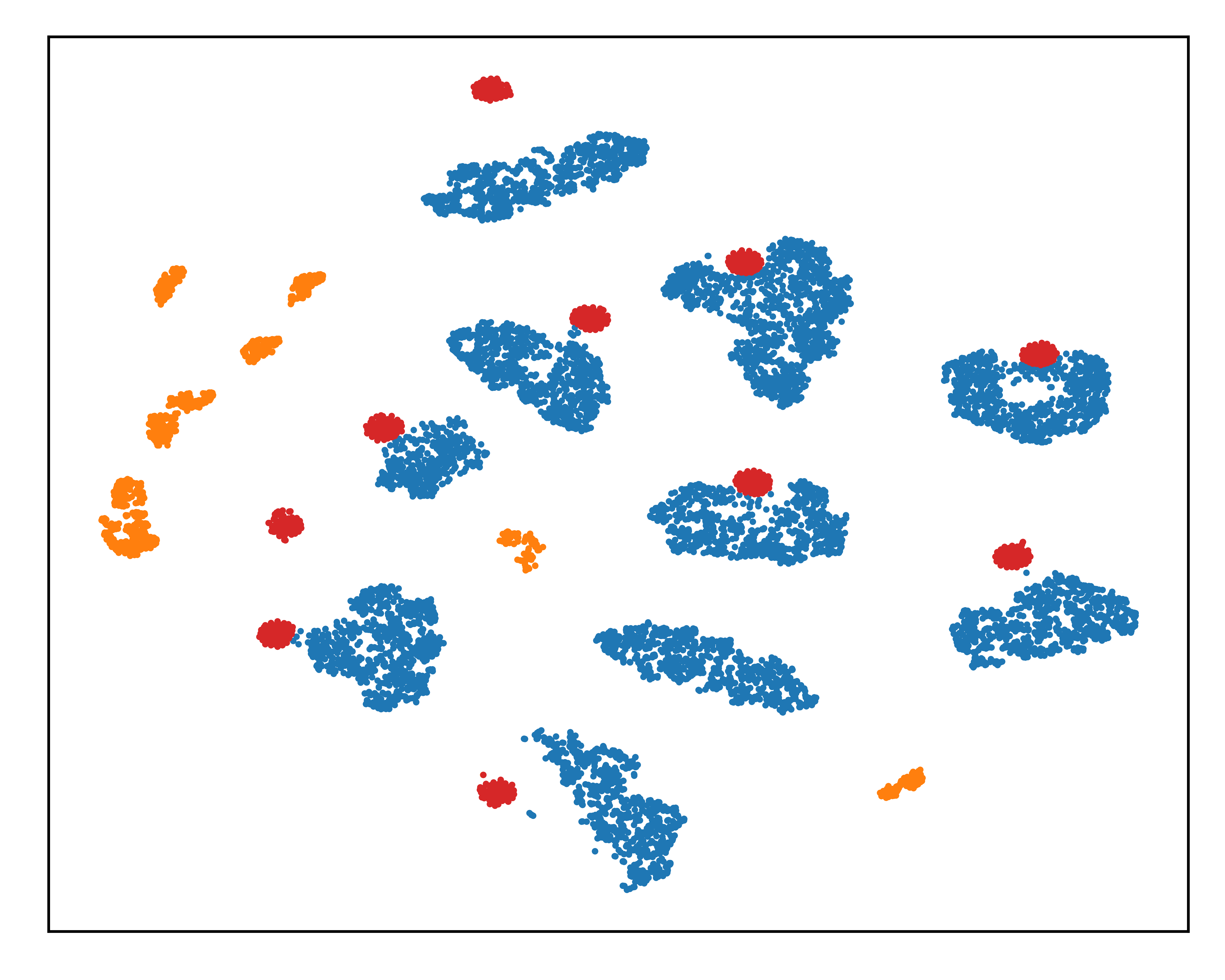}
    	\subcaption{Sinkhorn, CC.}
    	\label{image_app_emb_10}
    \end{minipage}
    \hfill
	\begin{minipage}[b]{\widthimages\linewidth}
        \centering
    	\includegraphics[trim=12 12 10 10, clip, width=1.0\linewidth]{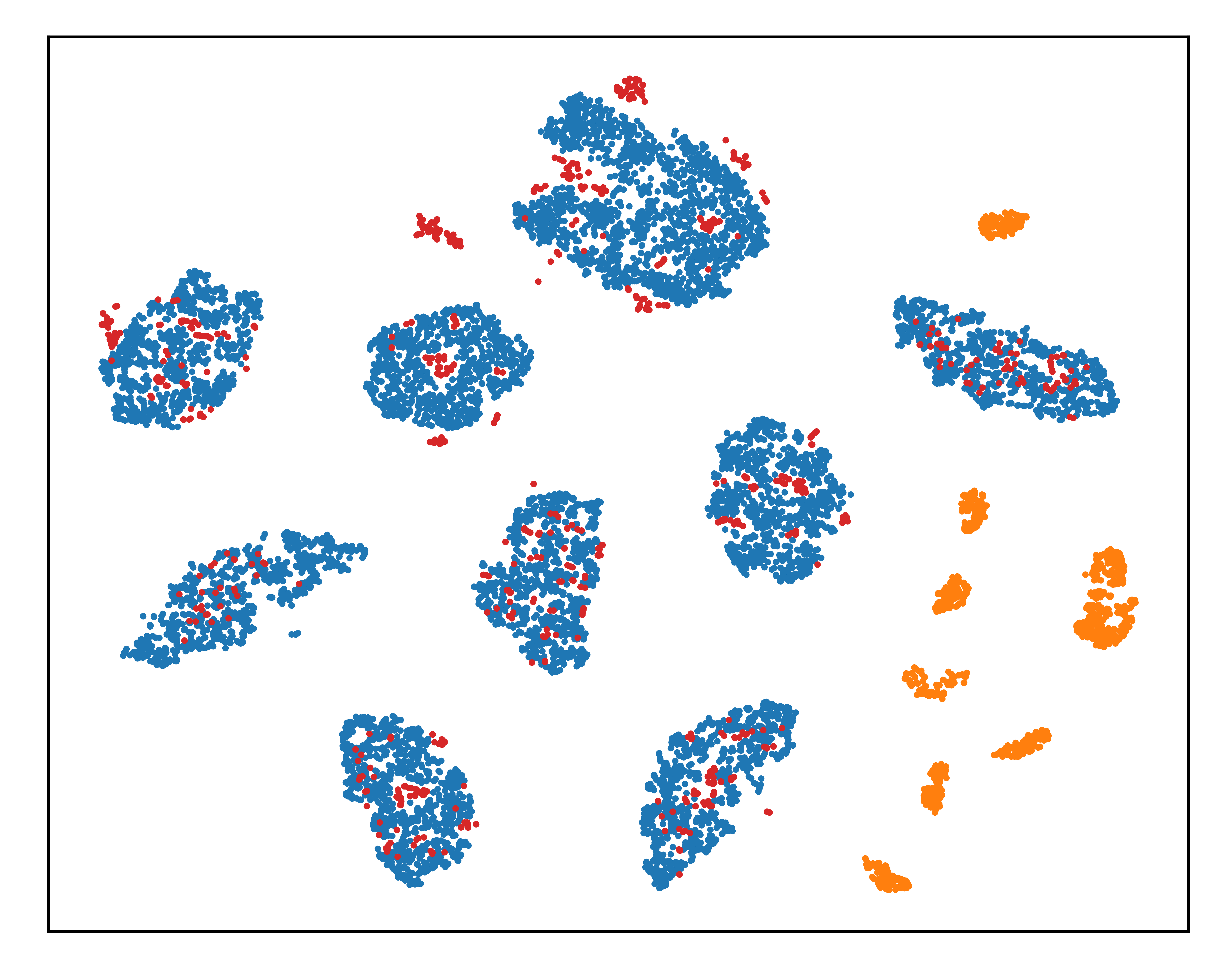}
    	\subcaption{EMD, PC.}
    	\label{image_app_emb_11}
    \end{minipage}
    \hfill
	\begin{minipage}[b]{\widthimages\linewidth}
        \centering
    	\includegraphics[trim=12 12 10 10, clip, width=1.0\linewidth]{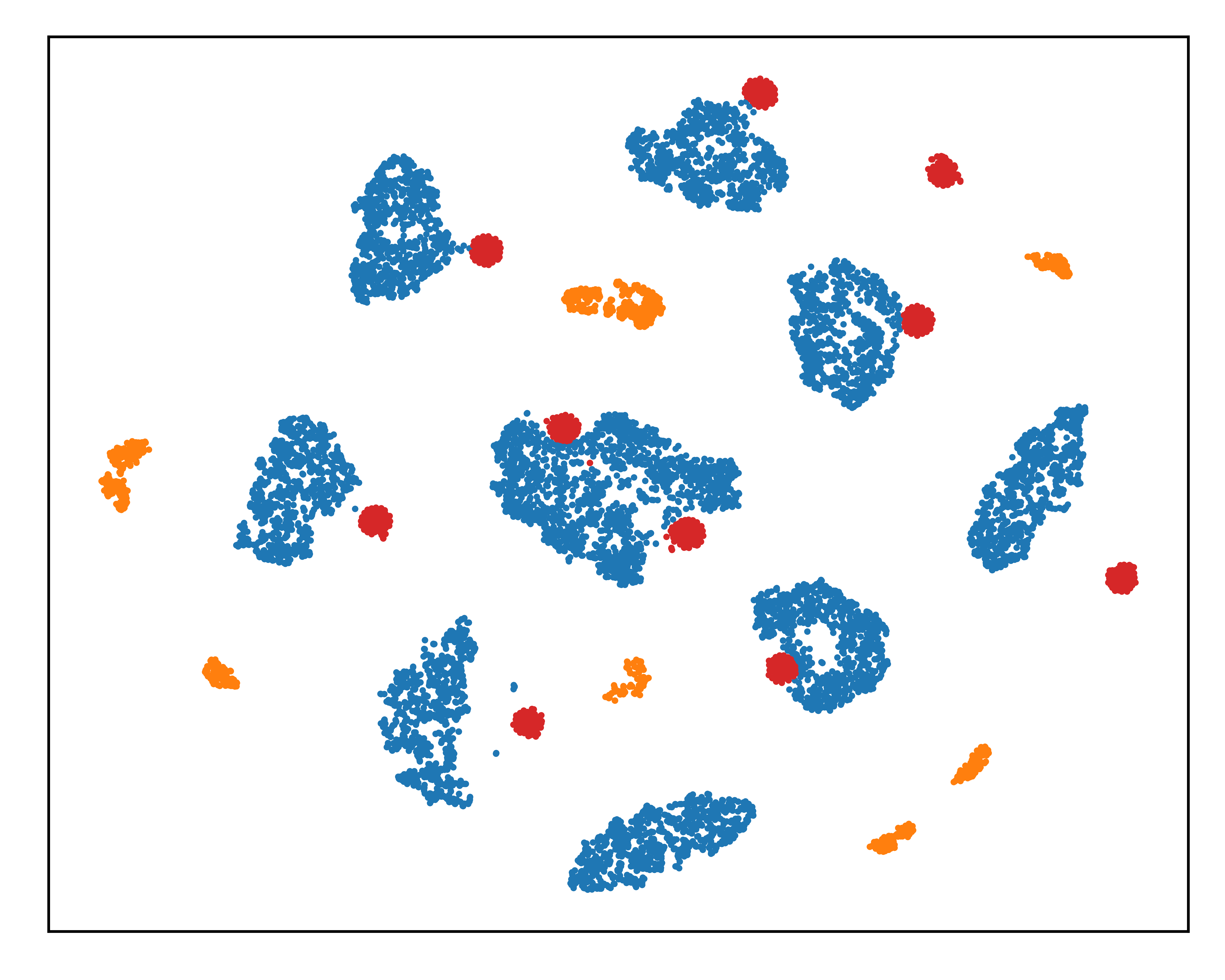}
    	\subcaption{Sinkhorn, PC.}
    	\label{image_app_emb_12}
    \end{minipage}
	\begin{minipage}[b]{\widthimages\linewidth}
        \centering
    	\includegraphics[trim=12 12 10 10, clip, width=1.0\linewidth]{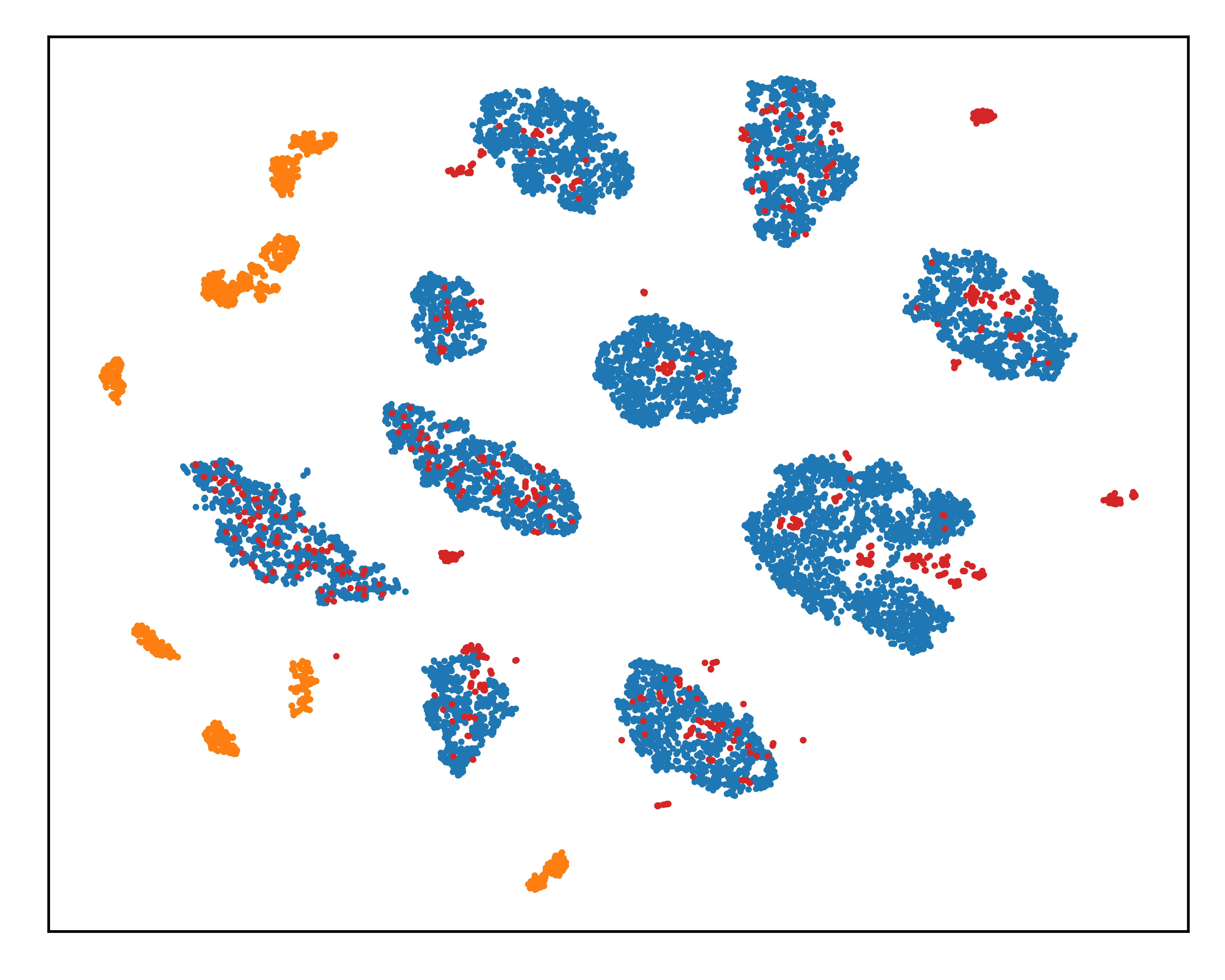}
    	\subcaption{EMD, Stein CORAL.}
    	\label{image_app_emb_13}
    \end{minipage}
    \hfill
	\begin{minipage}[b]{\widthimages\linewidth}
        \centering
    	\includegraphics[trim=12 12 10 10, clip, width=1.0\linewidth]{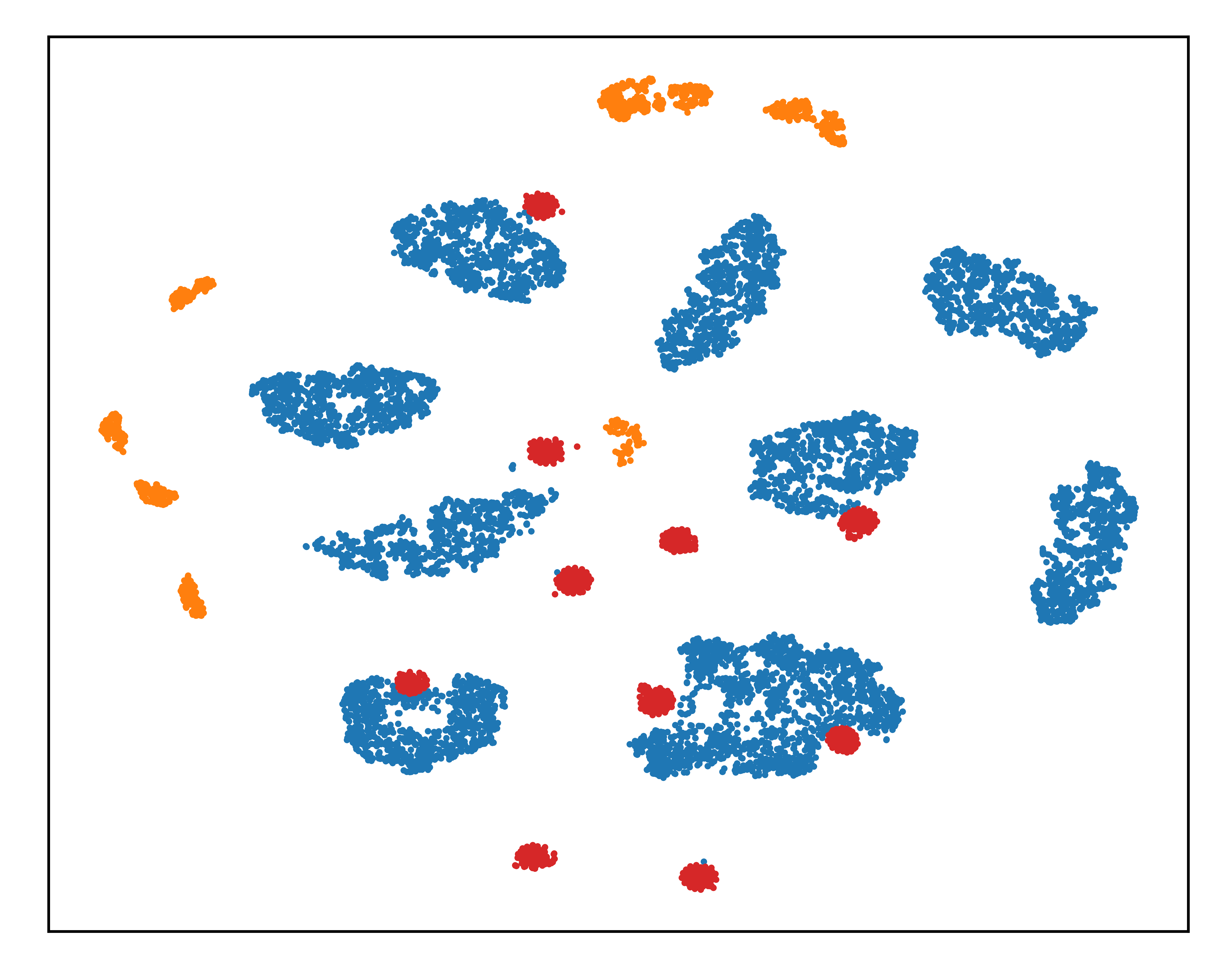}
    	\subcaption{Sinkhorn, Stein CORAL.}
    	\label{image_app_emb_14}
    \end{minipage}
    \hfill
	\begin{minipage}[b]{\widthimages\linewidth}
        \centering
    	\includegraphics[trim=12 12 10 10, clip, width=1.0\linewidth]{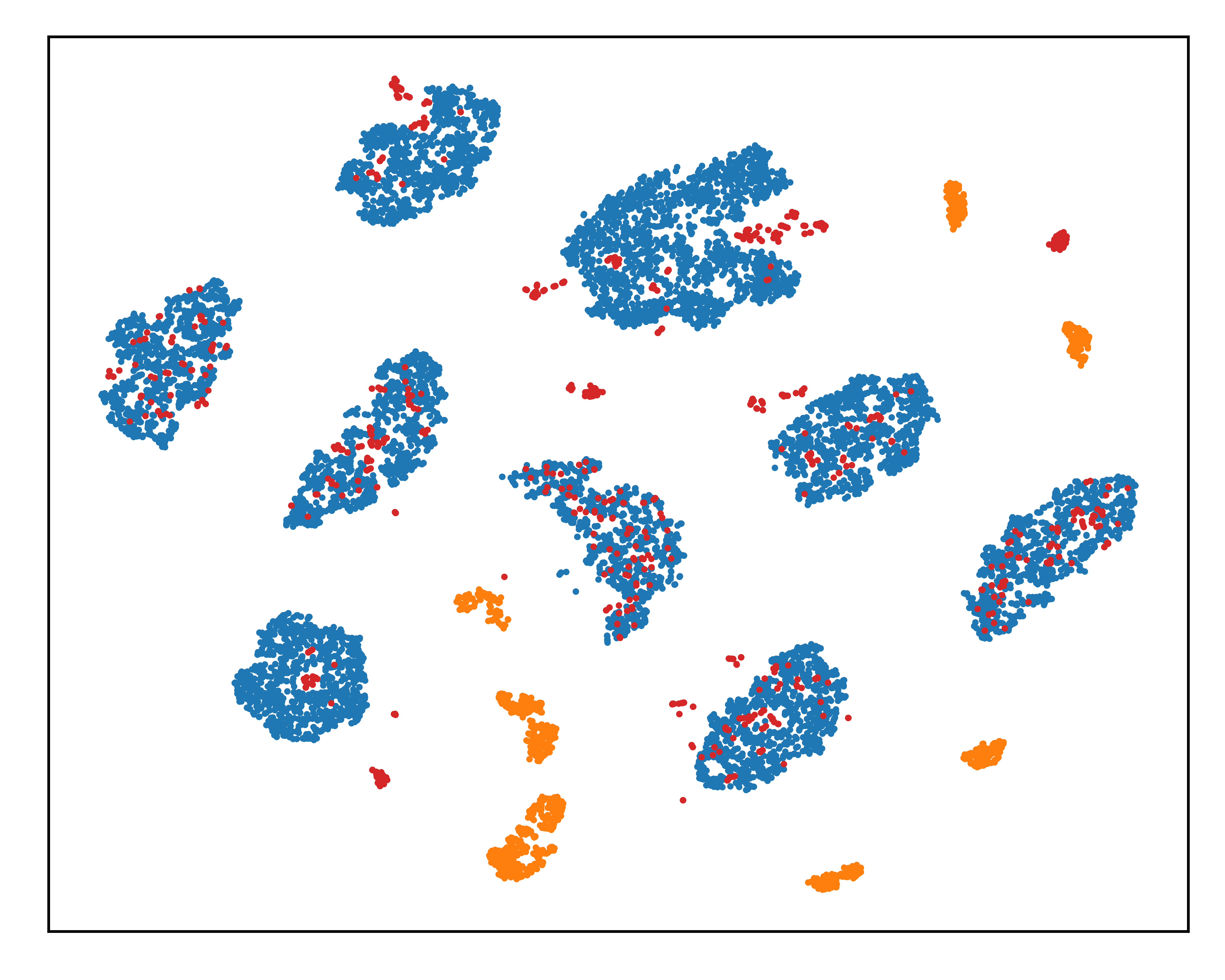}
    	\subcaption{EMD, Jeff CORAL.}
    	\label{image_app_emb_15}
    \end{minipage}
    \hfill
	\begin{minipage}[b]{\widthimages\linewidth}
        \centering
    	\includegraphics[trim=12 12 10 10, clip, width=1.0\linewidth]{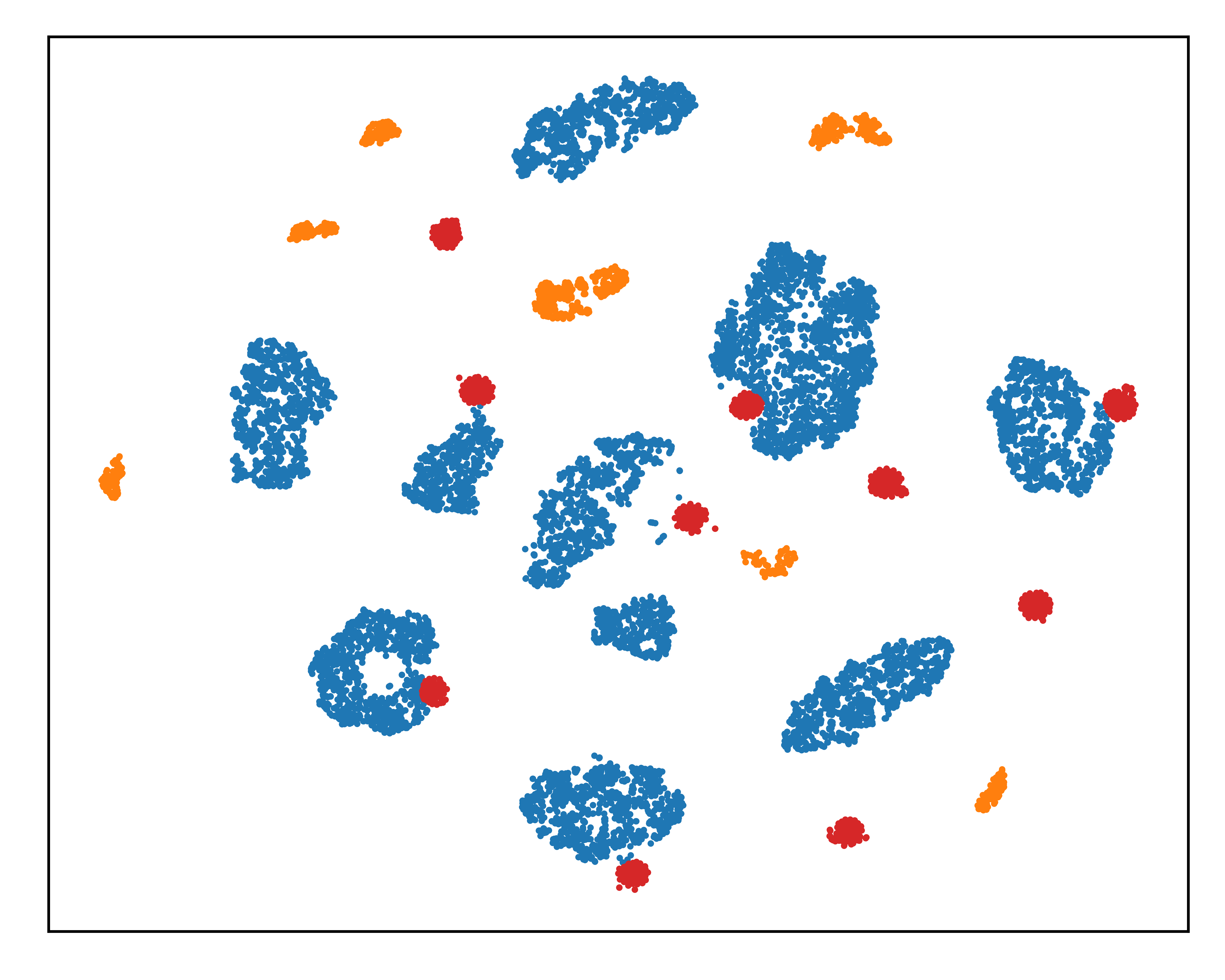}
    	\subcaption{Sinkhorn, Jeff CORAL.}
    	\label{image_app_emb_16}
    \end{minipage}
    \caption{For the caption, see Figure~\ref{image_app_emb}.}
    \label{image_app_emb_follow}
\end{figure*}

To better visualize the differences between domain-dependent embeddings and different methods, we plot two dimensional embeddings of the features $f(\mathbf{U})$. For the sinusoidal dataset, $f(\mathbf{U})$ is of size $50 \times 30$ (reshaped 1,500). We use t-SNE~\cite{maaten} with initial dimension 1,500, perplexity of 30, an initial momentum of 0.5, and a final momentum of 0.8. Figure~\ref{image_app_sin_cos_class} visualizes the feature embeddings for different noise parameters $b$ with class label dependent colors. Without noise ($b=0$), the clusters are clearly separable, while only label \texttt{'8'} and \texttt{'9'} overlap. As the noise increases $b=1.5$, also the cluster with label \texttt{'7'} overlaps, and finally label \texttt{'6'} and \texttt{'3'} for $b=1.9$. This is reflected by the results in Figure~\ref{image_eval_sinus}, where the accuracy drops for $b > 0.7$. The validation samples with domain shift are notably distant to the target samples.

Figure~\ref{image_app_emb} and \ref{image_app_emb_follow} visualize the feature embeddings $f(\mathbf{U})$ for $\mathcal{U}_T$ (blue), $\mathcal{U}_S$ (orange), and the transformed $\mathcal{U}_S^T$ (red). It is notable that EMD forms spread clusters of the specific sample embedding for the transformed source domain, while Sinkhorn shapes small clusters that are close to the target domain clusters, but are outlying. This distance increases for a higher noise. Choosing the best transformation (Figure~\ref{image_app_emb_1} and \ref{image_app_emb_2}), the transformed embeddings are mostly consistent, the kMMD (Figure~\ref{image_app_emb_5} and \ref{image_app_emb_6}) and cross correlation (Figure~\ref{image_app_emb_9} and \ref{image_app_emb_10}) distance metrics (that result in the highest classification accuracy), are similar to the best transformation. HoMM (Figure~\ref{image_app_emb_7} and \ref{image_app_emb_8}) with the lowest classification accuracy, also results in a larger spread of transformed clusters. Different are the variances Stein and Jeff CORAL trained with EMD where the transformed source domain samples are notably out-of the target domain samples (see Figure~\ref{image_app_emb_13} and \ref{image_app_emb_15}).

\begin{figure*}[t!]
	\centering
	\begin{minipage}[b]{0.33\linewidth}
        \centering
    	\includegraphics[trim=12 12 7 10, clip, width=1.0\linewidth]{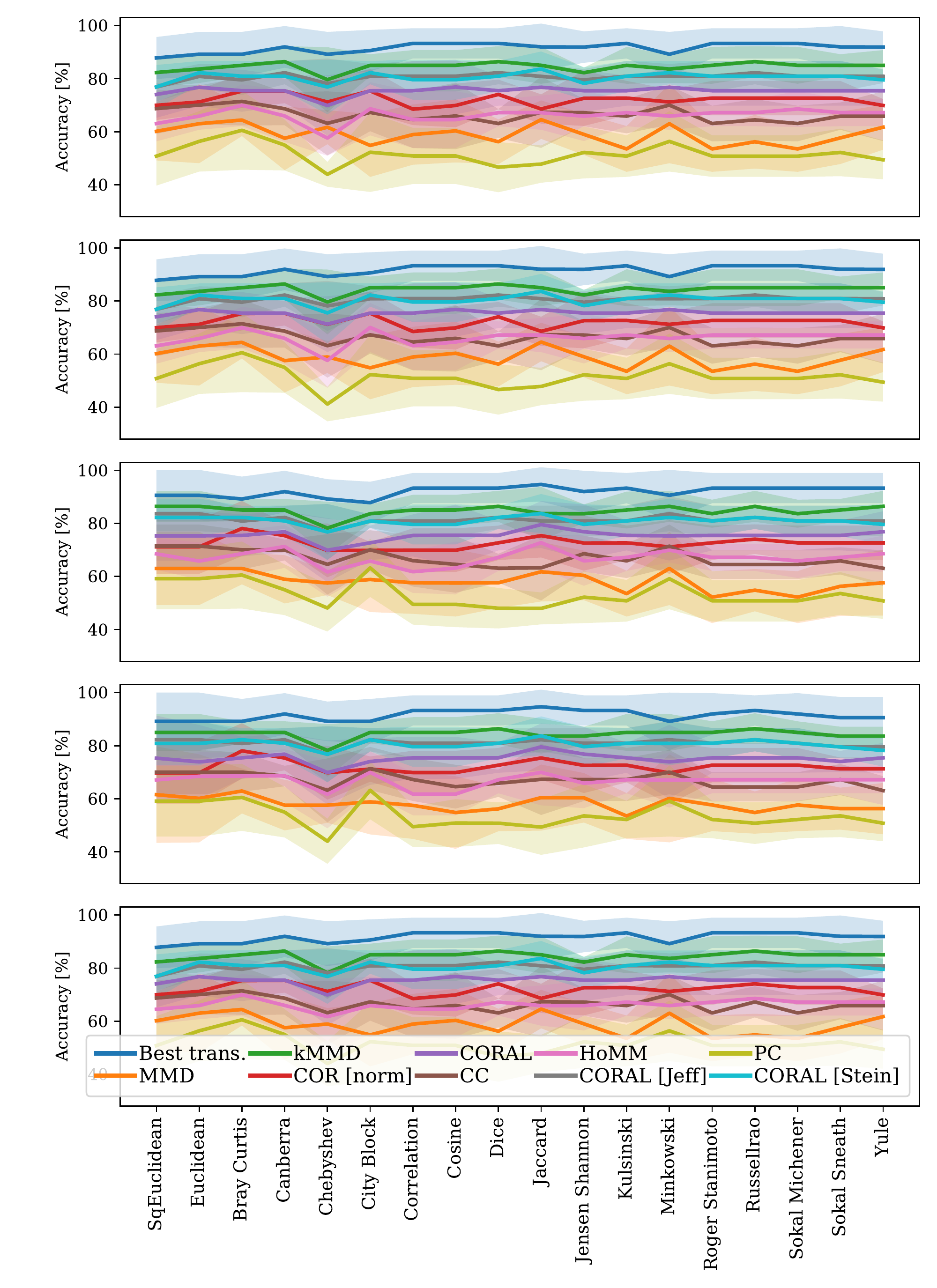}
    	\subcaption{EMD.}
    	\label{image_symbols_search_1}
    \end{minipage}
    \hfill
	\begin{minipage}[b]{0.33\linewidth}
        \centering
    	\includegraphics[trim=12 12 7 10, width=1.0\linewidth]{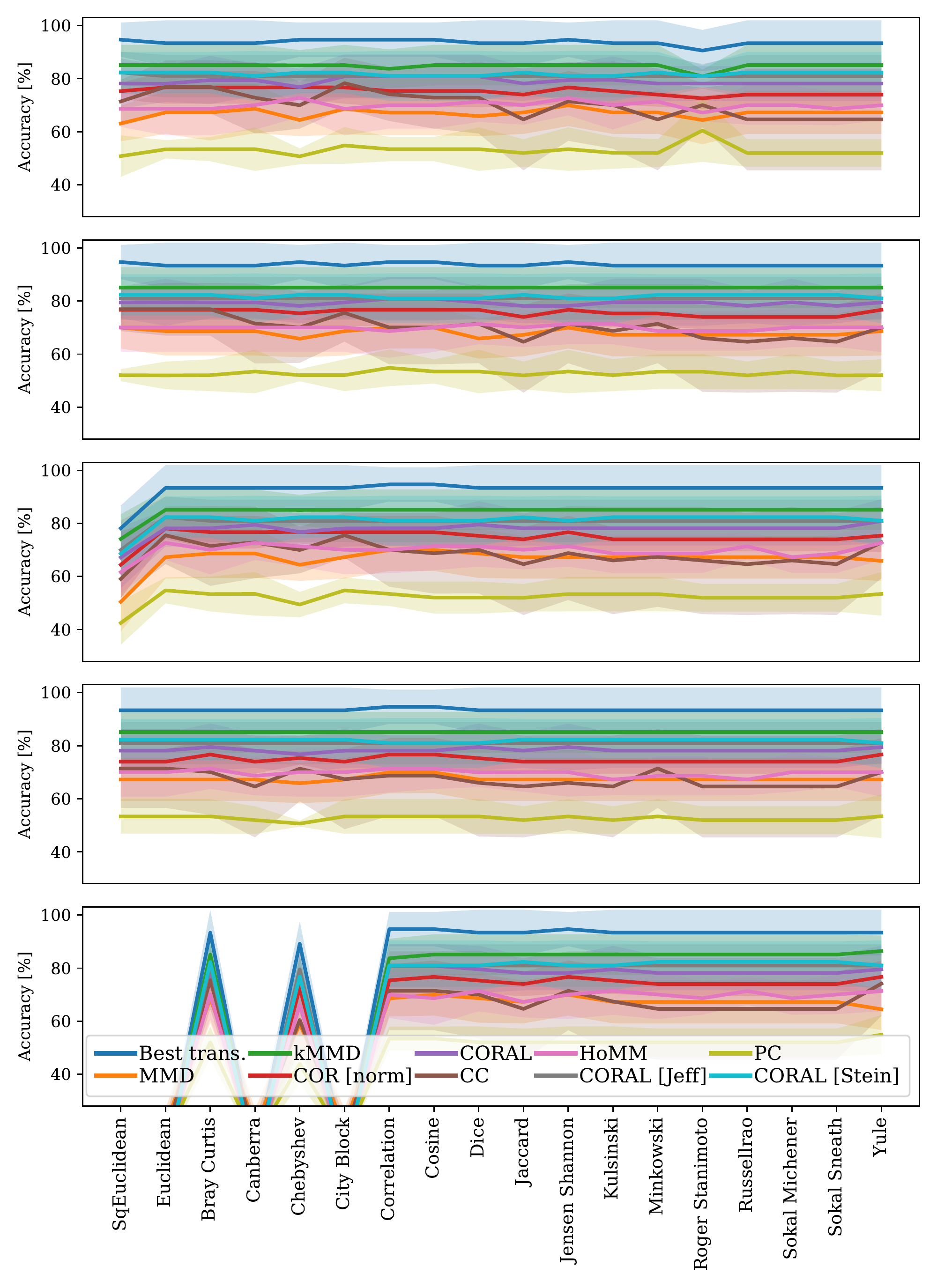}
    	\subcaption{Sinkhorn.}
    	\label{image_symbols_search_2}
    \end{minipage}
    \hfill
	\begin{minipage}[b]{0.33\linewidth}
        \centering
    	\includegraphics[trim=12 12 7 10, width=1.0\linewidth]{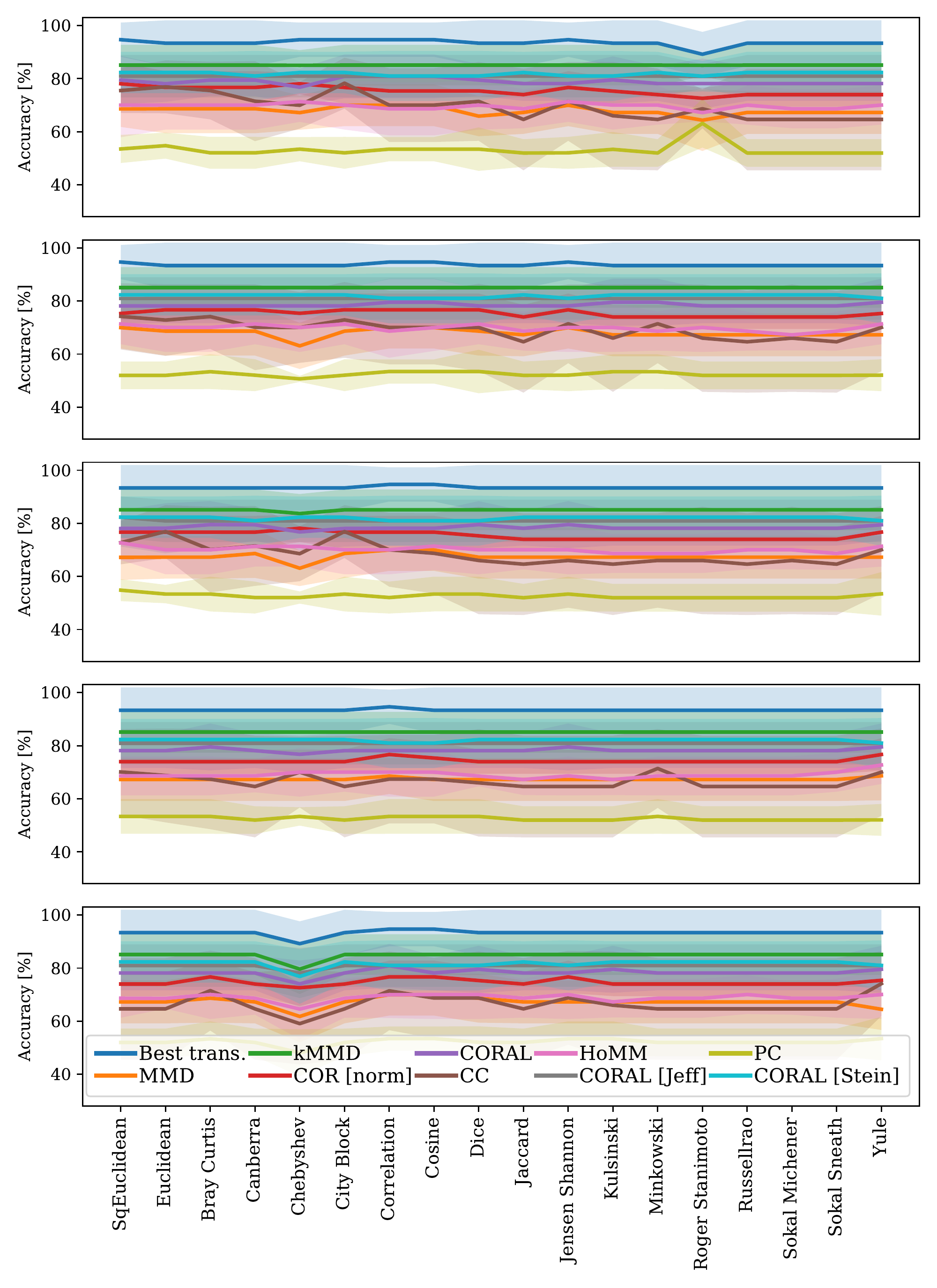}
    	\subcaption{Sinkhorn ($L_{1}L_{2}$).}
    	\label{image_symbols_search_3}
    \end{minipage}
    \caption{Hyperparameter search for distance metrics for optimal transport methods and transformation selection methods on the OnHW-symbols dataset. Results are averaged over four writers. 1: $median$. 2: $max$. 3: $log$. 4: $loglog$. 5: $None$.}
    \label{image_symbols_search}
\end{figure*}

\begin{figure*}[t!]
	\centering
	\begin{minipage}[b]{0.196\linewidth}
        \centering
    	\includegraphics[trim=12 14 10 11, clip, width=1.0\linewidth]{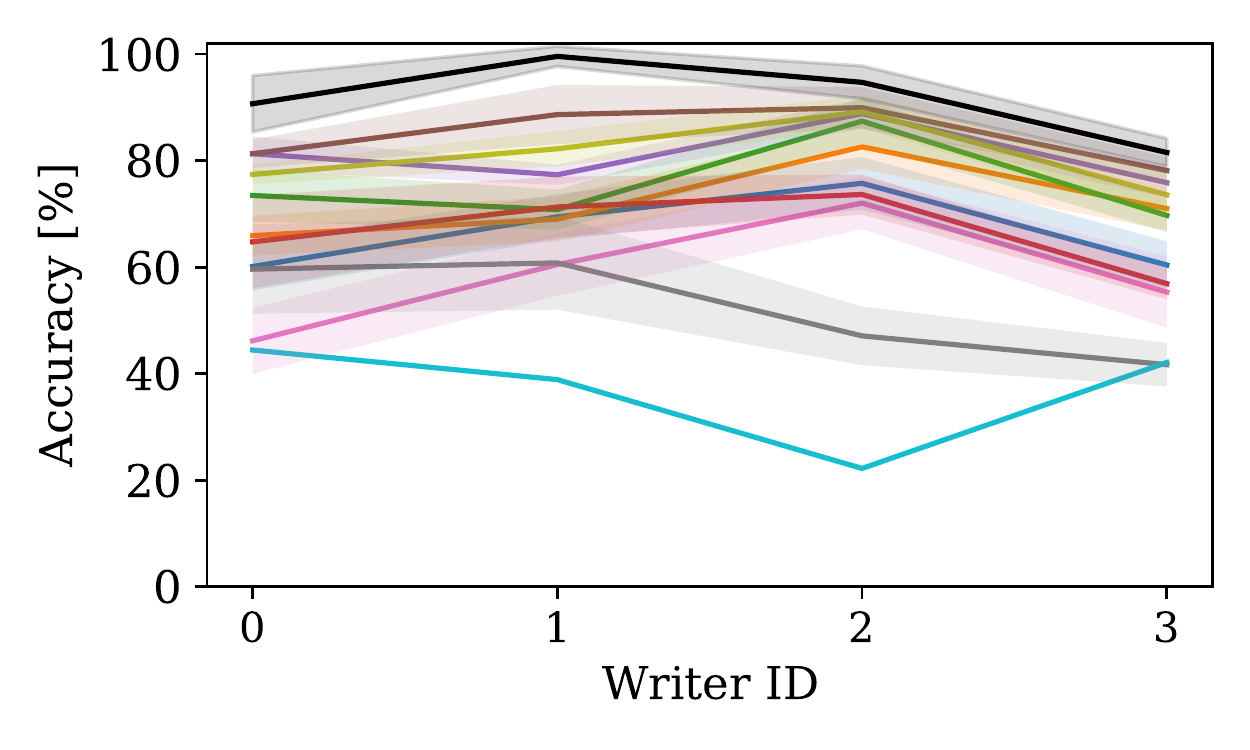}
    	\subcaption{EMD.}
    	\label{image_writer_symb_1}
    \end{minipage}
    \hfill
	\begin{minipage}[b]{0.196\linewidth}
        \centering
    	\includegraphics[trim=12 14 10 11, clip, width=1.0\linewidth]{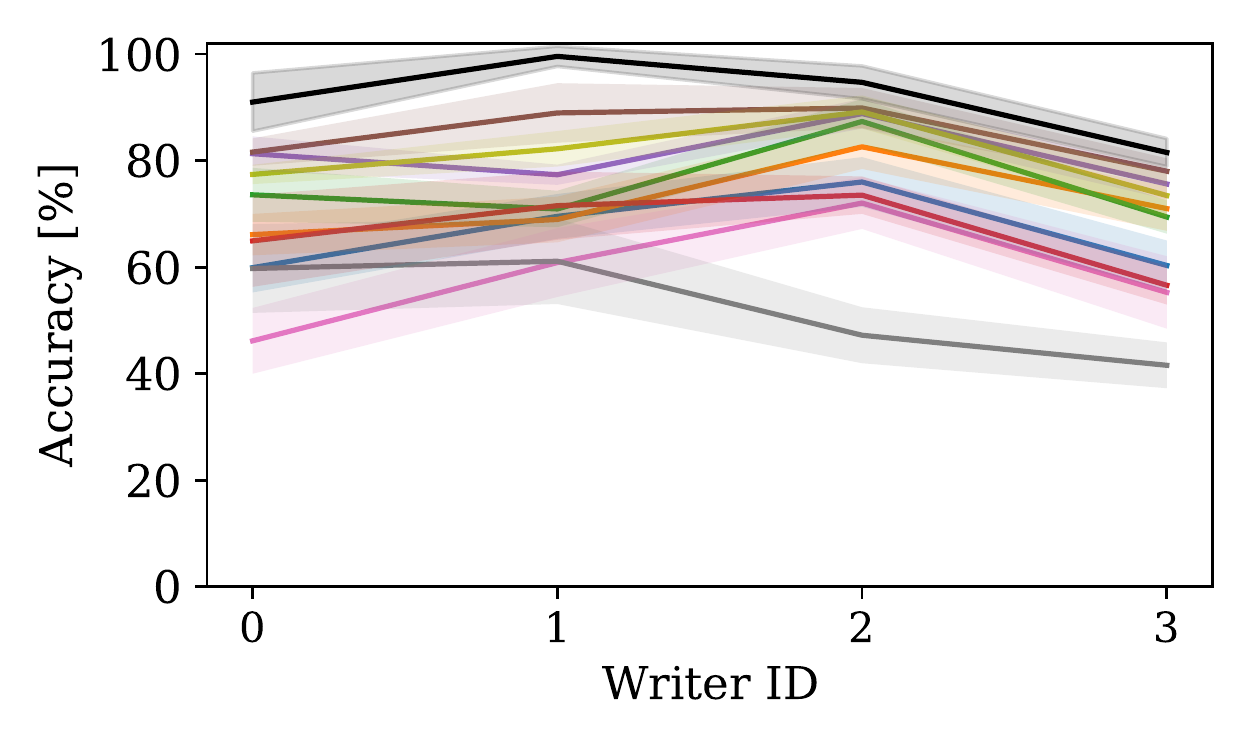}
    	\subcaption{SEMD.}
    	\label{image_writer_symb_2}
    \end{minipage}
    \hfill
	\begin{minipage}[b]{0.196\linewidth}
        \centering
    	\includegraphics[trim=12 14 10 11, clip, width=1.0\linewidth]{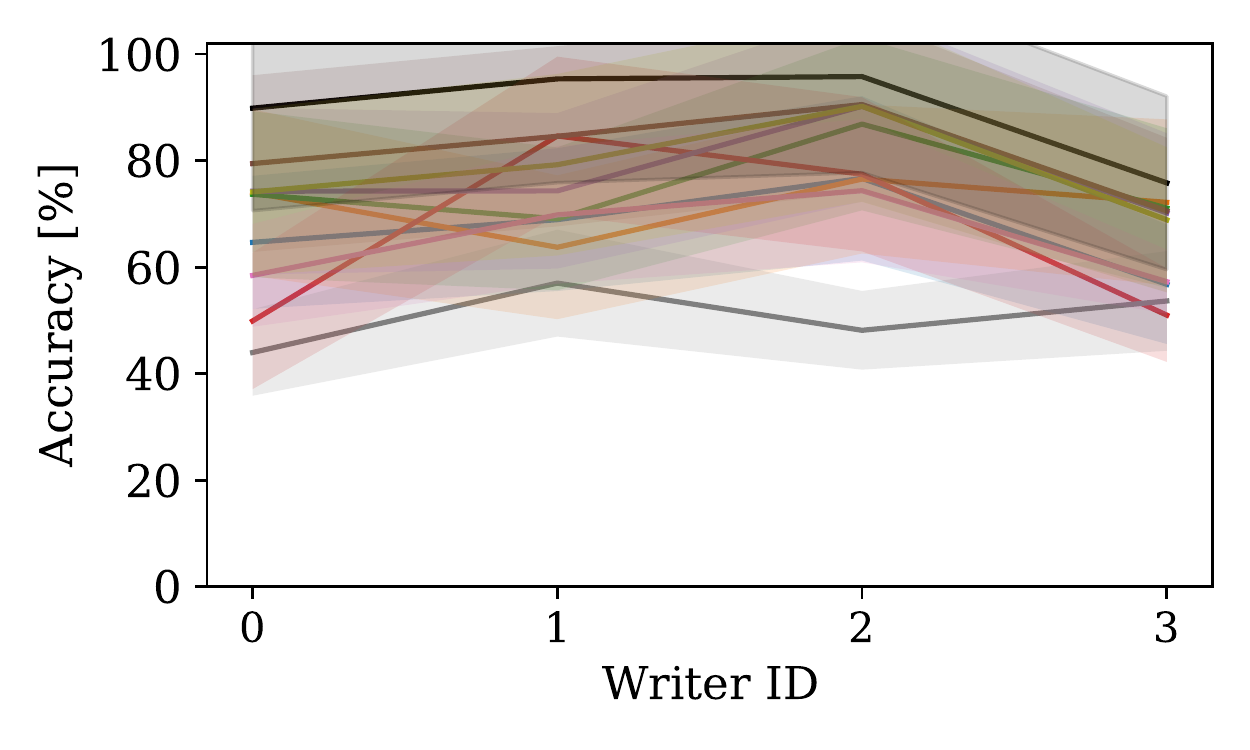}
    	\subcaption{Sinkhorn.}
    	\label{image_writer_symb_3}
    \end{minipage}
    \hfill
	\begin{minipage}[b]{0.196\linewidth}
        \centering
    	\includegraphics[trim=12 14 10 11, clip, width=1.0\linewidth]{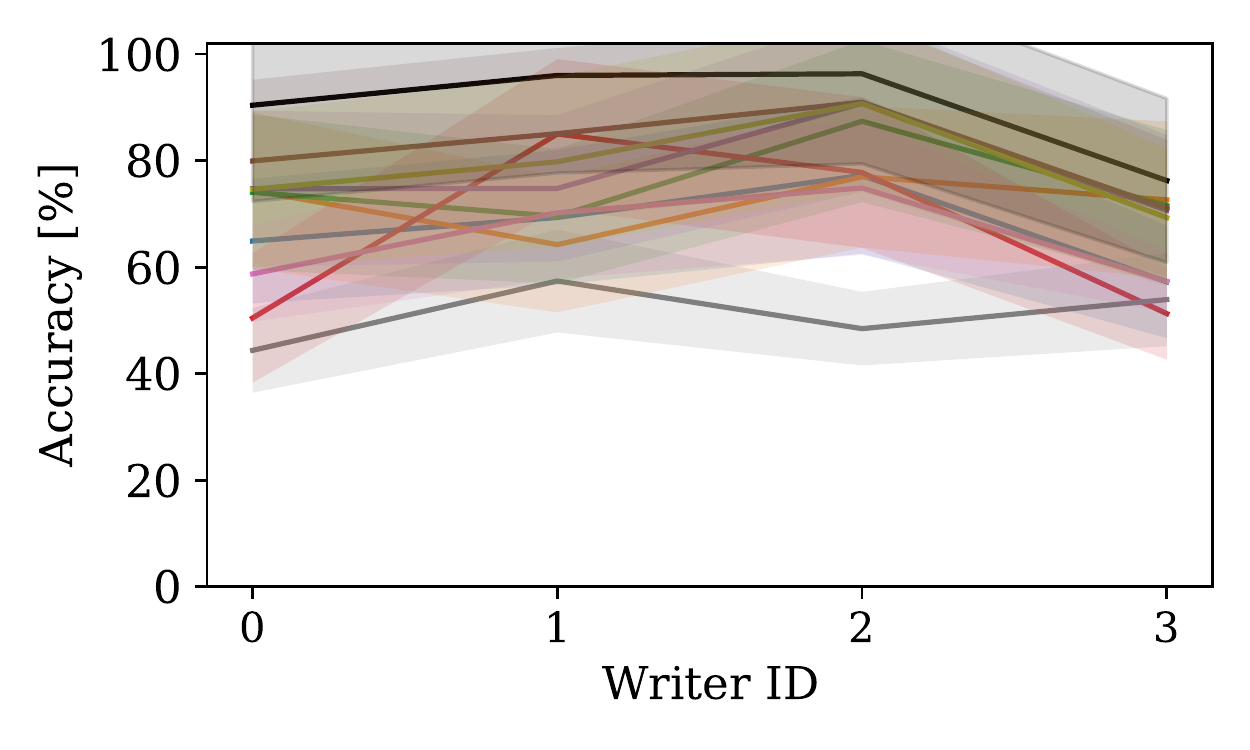}
    	\subcaption{Sinkhorn ($L_{p}L_{1}$).}
    	\label{image_writer_symb_4}
    \end{minipage}
    \hfill
	\begin{minipage}[b]{0.196\linewidth}
        \centering
    	\includegraphics[trim=12 14 10 11, clip, width=1.0\linewidth]{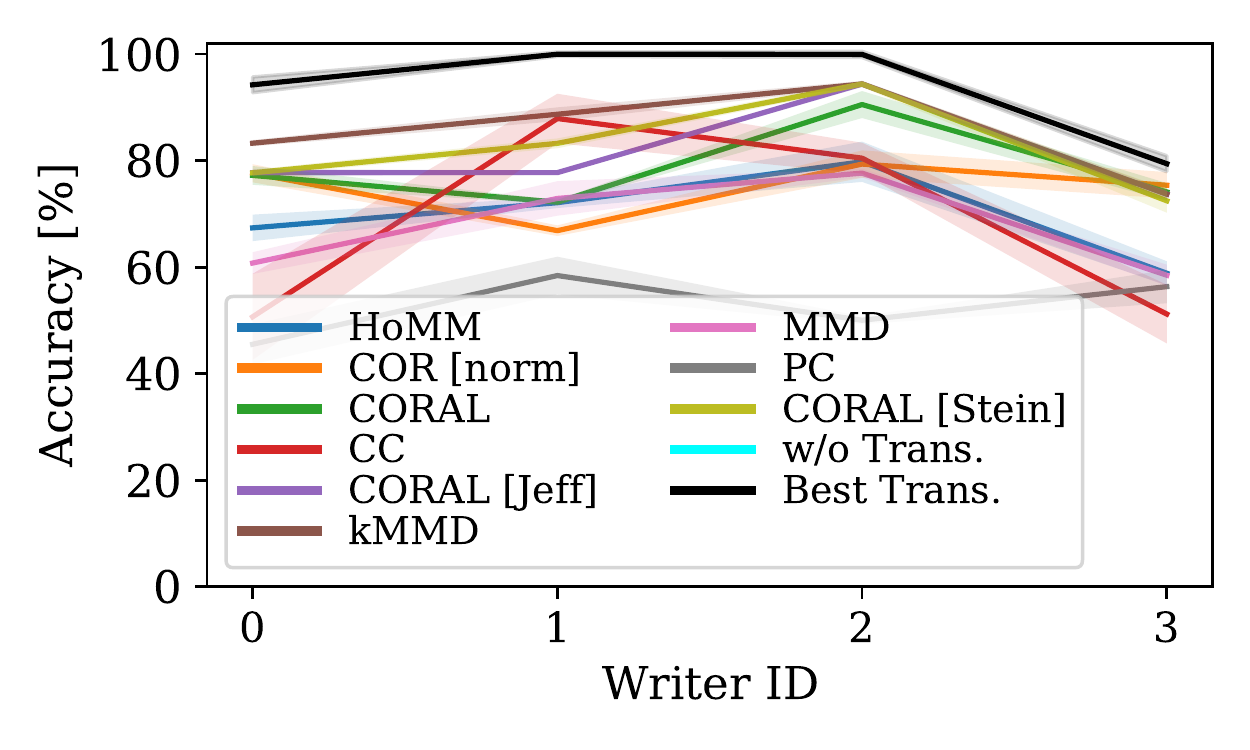}
    	\subcaption{Sinkhorn ($L_{1}L_{2}$).}
    	\label{image_writer_symb_5}
    \end{minipage}
    \caption{Evaluation of the transformed embeddings for the left-handed OnHW-symbols dataset for each of the four writers.}
    \label{image_writer_symb}
\end{figure*}

\begin{figure*}[t!]
	\centering
	\begin{minipage}[b]{0.196\linewidth}
        \centering
    	\includegraphics[trim=12 14 10 11, clip, width=1.0\linewidth]{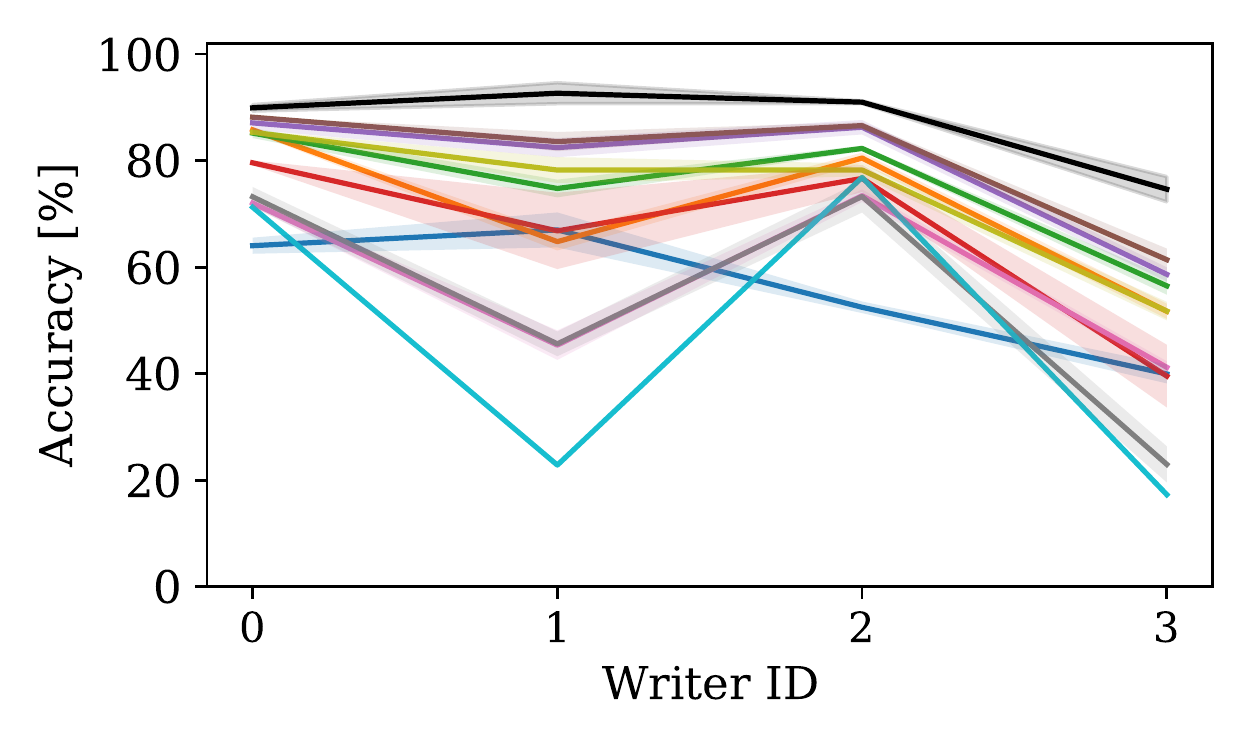}
    	\subcaption{EMD.}
    	\label{image_writer_equ_1}
    \end{minipage}
    \hfill
	\begin{minipage}[b]{0.196\linewidth}
        \centering
    	\includegraphics[trim=12 14 10 11, clip, width=1.0\linewidth]{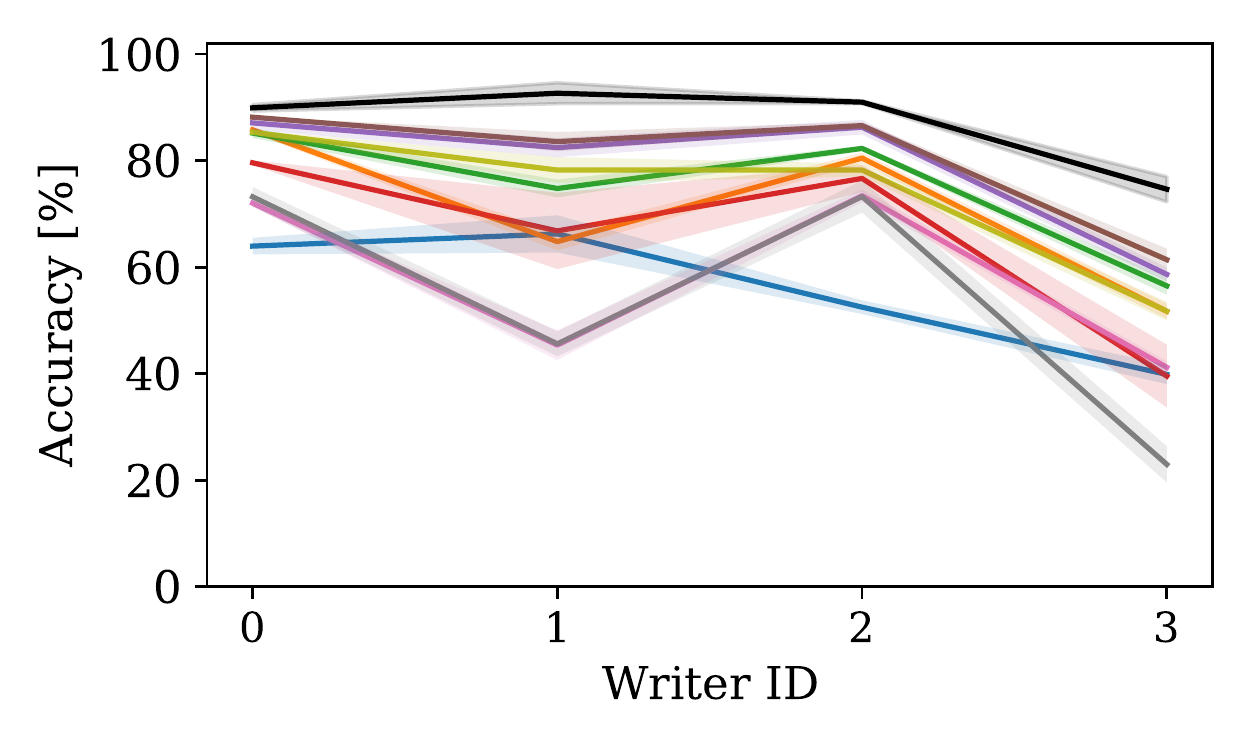}
    	\subcaption{SEMD.}
    	\label{image_writer_equ_2}
    \end{minipage}
    \hfill
	\begin{minipage}[b]{0.196\linewidth}
        \centering
    	\includegraphics[trim=12 14 10 11, clip, width=1.0\linewidth]{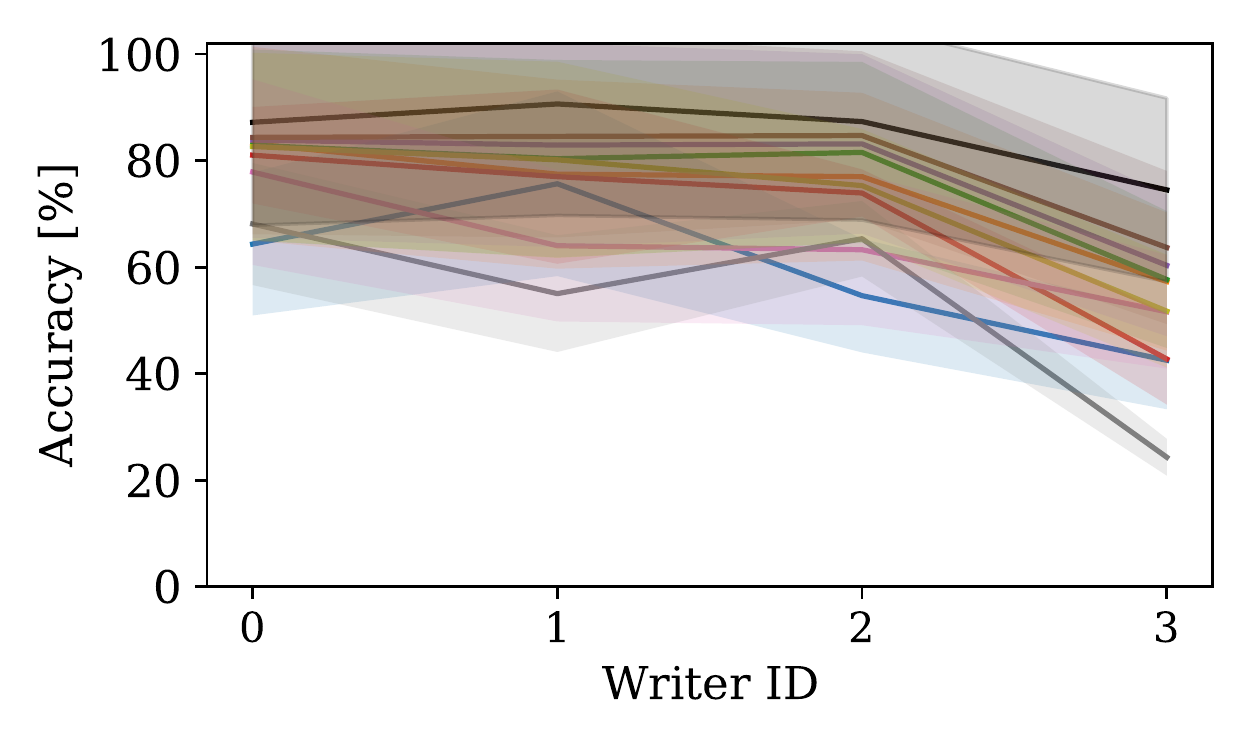}
    	\subcaption{Sinkhorn.}
    	\label{image_writer_equ_3}
    \end{minipage}
    \hfill
	\begin{minipage}[b]{0.196\linewidth}
        \centering
    	\includegraphics[trim=12 14 10 11, clip, width=1.0\linewidth]{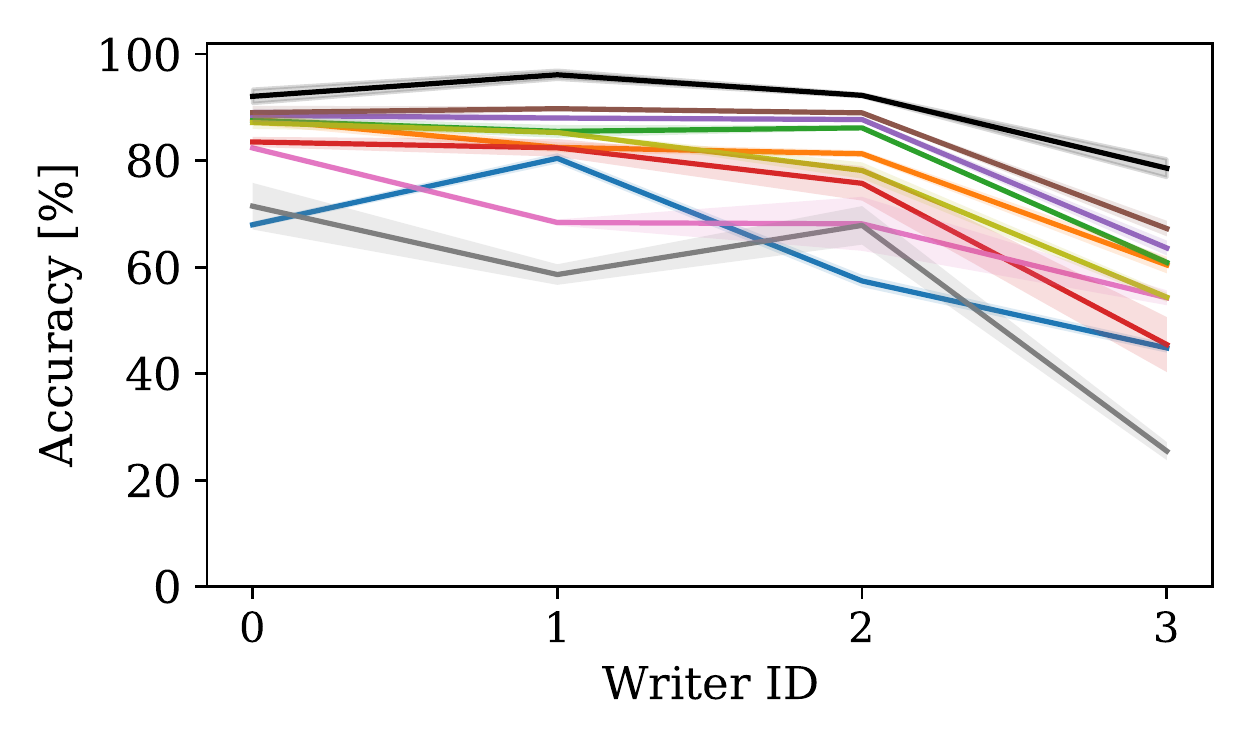}
    	\subcaption{Sinkhorn ($L_{p}L_{1}$).}
    	\label{image_writer_equ_4}
    \end{minipage}
    \hfill
	\begin{minipage}[b]{0.196\linewidth}
        \centering
    	\includegraphics[trim=12 14 10 11, clip, width=1.0\linewidth]{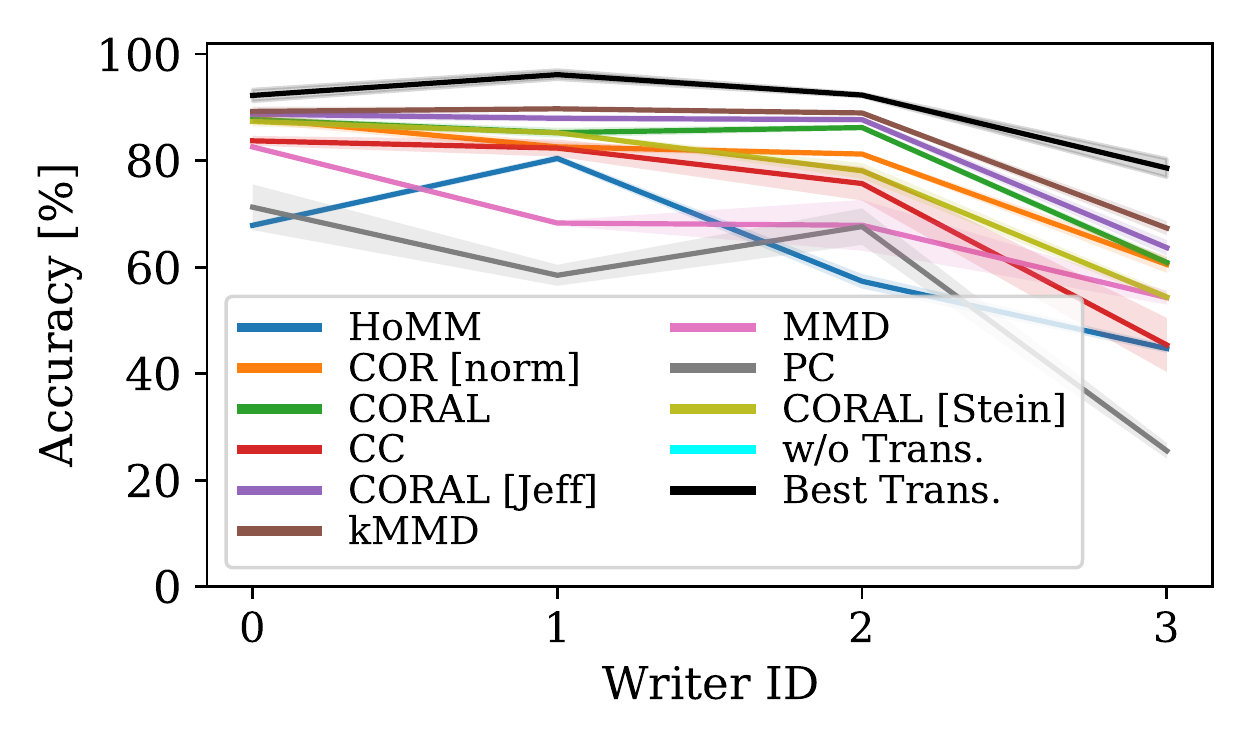}
    	\subcaption{Sinkhorn ($L_{1}L_{2}$).}
    	\label{image_writer_equ_5}
    \end{minipage}
    \caption{Evaluation of the transformed embeddings for the left-handed split OnHW-equations dataset for each of the four writers.}
    \label{image_writer_equ}
\end{figure*}

\begin{figure*}[t!]
	\centering
	\begin{minipage}[b]{0.196\linewidth}
        \centering
    	\includegraphics[trim=12 14 10 11, clip, width=1.0\linewidth]{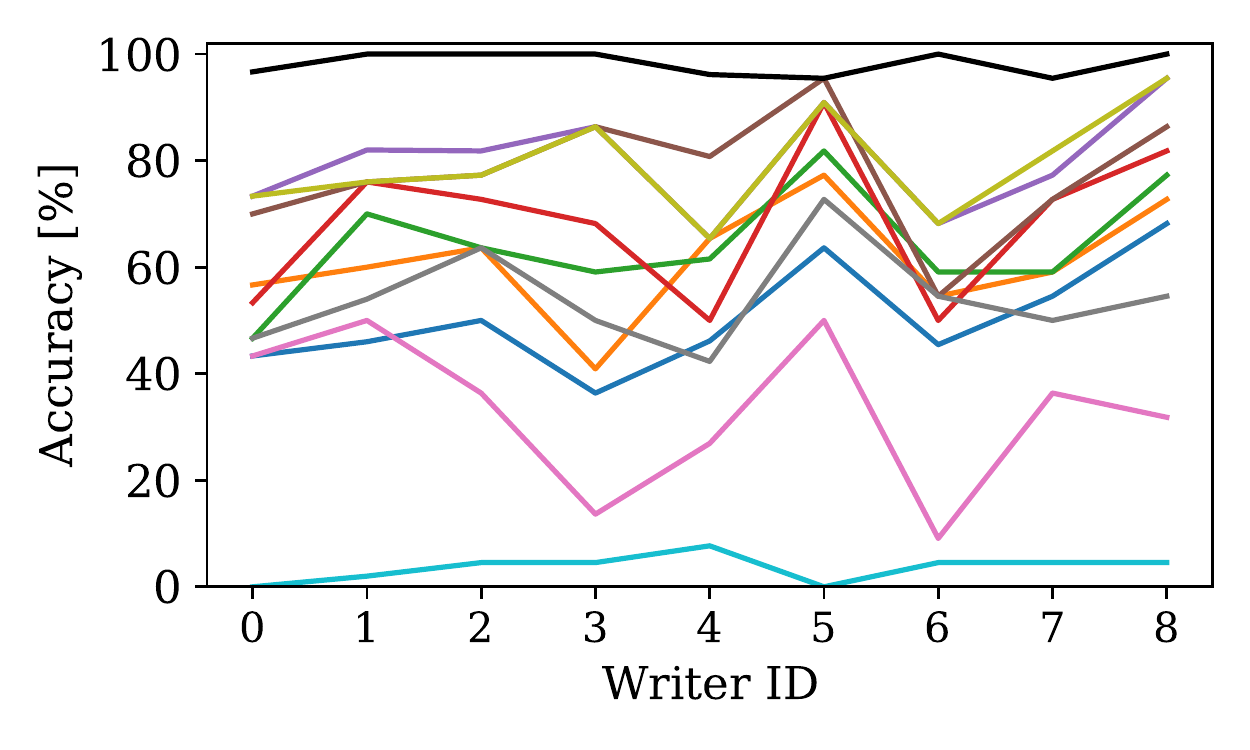}
    	\subcaption{EMD.}
    	\label{image_writer_lower_1}
    \end{minipage}
    \hfill
	\begin{minipage}[b]{0.196\linewidth}
        \centering
    	\includegraphics[trim=12 14 10 11, clip, width=1.0\linewidth]{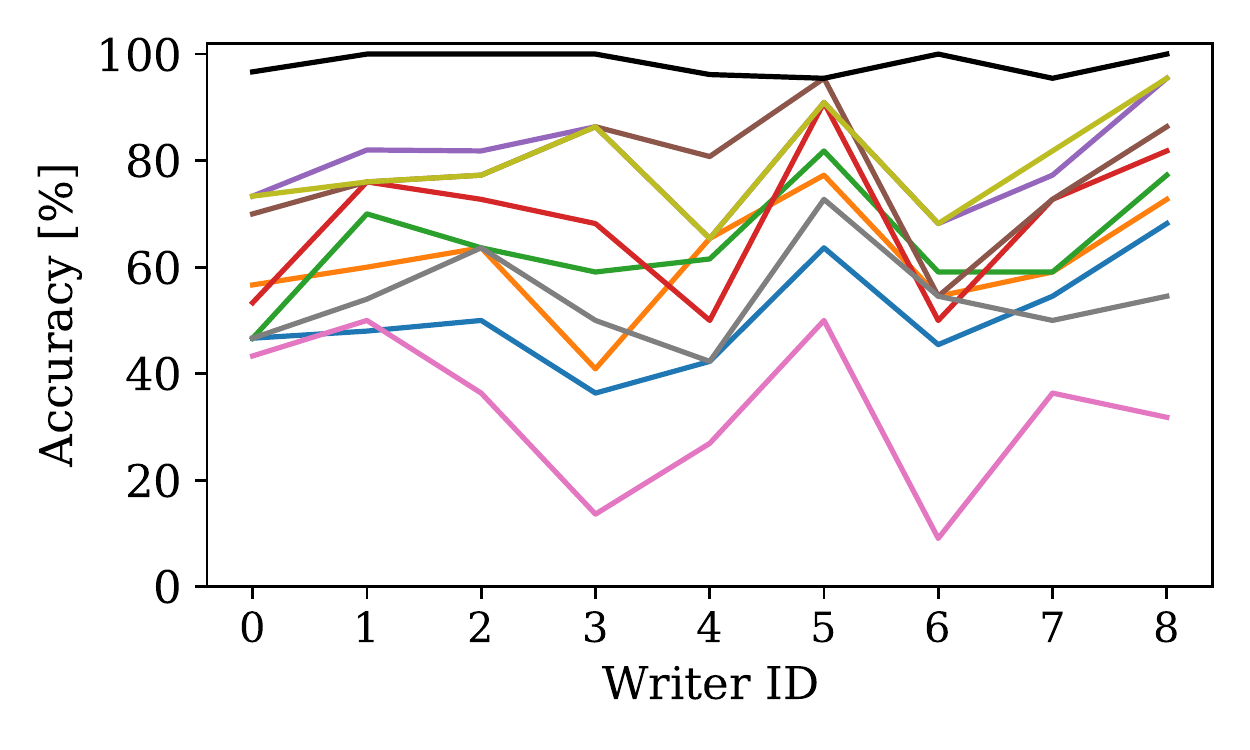}
    	\subcaption{SEMD.}
    	\label{image_writer_lower_2}
    \end{minipage}
    \hfill
	\begin{minipage}[b]{0.196\linewidth}
        \centering
    	\includegraphics[trim=12 14 10 11, clip, width=1.0\linewidth]{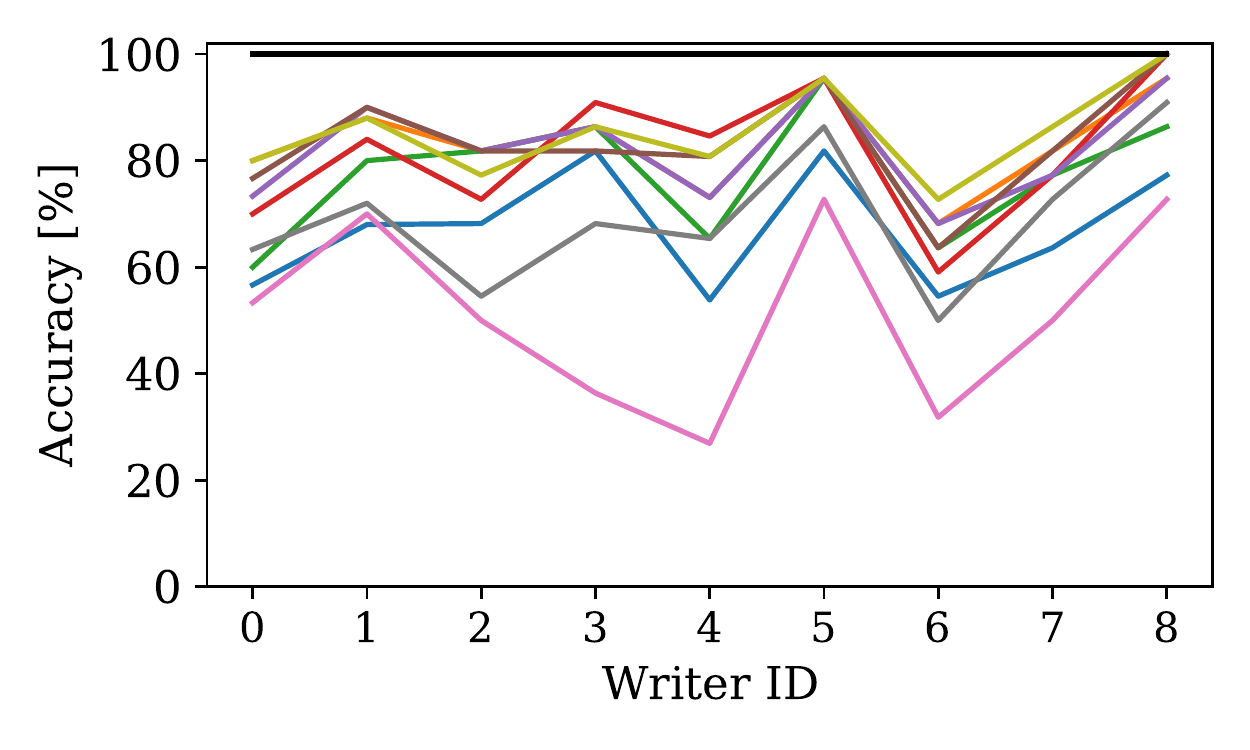}
    	\subcaption{Sinkhorn.}
    	\label{image_writer_lower_3}
    \end{minipage}
    \hfill
	\begin{minipage}[b]{0.196\linewidth}
        \centering
    	\includegraphics[trim=12 14 10 11, clip, width=1.0\linewidth]{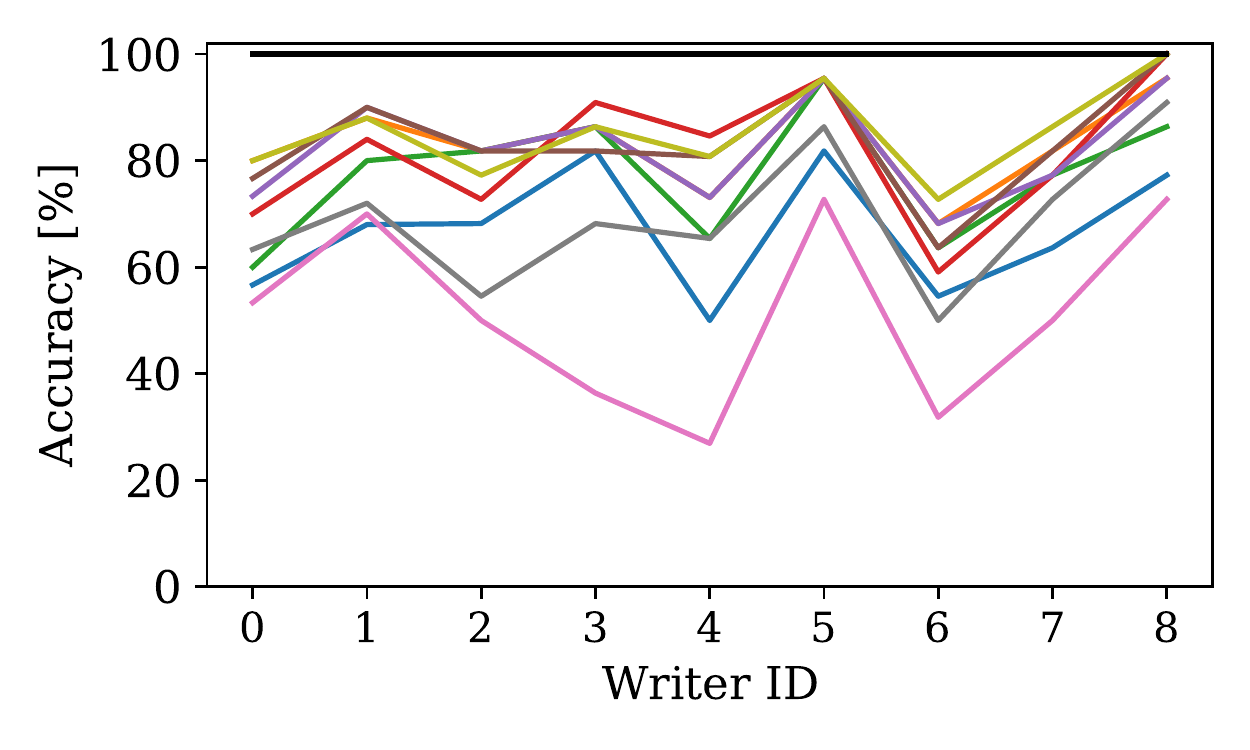}
    	\subcaption{Sinkhorn ($L_{p}L_{1}$).}
    	\label{image_writer_lower_4}
    \end{minipage}
    \hfill
	\begin{minipage}[b]{0.196\linewidth}
        \centering
    	\includegraphics[trim=12 14 10 11, clip, width=1.0\linewidth]{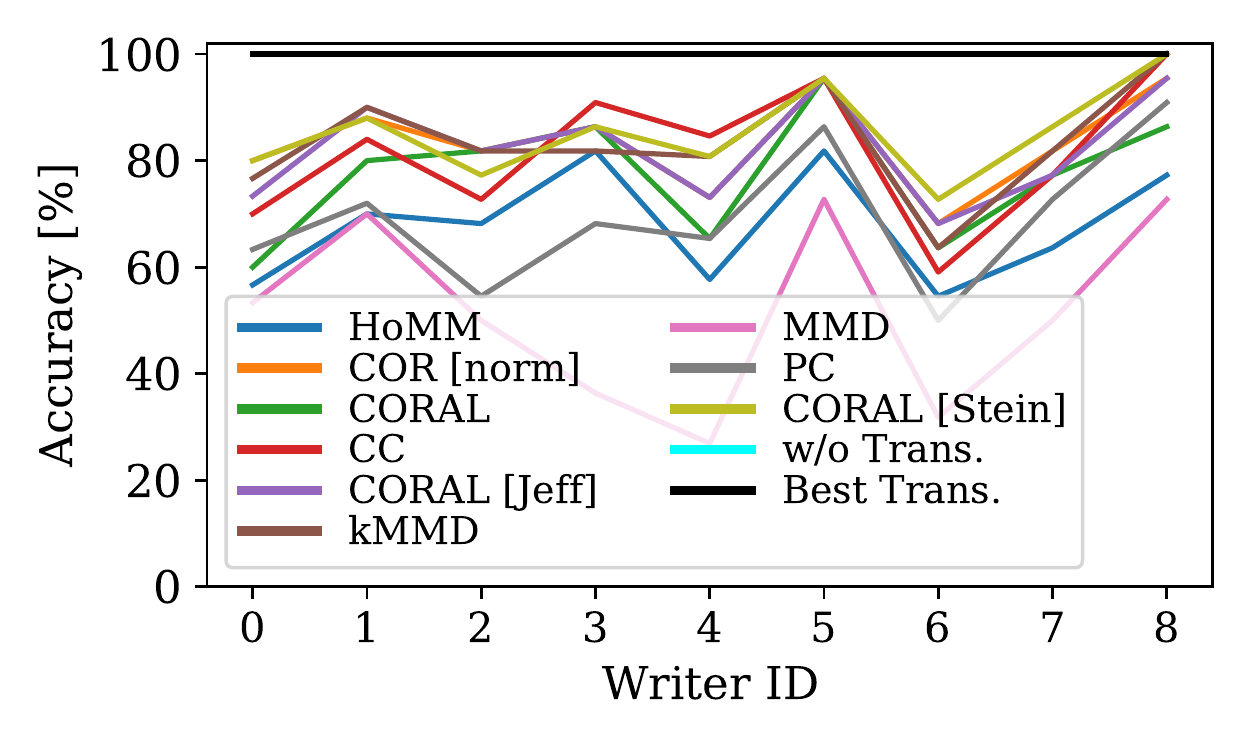}
    	\subcaption{Sinkhorn ($L_{1}L_{2}$).}
    	\label{image_writer_lower_5}
    \end{minipage}
    \caption{Evaluation of the transformed embeddings for the left-handed OnHW-chars (lower) dataset for each writer.}
    \label{image_writer_lower}
\end{figure*}

\begin{figure*}[t!]
	\centering
	\begin{minipage}[b]{0.196\linewidth}
        \centering
    	\includegraphics[trim=12 14 10 11, clip, width=1.0\linewidth]{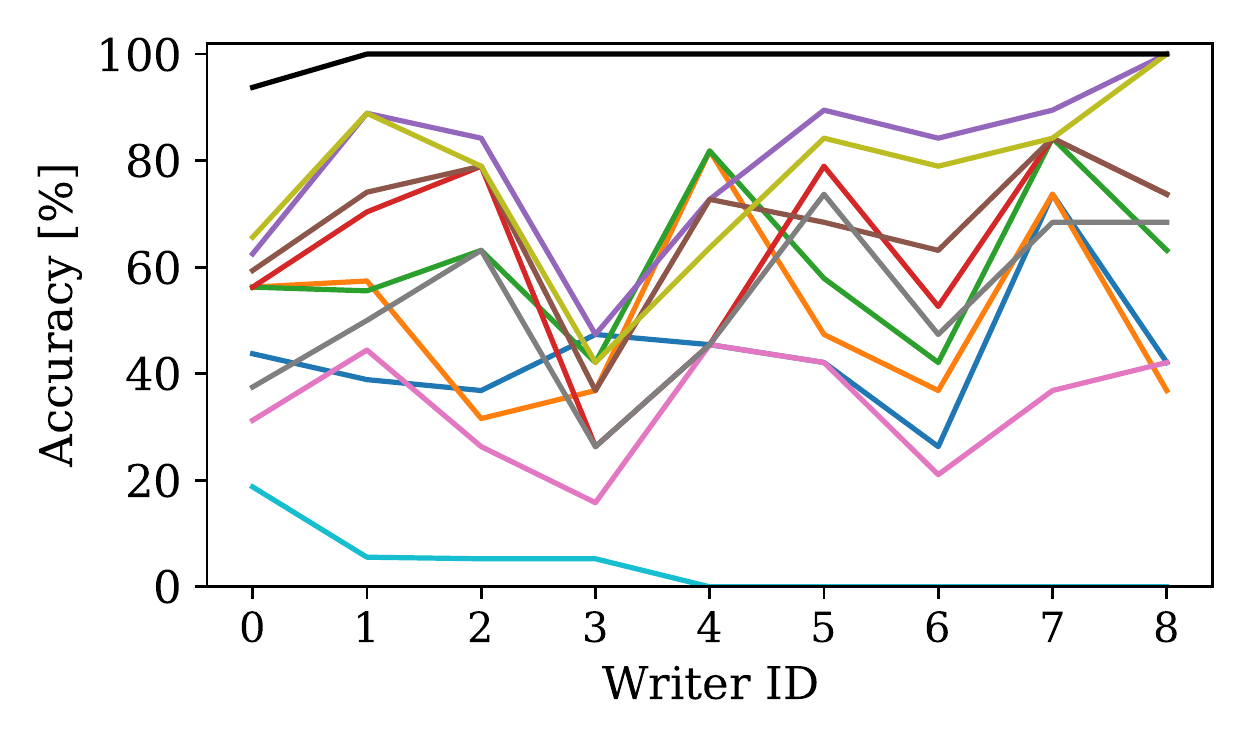}
    	\subcaption{EMD.}
    	\label{image_writer_upper_1}
    \end{minipage}
    \hfill
	\begin{minipage}[b]{0.196\linewidth}
        \centering
    	\includegraphics[trim=12 14 10 11, clip, width=1.0\linewidth]{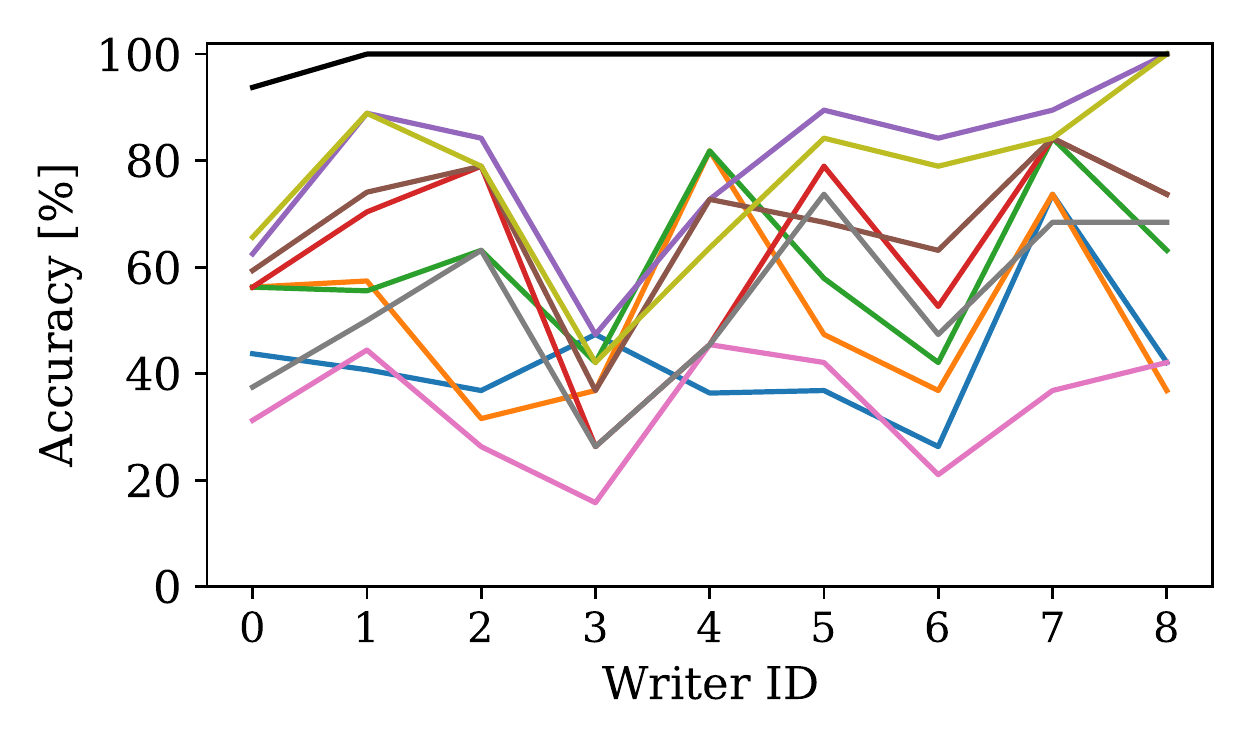}
    	\subcaption{SEMD.}
    	\label{image_writer_upper_2}
    \end{minipage}
    \hfill
	\begin{minipage}[b]{0.196\linewidth}
        \centering
    	\includegraphics[trim=12 14 10 11, clip, width=1.0\linewidth]{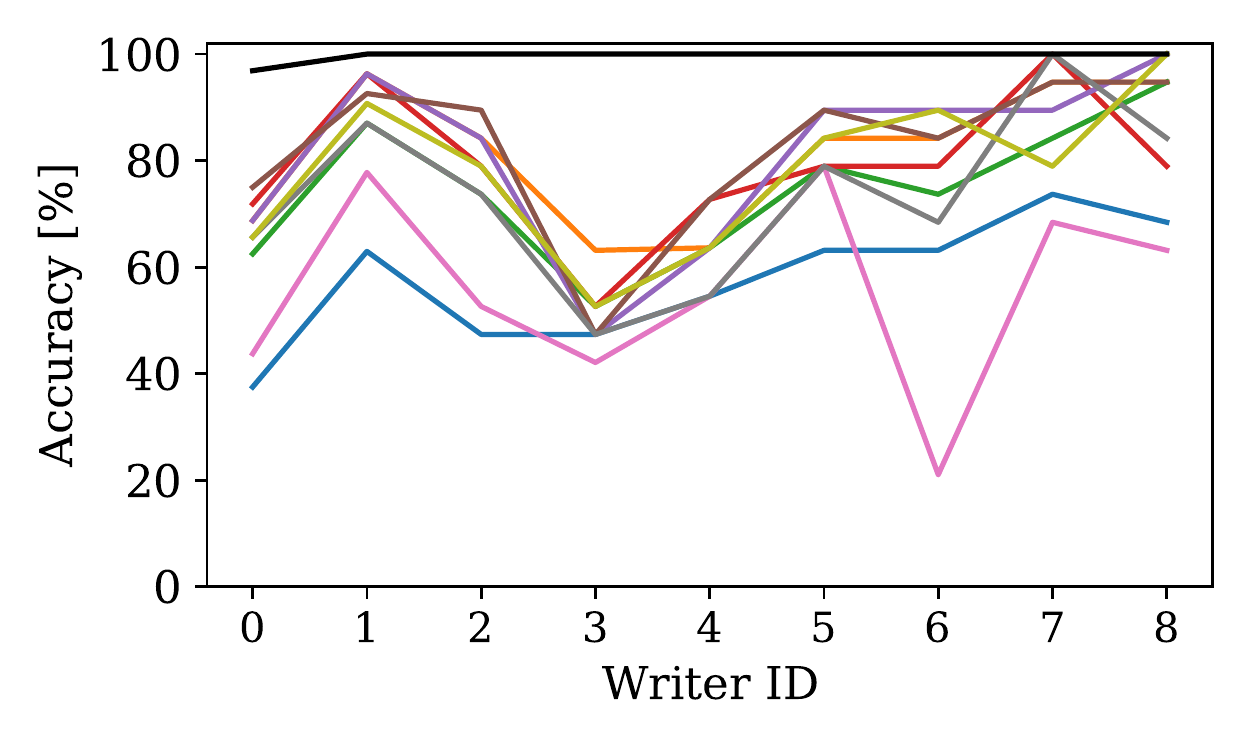}
    	\subcaption{Sinkhorn.}
    	\label{image_writer_upper_3}
    \end{minipage}
    \hfill
	\begin{minipage}[b]{0.196\linewidth}
        \centering
    	\includegraphics[trim=12 14 10 11, clip, width=1.0\linewidth]{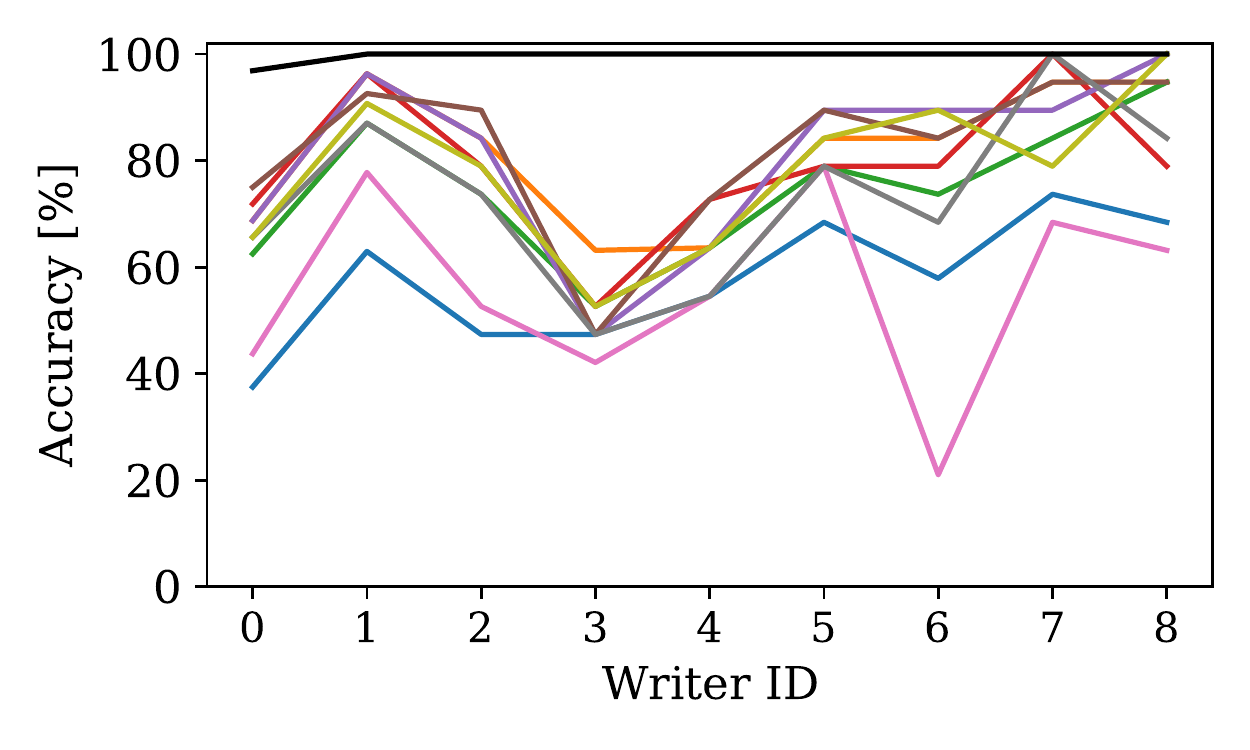}
    	\subcaption{Sinkhorn ($L_{p}L_{1}$).}
    	\label{image_writer_upper_4}
    \end{minipage}
    \hfill
	\begin{minipage}[b]{0.196\linewidth}
        \centering
    	\includegraphics[trim=12 14 10 11, clip, width=1.0\linewidth]{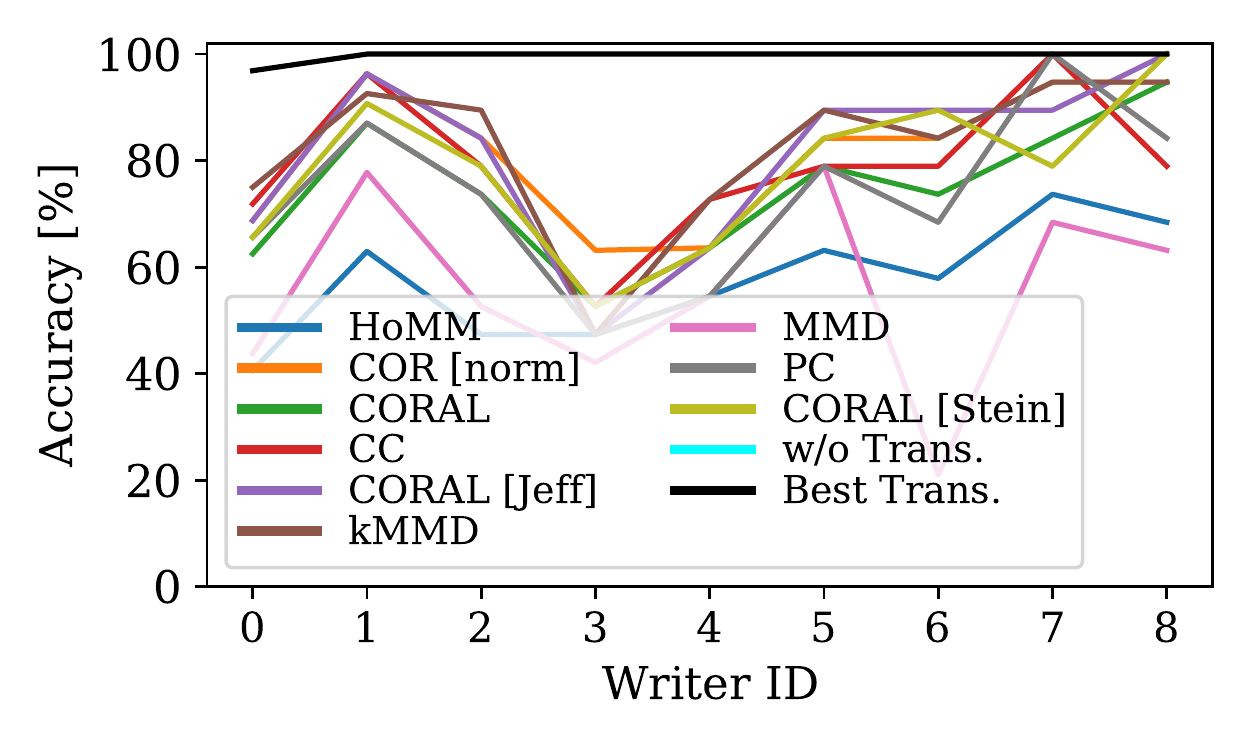}
    	\subcaption{Sinkhorn ($L_{1}L_{2}$).}
    	\label{image_writer_upper_5}
    \end{minipage}
    \vspace{-0.2cm}
    \caption{Evaluation of the transformed embeddings for the left-handed OnHW-chars (upper) dataset for each writers.}
    \label{image_writer_upper}
\end{figure*}

\begin{figure*}[t!]
	\centering
	\begin{minipage}[b]{0.196\linewidth}
        \centering
    	\includegraphics[trim=12 14 10 11, clip, width=1.0\linewidth]{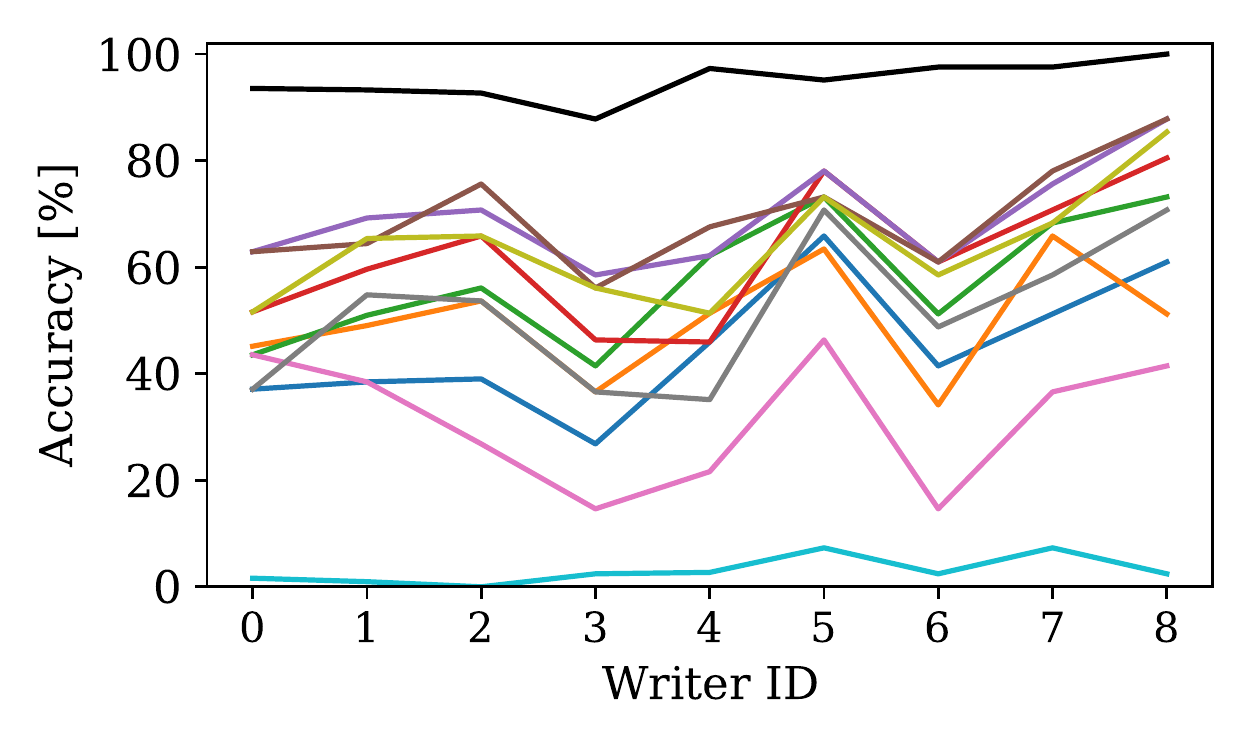}
    	\subcaption{EMD.}
    	\label{image_writer_comb_1}
    \end{minipage}
    \hfill
	\begin{minipage}[b]{0.196\linewidth}
        \centering
    	\includegraphics[trim=12 14 10 11, clip, width=1.0\linewidth]{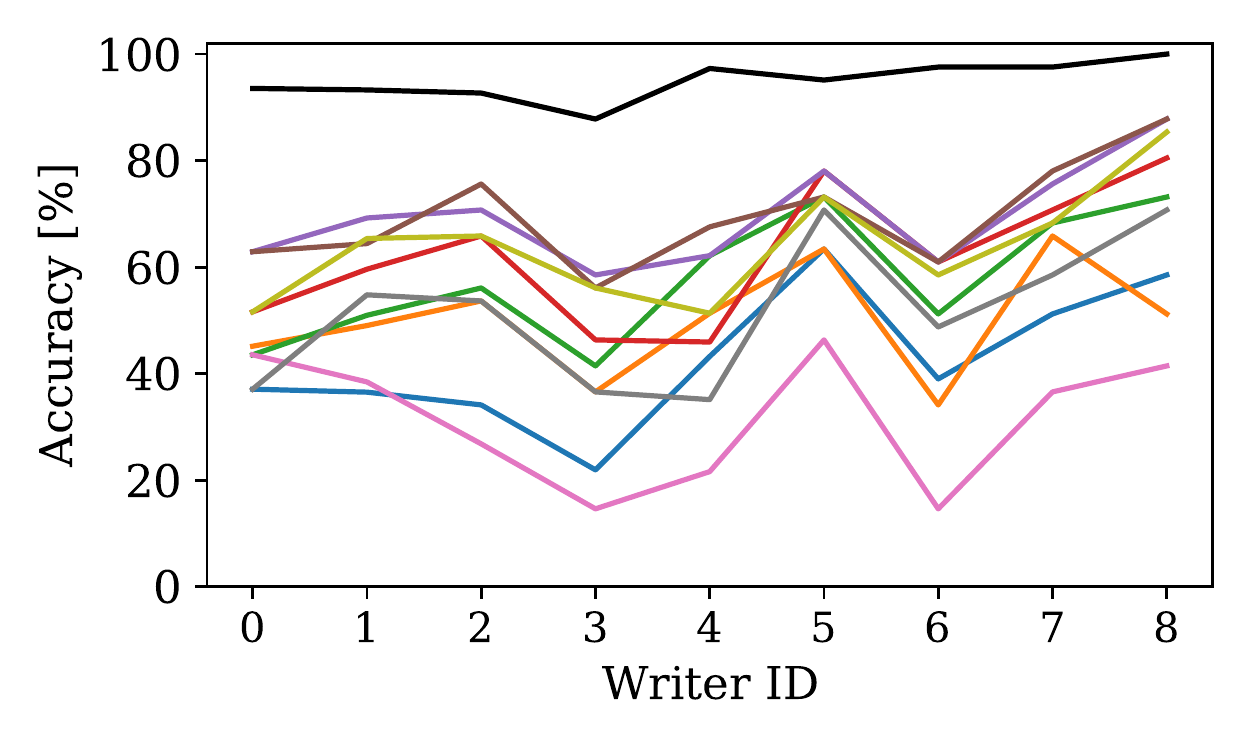}
    	\subcaption{SEMD.}
    	\label{image_writer_comb_2}
    \end{minipage}
    \hfill
	\begin{minipage}[b]{0.196\linewidth}
        \centering
    	\includegraphics[trim=12 14 10 11, clip, width=1.0\linewidth]{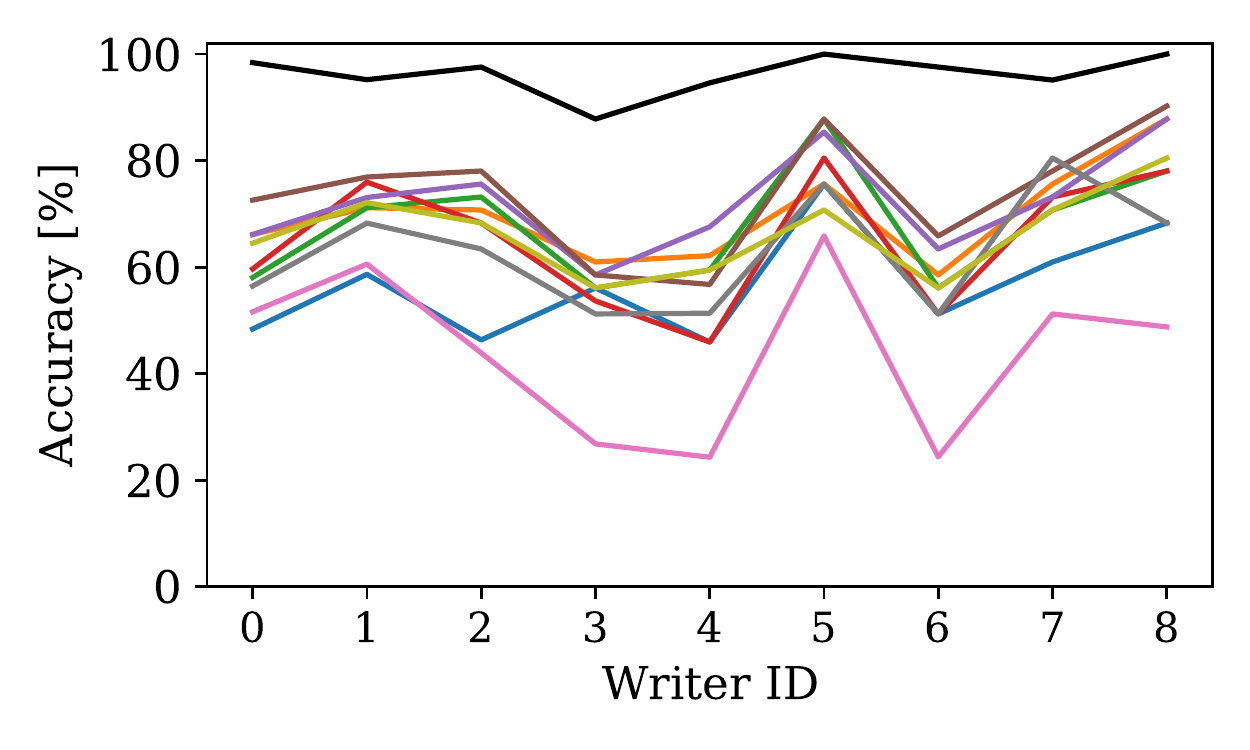}
    	\subcaption{Sinkhorn.}
    	\label{image_writer_comb_3}
    \end{minipage}
    \hfill
	\begin{minipage}[b]{0.196\linewidth}
        \centering
    	\includegraphics[trim=12 14 10 11, clip, width=1.0\linewidth]{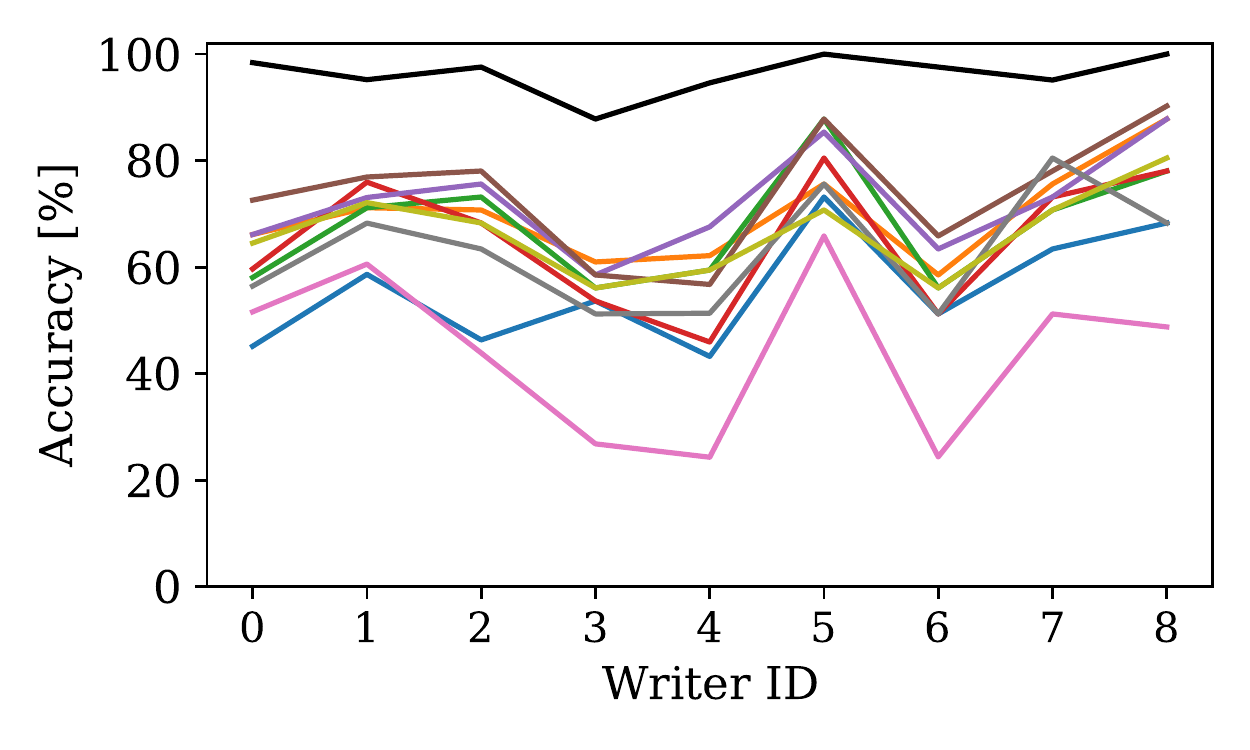}
    	\subcaption{Sinkhorn ($L_{p}L_{1}$).}
    	\label{image_writer_comb_4}
    \end{minipage}
    \hfill
	\begin{minipage}[b]{0.196\linewidth}
        \centering
    	\includegraphics[trim=12 14 10 11, clip, width=1.0\linewidth]{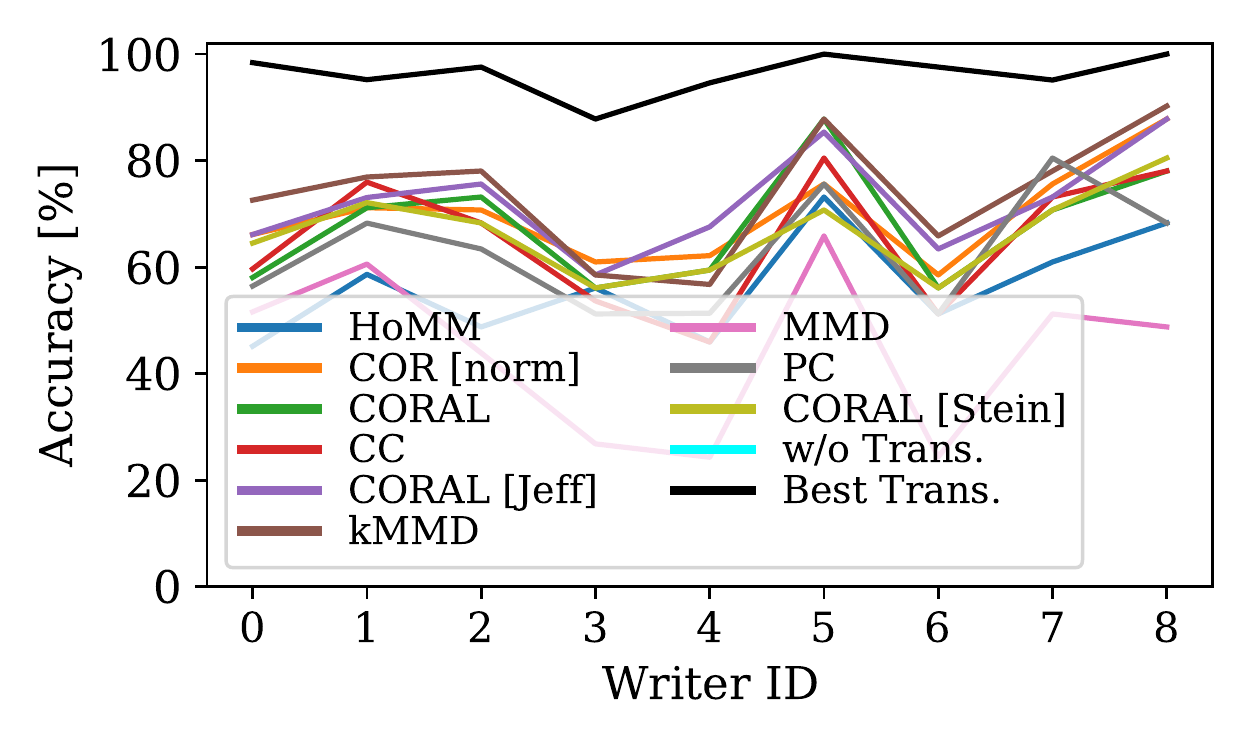}
    	\subcaption{Sinkhorn ($L_{1}L_{2}$).}
    	\label{image_writer_comb_5}
    \end{minipage}
    \vspace{-0.2cm}
    \caption{Evaluation of the transformed embeddings for the left-handed OnHW-chars (combined) dataset for each writer.}
    \label{image_writer_comb}
\end{figure*}

\begin{figure*}[t!]
	\centering
	\begin{minipage}[b]{0.196\linewidth}
        \centering
    	\includegraphics[trim=12 14 10 11, clip, width=1.0\linewidth]{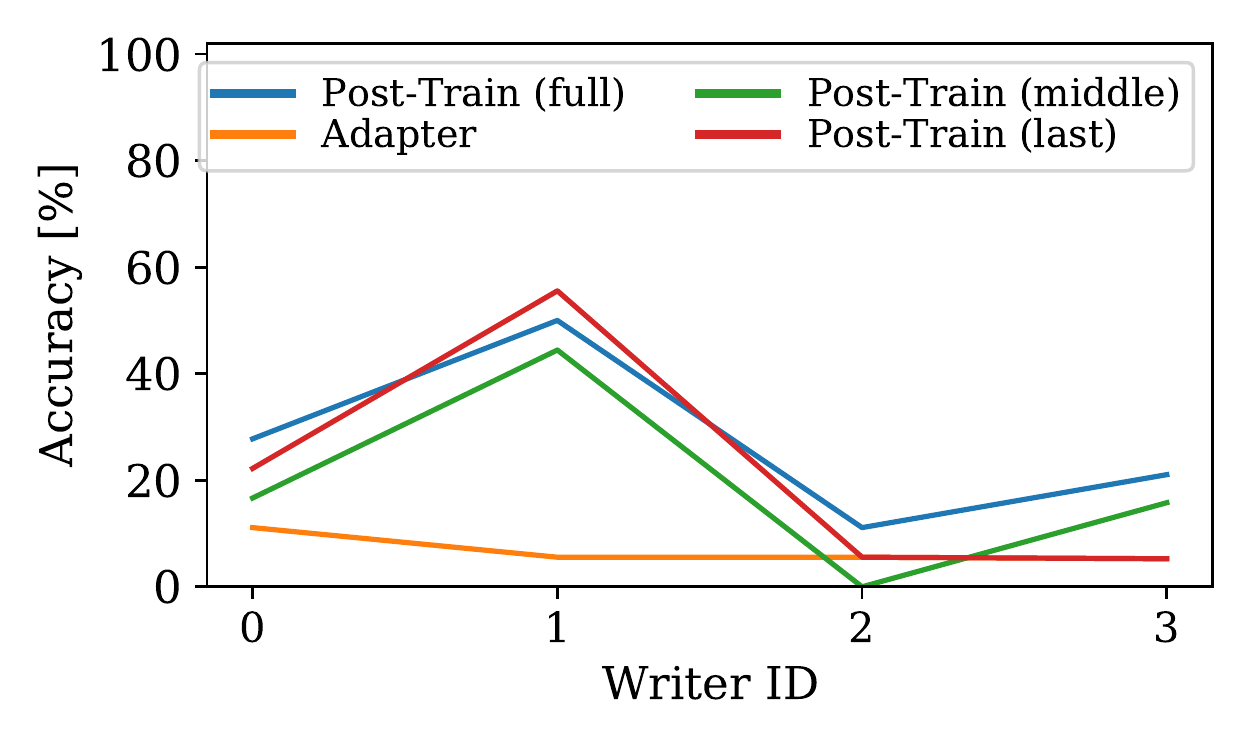}
    	\subcaption{OnHW-symbols.}
    	\label{image_post_training_1}
    \end{minipage}
    \hfill
	\begin{minipage}[b]{0.196\linewidth}
        \centering
    	\includegraphics[trim=12 14 10 11, clip, width=1.0\linewidth]{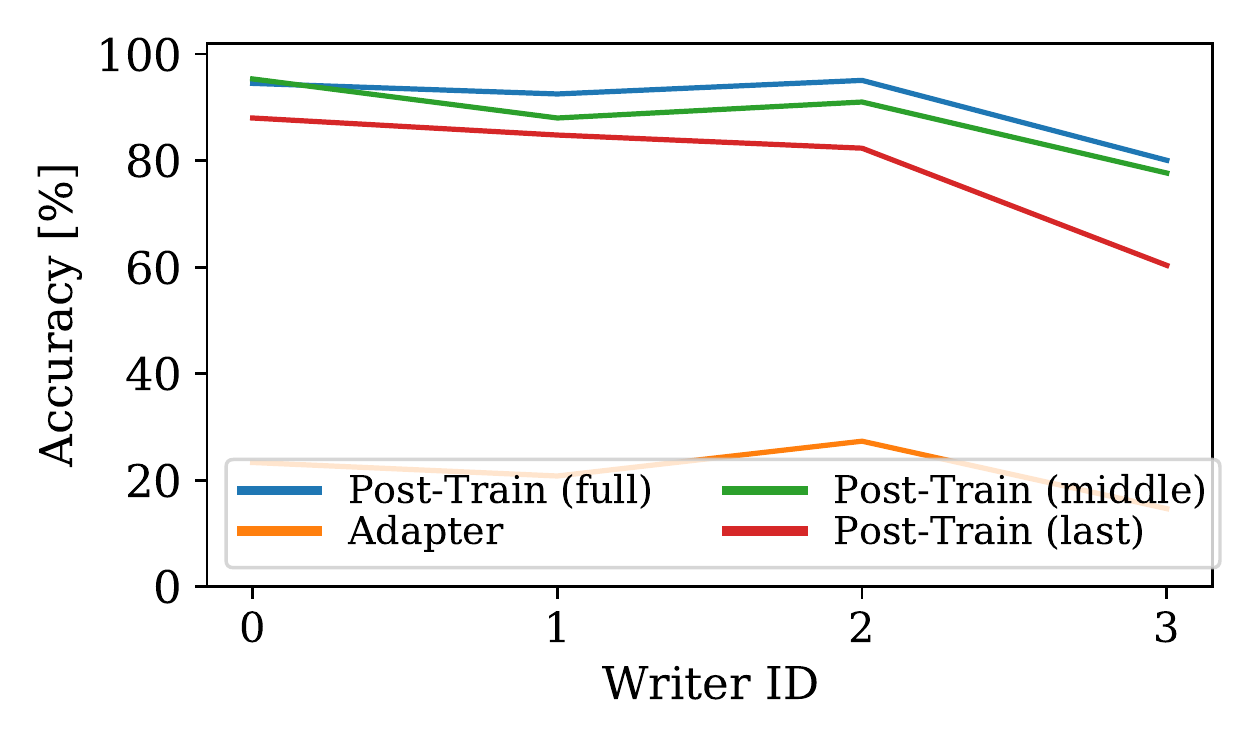}
    	\subcaption{Split OnHW-equations.}
    	\label{image_post_training_2}
    \end{minipage}
    \hfill
	\begin{minipage}[b]{0.196\linewidth}
        \centering
    	\includegraphics[trim=12 14 10 11, clip, width=1.0\linewidth]{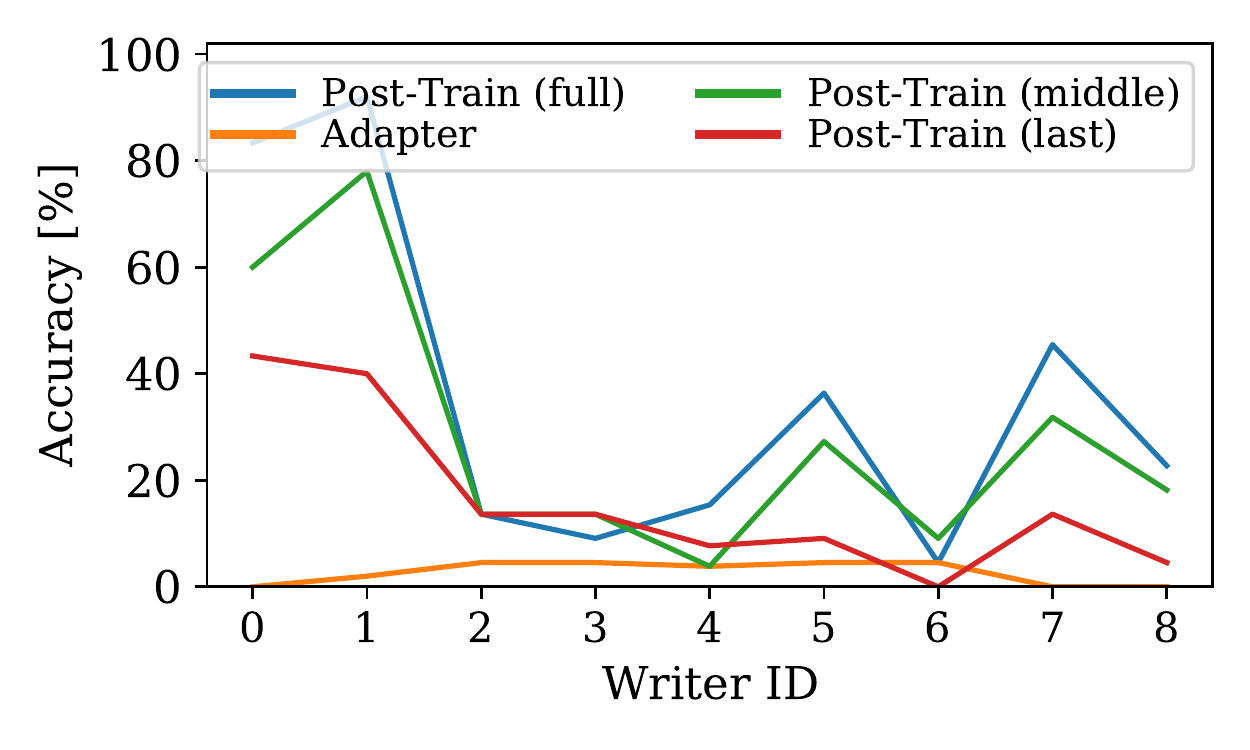}
    	\subcaption{OnHW-chars (lower).}
    	\label{image_post_training_3}
    \end{minipage}
    \hfill
	\begin{minipage}[b]{0.196\linewidth}
        \centering
    	\includegraphics[trim=12 14 10 11, clip, width=1.0\linewidth]{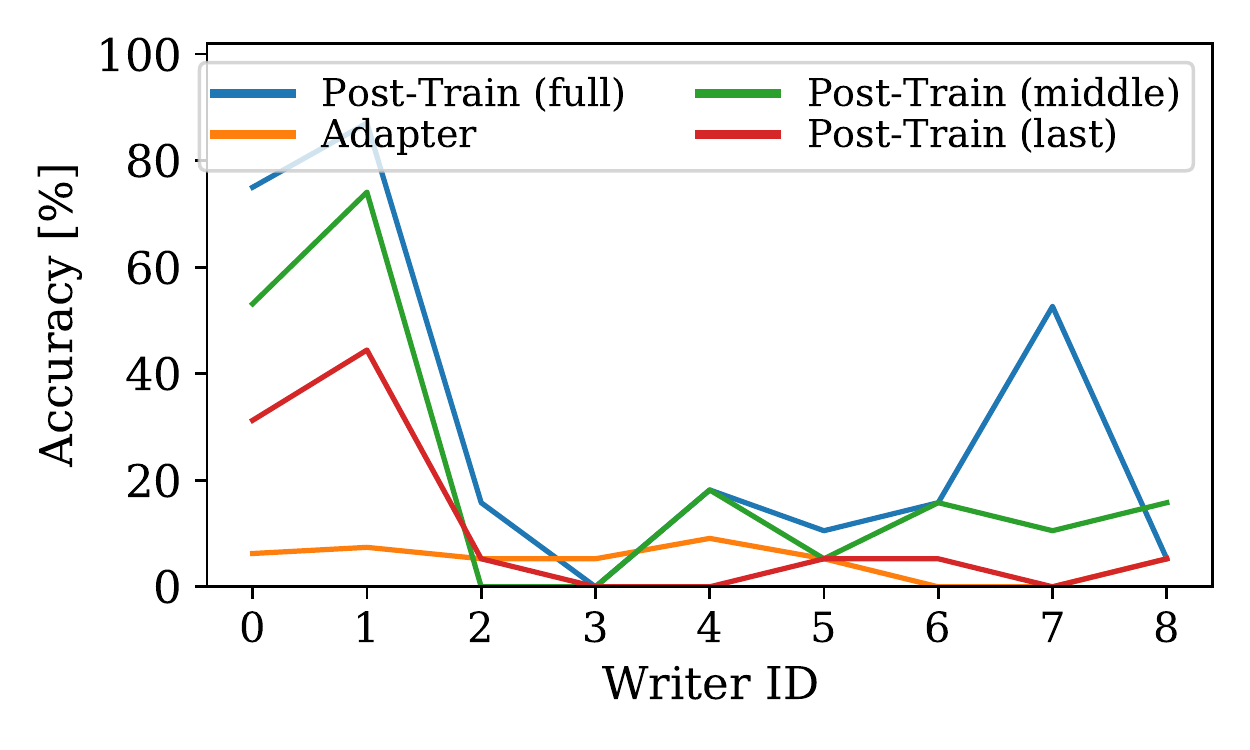}
    	\subcaption{OnHW-chars (upper).}
    	\label{image_post_training_4}
    \end{minipage}
    \hfill
	\begin{minipage}[b]{0.196\linewidth}
        \centering
    	\includegraphics[trim=12 14 10 11, clip, width=1.0\linewidth]{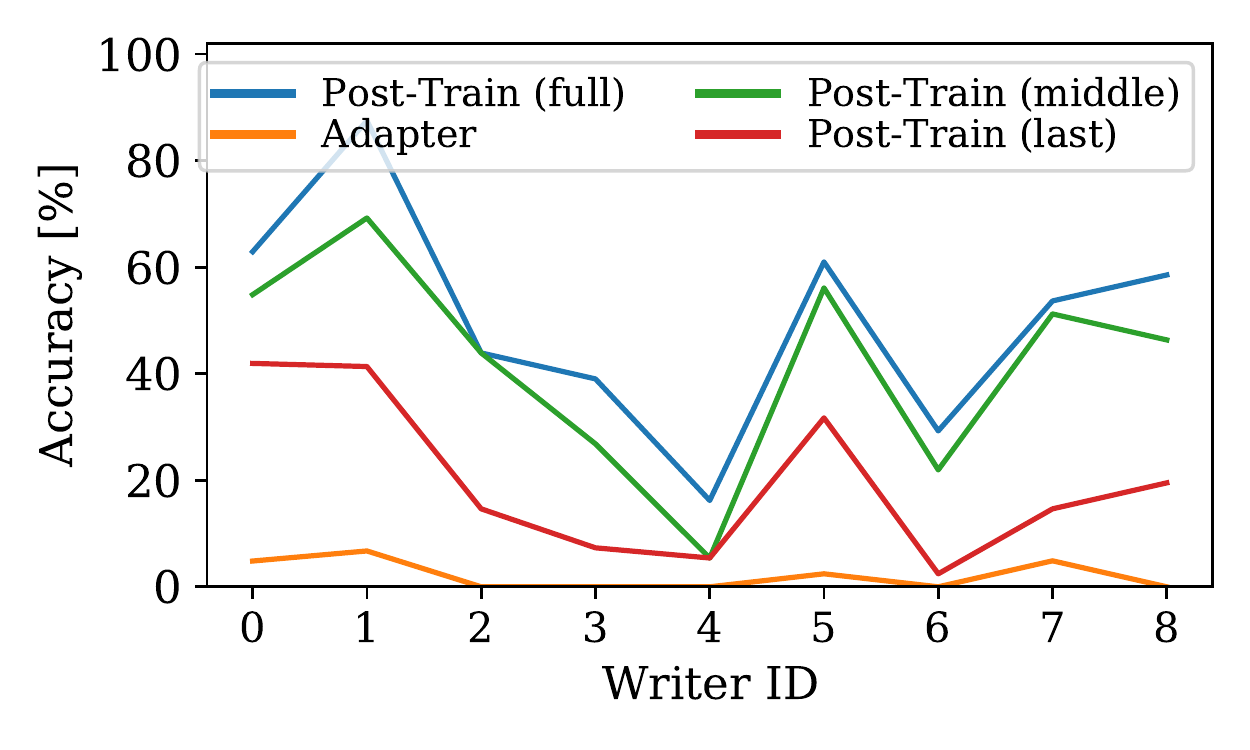}
    	\subcaption{OnHW-chars (comb.).}
    	\label{image_post_training_5}
    \end{minipage}
    \vspace{-0.2cm}
    \caption{Evaluation of transfer learning for the OnHW datasets for each of the four, respectively nine, writers.}
    \label{image_post_training}
\end{figure*}

\subsection{Deep Metric Learning Searches}
\label{sec_app_searches}

We perform a large hyperparameter search on the OnHW-symbols dataset for all optimal transport techniques. Results are shown in Figure~\ref{image_symbols_search}. The differences between EMD and SEMD are marginally, as well as between $L_{p}L_{1}$ and $L_{1}L_{2}$ regularization techniques. In general Sinkhorn performs better than EMD. While the $log$ metric performs best for EMD, $median$, $max$ and $loglog$ yield better results for Sinkhorn. The differences for distance metrics are marginal for Sinkhorn. For follow-up trainings, we choose the squared Euclidean distance metric with $loglog$ metric. Indeed, the similarity comparison method for transformation selection has the highest impact. kMMD consistently yields the highest classification accuracies followed by Stein and Jeff CORAL. Again, HoMM of order 3 can outperform MMD, but not kMMD.

\subsection{Evaluation per Writer}
\label{sec_app_writer_eval}

We adapt each writer separately as writer can have very different writing styles, and hence, different domains of sensor features. Figure~\ref{image_writer_symb} to Figure~\ref{image_writer_comb} shows all results for five different optimal transport techniques and transformation selection methods. Using the best transformation mostly achieves 100\%, which is the upper bound for the optimal transformation. Without transformation, the right-handed model leads to a poor accuracy of below 10\% (the lower bound). For the OnHW-symbols and split OnHW-equations datasets, the accuracy of the models drop for the writer with ID 3. For the OnHW-chars dataset, the writers with ID 3 and 6 are outlier. The reason is that these persons wrote inconsistent. Again, the difference between EMD and SEMD, and Sinkhorn with and without regularization is marginal in accuracy.

Figure~\ref{image_post_training} shows results of transfer learning separated for all writers for all OnHW datasets. Consistently, adapting an additional layers yields the lowest classification results. While the OnHW-symbols dataset is rather small, all post-training techniques fail to successfully classify symbols (see Figure~\ref{image_post_training_1}). The split OnHW-equations dataset is large, and hence, post-training results in classification accuracies between 60\% and 100\% (see Figure~\ref{image_post_training_2}). Results for the OnHW-chars lower, upper and combined are similar (see Figure~\ref{image_post_training_3} to \ref{image_post_training_5}). Post-training the full model yields marginally better results than training only the last layer or the layers after the middle of the model. Hence, the model requires to overfit on the specific writer. Results highly vary with the writer ID.

\subsection{Details on the Architectures}
\label{sec_app_architecture}

\begin{table}[t!]
\begin{center}
\setlength{\tabcolsep}{3.7pt}
    \caption{Architecture details for the sinusoidal datasets.}
    \label{table_architecture_sinus}
    \small \begin{tabular}{ p{0.5cm} | p{0.5cm} }
    \multicolumn{1}{c|}{\textbf{Layer}} & \multicolumn{1}{c}{\textbf{Features}} \\ \hline
    \multicolumn{1}{l|}{\textbf{Encoder}} & \multicolumn{1}{l}{} \\
    \multicolumn{1}{l|}{Input} & \multicolumn{1}{l}{Size: (Sequence length = 200, 13)} \\
    \multicolumn{1}{l|}{Convolution (1D)} & \multicolumn{1}{l}{Filters: 50, kernel size: 4, activation: \textit{softmax}} \\
    \multicolumn{1}{l|}{Max Pooling (1D)} & \multicolumn{1}{l}{Pool size: 4} \\ \hline
    \multicolumn{1}{l|}{\textbf{Classifier}} & \multicolumn{1}{l}{} \\
    \multicolumn{1}{l|}{Batch Normalization} & \multicolumn{1}{l}{} \\
    \multicolumn{1}{l|}{Dropout} & \multicolumn{1}{l}{Rate 20\%} \\
    \multicolumn{1}{l|}{LSTM} & \multicolumn{1}{l}{Units: 10, activation: \textit{tanh}} \\
    \multicolumn{1}{l|}{Dense} & \multicolumn{1}{l}{Units: 20} \\
    \multicolumn{1}{l|}{Output} & \multicolumn{1}{l}{Size: \textit{number classes} (10)} \\
    \end{tabular}
\end{center}
\end{table}

\begin{table}[t!]
\begin{center}
\setlength{\tabcolsep}{3.7pt}
    \caption{Architecture details for the OnHW datasets.}
    \label{table_architecture_onhw}
    \small \begin{tabular}{ p{0.5cm} | p{0.5cm} }
    \multicolumn{1}{c|}{\textbf{Layer}} & \multicolumn{1}{c}{\textbf{Features}} \\ \hline
    \multicolumn{1}{l|}{\textbf{Encoder}} & \multicolumn{1}{l}{} \\
    \multicolumn{1}{l|}{Input} & \multicolumn{1}{l}{Size: (Sequence length, 1)} \\
    \multicolumn{1}{l|}{Convolution (1D)} & \multicolumn{1}{l}{Filters: 200, kernel size: 4, activation: \textit{relu}} \\
    \multicolumn{1}{l|}{Max Pooling (1D)} & \multicolumn{1}{l}{Pool size: 2} \\
    \multicolumn{1}{l|}{Batch Normalization} & \multicolumn{1}{l}{} \\
    \multicolumn{1}{l|}{Dropout} & \multicolumn{1}{l}{Rate: 20\%} \\
    \multicolumn{1}{l|}{Convolution (1D)} & \multicolumn{1}{l}{Filters: 200, kernel size: 4, activation: \textit{relu}} \\
    \multicolumn{1}{l|}{Max Pooling (1D)} & \multicolumn{1}{l}{Pool size: 2} \\ \hline
    \multicolumn{1}{l|}{\textbf{Classifier}} & \multicolumn{1}{l}{} \\
    \multicolumn{1}{l|}{Batch Normalization} & \multicolumn{1}{l}{} \\
    \multicolumn{1}{l|}{Dropout} & \multicolumn{1}{l}{Rate: 20\%} \\
    \multicolumn{1}{l|}{BiLSTM} & \multicolumn{1}{l}{Units: 60, activation: \textit{tanh}, return sequ.: True} \\
    \multicolumn{1}{l|}{BiLSTM} & \multicolumn{1}{l}{Units: 60, activation: \textit{tanh}, return sequ.: True} \\
    \multicolumn{1}{l|}{Dense} & \multicolumn{1}{l}{Units: 100} \\
    \multicolumn{1}{l|}{Dense (time distributed)} & \multicolumn{1}{l}{Units: \textit{number classes}, activation: \textit{softmax}} \\
    \multicolumn{1}{l|}{Output} & \multicolumn{1}{l}{Size: \textit{number classes}} \\
    \end{tabular}
\end{center}
\end{table}

Table~\ref{table_architecture_sinus} and \ref{table_architecture_onhw} show architecture details for the sinusoidal dataset, and OnHW datasets, respectively. Both models contain a feature extractor of the time-series datasets, and temporal unit, and \textit{dense} layers for classification. We use a small LSTM of 10 units for the synthetic dataset, and two stacked bidirectional LSTM layers of 60 units each for the OnHW datasets. For the optimal transport methods, we use the output of the \textit{max pooling} layer before the \textit{batch normalization} and \textit{dropout} layers and the temporal units. We train the synthetic dataset for 100 epochs, a batch size of 100, the Adam optimizer with learning rate 0.0001, and the categorical cross-entropy loss. We train the OnHW datasets for 1,000 epochs, a batch size of 50, the Adam optimizer with learning rate 0.0001, and the categorical cross-entropy loss. For transfer learning techniques we train the synthetic dataset for 80 epochs, and the OnHW datasets for 100 epochs with the same optimizer parameters.

\end{document}